%% file: arXiv.tex
\documentclass[11pt]{article}
\PassOptionsToPackage{numbers, sort&compress}{natbib}

\usepackage{neuroai-epfl}%
\usepackage{graphicx}%
\usepackage{multirow}%
\usepackage{amsmath,amssymb,amsfonts}%
\usepackage{amsthm}%
\usepackage{mathrsfs}%
\usepackage[title]{appendix}%
\usepackage{xcolor}%
\usepackage{textcomp}%
\usepackage{manyfoot}%
\usepackage{booktabs}%
\usepackage{algorithm}%
\usepackage{algorithmicx}%
\usepackage{algpseudocode}%
\usepackage{xspace}%
\usepackage{listings}%
\usepackage{caption}%
\usepackage{natbib}

\usepackage[utf8]{inputenc} %
\usepackage[T1]{fontenc}    %
\usepackage{hyperref}       %
\usepackage{url}            %
\usepackage{xurl}           %
\usepackage{nicefrac}       %
\usepackage{microtype}      %
\usepackage{multirow}
\usepackage{threeparttable}
\usepackage{array}

\hypersetup{
  pdftitle={MIRAGE: Adaptive Multimodal Gating for Whole-Brain fMRI Encoding},
  pdfauthor={Abdulkadir Gokce, Badr AlKhamissi, Martin Schrimpf},
  pdfpagemode=UseOutlines,
  bookmarksopen=true,
  bookmarksnumbered=true,
}

\makeatletter
\newcommand{\supplementarysection}{%
  \setcounter{figure}{0}%
  \setcounter{table}{0}%
  \let\oldthefigure\thefigure%
  \let\oldthetable\thetable%
  \renewcommand{\thefigure}{S\oldthefigure}%
  \renewcommand{\thetable}{S\oldthetable}%
  \@ifundefined{theHtable}{}{\renewcommand{\theHtable}{S\oldthetable}\renewcommand{\theHfigure}{S\oldthefigure}}%
}

\usepackage{etoolbox}
\AtBeginEnvironment{table}{\def\arxiv@intable{}}
\AtBeginEnvironment{tabular}{%
  \ifdefined\arxiv@intable
    \ifdim\f@size pt>10.5pt \small
    \else\ifdim\f@size pt>9.5pt \footnotesize
    \else\ifdim\f@size pt>8.5pt \scriptsize
    \else \tiny\fi\fi\fi
  \fi}

\let\arxiv@origfigure\figure
\def\figure{\@ifnextchar[\arxiv@firstfigure{\arxiv@firstfigure[]}}
\def\arxiv@firstfigure[#1]{%
  \global\let\figure\arxiv@origfigure
  \arxiv@origfigure[t]}
\makeatother

\begin{document}

\definecolor{ggreen}{HTML}{00A64F}
\definecolor{light-gray}{gray}{0.9}
\let\eb\eqnmarkbox
\newcommand*{\tightcolorbox}[2]{%
    \begingroup\setlength{\fboxsep}{1pt}%
        \colorbox{#1}{{\hspace*{2pt}\vphantom{Ay}#2\hspace*{2pt}}}%
    \endgroup
}
\newcommand*{\code}[1]{\tightcolorbox{light-gray}{\texttt{#1}}}
\newcommand*{\modelname}[1]{{\textsc{#1}}}
\newcommand*{\datasetname}[1]{{\textsc{#1}}}
\newcommand{\ourmodel}{\textsc{MIRAGE}\xspace}

\begin{titlecard}

\papertitle{MIRAGE: Adaptive Multimodal Gating\\for Whole-Brain fMRI Encoding}

\paperauthors{%
    \authorlink{https://akgokce.github.io}{Abdulkadir Gokce}\textsuperscript{*},\
    \authorlink{https://bkhmsi.github.io}{Badr AlKhamissi}\textsuperscript{*},\
    \authorlink{https://mschrimpf.com}{Martin Schrimpf}%
}

\paperaffiliations{%
    NeuroAI Lab, EPFL\hspace{0.6cm}
    \textsuperscript{*}Equal Contribution\hspace{0.6cm}
    Published at NeurIPS 2026 (\href{https://openreview.net/forum?id=o0Kim6sdzg}{OpenReview})%
}

\paperabstract{%
Recent progress in task-optimized neural networks has established encoding models as a powerful tool for predicting brain responses to naturalistic stimuli, yet most existing approaches rely on unimodal representations. The emergence of omni-modal foundation models and rich multimodal neural datasets enables encoding models that jointly integrate visual, auditory, and linguistic information across subjects.
We introduce \ourmodel, a brain encoding framework for predicting whole-brain fMRI responses to naturalistic audiovisual stimuli with paired transcripts.
\ourmodel\ extracts representations from a single pretrained omni-modal backbone through three modality-specific cross-attention modules whose latent queries adaptively aggregate features across the backbone's $48$ layers, and combines them through a transformer-based brain encoder and a subject-specific linear head over the cortical parcels.
\ourmodel achieves state-of-the-art results on the CNeuroMod/Algonauts out-of-distribution benchmark.
Within the Qwen-Omni family, controlled comparisons show that natively fused features outperform post-hoc fusion of independently extracted unimodal streams at every architectural level and backbone scale, and retain a small but consistent advantage over a depth- and representation-matched control. The frozen representation also transfers to held-out participants and to two external movie-watching fMRI datasets.
Beyond predictive accuracy, the learned attention weights are directly inspectable: each modality’s gating module discovers a distinct depth profile over the backbone, and each modality traces a distinct, anatomically structured pattern across cortex.
Together, these results establish adaptive layer-wise aggregation of natively multimodal features as a generalizable, interpretable, and accurate approach for whole-brain encoding.%
}

\papermeta{%
    \metaentry{Correspondence}{%
        \href{mailto:abdulkadir.gokce@epfl.ch}{abdulkadir.gokce@epfl.ch},\
        \href{mailto:badr.alkhamissi@epfl.ch}{badr.alkhamissi@epfl.ch},\
        \href{mailto:martin.schrimpf@epfl.ch}{martin.schrimpf@epfl.ch}}
    \metaentry{Code}{\metalink[github.com/epflneuroailab/mirage]%
        {https://github.com/epflneuroailab/mirage}}
    \metaentry{Model}{\metalink[huggingface.co/epfl-neuroai/mirage]%
        {https://huggingface.co/epfl-neuroai/mirage}}
    \metaentry{Website}{\metalink[mirage-brain.epfl.ch]%
        {https://mirage-brain.epfl.ch}}
}

\end{titlecard}

\section{Introduction}

\input{sections/01-intro}

\section{Methodology}
\label{sec:methods}

\input{sections/03-methods}

\section{Results}
\label{sec:results}

\input{sections/04-results}

\section{Discussion \& Future Work}
\label{sec:discussion}

\input{sections/05-discussion}

\section{Related Work}

\input{sections/02-related_work}

\section{Conclusion}
\label{sec:conclusion}

\input{sections/06-conclusion}

\section*{Acknowledgements}
This work was supported by the Swiss National Science Foundation (SNSF; Grant No.\ 10.003.772; A.G. and B.A.) and the Swiss AI Initiative (2025 Fellowship Program; A.G. and B.A.).  This work was supported as part of the ``Swiss AI initiative'' by a grant from the Swiss National Supercomputing Centre (CSCS) under project ID a0173 on Alps. We thank the members of the EPFL NeuroAI Lab for their valuable discussions and feedback throughout the project, the Courtois NeuroMod project and the Algonauts 2025 organizers for the data and the evaluation platform, and the CineBrain authors for releasing their dataset.

\textbf{HCP data use.} Data were provided in part by the Human Connectome Project, WU-Minn Consortium (Principal Investigators: David Van Essen and Kamil Ugurbil; 1U54MH091657) funded by the 16 NIH Institutes and Centers that support the NIH Blueprint for Neuroscience Research; and by the McDonnell Center for Systems Neuroscience at Washington University.

\bibliographystyle{unsrtnat}

\bibliography{references}

\clearpage
\input{appendices/00-appendix}

\end{document}

%% file: sections/01-intro.tex
Humans perceive the natural world as integrated streams of multimodal information, rather than as isolated sensory channels. A central goal of computational neuroscience is to explain how such multimodal sensory information is represented in the brain. Encoding models provide a quantitative framework for this goal: representations from task-optimized neural networks can predict held-out neural responses and reveal which computational variables align with cortical activity \citep{Yamins_Hong_Cadieu_Solomon_Seibert_DiCarlo_2014,yamins2016,SchrimpfKubilius2018BrainScore,Schrimpf2021Language,richards2019_dl4neuroscience,conwell2024bias,gokce2025scaling, tang2025manytwoone,dascoli2026tribev1}. Most progress, however, has been made with unimodal stimuli and unimodal backbones, often paired with simple linear or ridge readouts \citep{Schrimpf2021Language,alkhamissi2025language2cognition,styves2023brainoptimized_nature,adeli2025nsd_transformer_scale}.

\begin{figure}[ht]
    \centering
    \includegraphics[width=1\linewidth]{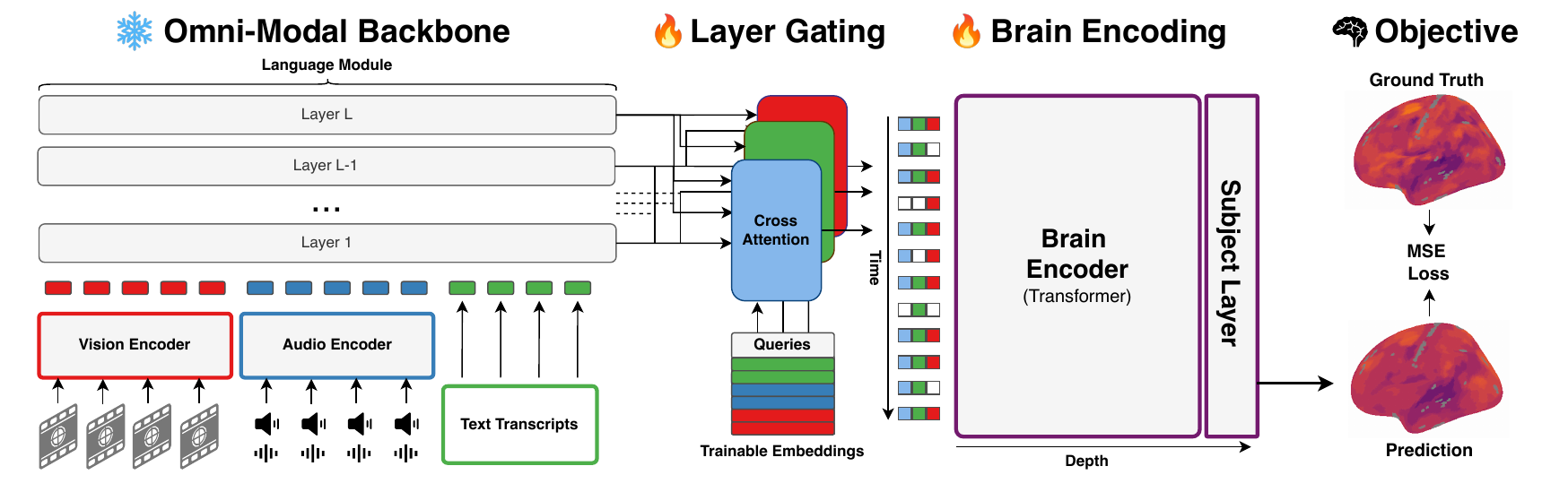}
    \caption{
    \textbf{\ourmodel architecture for predicting brain responses from naturalistic multimodal stimuli.} Aligned video, audio, and text-transcript inputs are encoded by \texttt{Qwen3-Omni-30B-A3B-Thinking} (left, frozen), exposing the hidden states of every layer of its Language Module. Modality-specific trainable Layer Gating modules each use a small bank of learnable query embeddings in a cross-attention block to aggregate information across all $L=48$ layers, producing one pooled feature vector per modality per time step, which are concatenated along the hidden dimension. The resulting time-aligned sequence, colored by source modality (vision in red, audio in blue, text in green), is passed to the Brain Encoding module: a transformer applied along the time axis, followed by a per-subject linear projection (Subject Layer) that maps to cortical parcels. Only the gating and encoding parameters are trained; the model is optimized end-to-end by minimizing the mean-squared error between predicted and measured (ground-truth) fMRI responses.
    }
    \label{fig:graphical-abstract}
\end{figure}

Naturalistic fMRI datasets in which subjects watch movies now capture the temporal interplay of vision, speech, and language~\citep{boyle2020courtois_cneuromod,gifford2025algonauts}, and combining these streams improves brain prediction over unimodal features~\citep{dascoli2026tribev1,schad2025-algonauts2025_vibe,eren2025-algonauts2025_rnn}. Yet most such pipelines still treat multimodal integration as a downstream brain-mapping problem: features are extracted from separate modality-specific models and fused only later by the temporal encoder or neural readout.

This raises a basic question: are the cross-modal interactions learned during multimodal pretraining themselves brain-relevant, or is fusion better deferred to the downstream readout? 
Beyond where fusion happens, many encoding pipelines use rigid feature aggregation, a single selected layer, a fixed group of layers, or concatenated features from predefined streams~\citep{gokce2025scaling,schad2025-algonauts2025_vibe,eren2025-algonauts2025_rnn}, which is poorly matched to a cortex in which different regions may align with different representational depths~\citep{yamins2016,conwell2024bias,tang2025manytwoone}. Whole-brain encoding therefore calls for adaptive, depth-aware layer aggregation.

We introduce \textbf{\ourmodel} (\textbf{M}ultimodal \textbf{I}ntegration with \textbf{R}epresentation-\textbf{A}daptive \textbf{G}ated \textbf{E}ncoding), an adaptive multimodal gating encoding framework for whole-brain fMRI prediction from naturalistic movies. Unlike prior approaches that combine separate vision, audio, and language models, \ourmodel uses a single multimodal foundation model as the stimulus feature encoder.
This enables a controlled comparison between modality-specific streams, a depth-matched unimodal control, and native-fusion representations produced by the encoder's own multimodal integration mechanism, with the downstream temporal encoder and neural readout held fixed.

\ourmodel\ disentangles components that are often conflated in encoding pipelines: frozen feature extraction, learned per-modality layer aggregation, modality fusion, and a temporal brain encoder with subject-specific readout. This design makes multimodal fusion an explicit experimental variable while allowing the readout to adapt across cortical regions and subjects.

\paragraph{Contributions.}
Our contributions are fourfold: 
\emph{(i)} we introduce \ourmodel, a brain encoding framework that predicts whole-brain fMRI responses to naturalistic audiovisual stimuli by combining a multimodal foundation model with per-modality adaptive layer aggregation through learned latent queries; 
\emph{(ii)} through controlled comparisons at every architectural level and across backbone scales, we show that native multimodal fusion improves on post-hoc fusion of independently extracted unimodal streams, an effect that survives a depth- and representation-matched control within the Qwen-Omni family;
\emph{(iii)} we show that \ourmodel's learned attention weights are directly interpretable, exposing modality-specific depth profiles over the backbone and spatially structured modality contributions across cortex; and 
\emph{(iv)} \ourmodel\ achieves state-of-the-art results on the CNeuroMod/Algonauts~2025 challenge out-of-distribution benchmark, and its frozen representation transfers to held-out participants and to two external datasets, HCP 7T Movie and CineBrain.

Together, our results suggest that multimodal brain encoding should treat fusion as a property of the features rather than a downstream readout operation, and adaptively fuse features across the processing hierarchy rather than selecting only a single fixed layer.

%% file: sections/03-methods.tex
\subsection{Problem Setup and Data}
We use the Algonauts 2025 challenge data~\citep{gifford2025algonauts}, derived from the Courtois NeuroMod project~\citep{boyle2020courtois_cneuromod}. The release provides whole-brain fMRI from $N_S=4$ subjects watching the television series \emph{Friends} and a curated movie set (\emph{Movie10}). We predict whole-brain fMRI responses to naturalistic movie stimuli from time-aligned visual, auditory, and language streams. Each training example is a stimulus window $x_i$, a subject index $s_i$, and a target response $y_i \in \mathbb{R}^{K \times P}$, where $P=1000$ denotes cortical parcels and $K=100$ denotes the number of fMRI samples in the window. Targets are sampled at the fMRI repetition time (TR) of $1.49$~s, so each window covers $\approx 149$~s of stimulus, and features are aligned to a common 2~Hz feature grid before TR pooling.

We hold out \emph{Friends} season~6 as the internal validation split used for model selection and ensembling, while the in-distribution challenge test \emph{Friends} season~7 and a curated out-of-distribution (OOD) movie set are accessed via the Codabench evaluation platform as test sets. Performance is measured by Pearson correlation between predicted and measured BOLD responses, computed per parcel across all time points and then averaged over parcels and subjects, matching the primary objective of the Algonauts 2025 benchmark.

\subsection{Multi-Layer Multimodal Features}
For each modality $m \in \{\mathrm{text}, \mathrm{audio}, \mathrm{vision}\}$, we extract a multi-layer feature tensor
\begin{equation*}
    H_i^m \in \mathbb{R}^{L_m \times T_i \times d_m},
\end{equation*}
where $L_m$ is the number of backbone layers, $T_i$ is the number of stimulus frames on a 2 Hz grid, and $d_m$ is the hidden dimension. %
We evaluate two feature families. The first follows the TRIBE-style unimodal pipeline \citep{dascoli2026tribev1}, with text features from Llama-3.2-3B \citep{grattafiori2024llama3herdmodels}, audio features from Wav2Vec-Bert-2.0 \citep{chung2021wav2vecbert}, and video features from V-JEPA~2 \citep{assran2025vjepa2}, all resampled to a common 2~Hz grid. The second uses the Qwen-Omni multimodal backbones \citep{xu2025qwen25omnitechnicalreport,xu2025qwen3omnitechnicalreport}, from which we read features under the three fusion conditions described below, giving a controlled comparison within the \emph{same} backbone.

\paragraph{Three fusion conditions.}
Using a multimodal backbone differs from a set of unimodal encoders in two ways: where the features of each modality are read, and whether the modalities interact while they are computed. We therefore compare three conditions that share the same downstream model, optimizer, schedule, and seed and differ only in the cached features.
\emph{Post-hoc (tower)} reads vision and audio from their encoder towers and text from the language module applied to the transcript alone; this is the standard pipeline that concatenates independently extracted unimodal features.
\emph{Post-hoc (decoder)} passes each modality through the language module in a separate forward pass and reads all three streams from the same layers as the native condition, so every stream receives the same computation and lives in the same representational space without attending to the other modalities.
\emph{Native fusion} passes all modalities together so their tokens attend to one another inside the language module. \ourmodel\ uses this condition with \modelname{Qwen3-Omni-30B-A3B-Thinking} (extraction details in App.~\ref{app:fusion-conditions}).

\subsection{Adaptive Layer Aggregation}

Fixed layer selection is a strong but restrictive baseline, as it assumes that the same representational depth is appropriate across modalities, subjects, and brain regions. \ourmodel\ instead learns an aggregation over the layer axis before mapping features to neural responses. For each modality we instantiate a dedicated cross-attention pooler with a bank of $n_q$ learnable query embeddings $Q^{(m)} \in \mathbb{R}^{n_q \times d_m}$. At every time step $t$, each query cross-attends to the layer tokens $H^m_{:,t,:}$ of that
modality:
\begin{equation*}
\tilde a_t^{m,q}
  = \sum_{\ell=1}^{L_m} \pi^{m,q}_{t,\ell}\, V^m_{\ell,t},
\quad
\pi^{m,q}_{t,\ell}
  = \mathrm{softmax}_\ell\!\left(
       \frac{(Q^{(m)}_q)^{\top} K^m_{\ell,t}}{\sqrt{d_m / h}}
     \right),
\quad q = 1,\dots,n_q,
\end{equation*}
implemented as standard multi-head attention with $h$ heads (single-head form shown; $K^m, V^m$ are linear projections of $H^m$). Rather than reducing each modality to a single weighted layer average, we concatenate the $n_q$ query outputs along the hidden dimension,

\begin{equation*}
a_t^m
  = \big[\tilde a_t^{m,1}\,\Vert\,\dots\,\Vert\,\tilde a_t^{m,n_q}\big]
  \in \mathbb{R}^{n_q\,d_m},
\end{equation*}
so that distinct queries can specialize to different mixtures of representational depths. \ourmodel uses $n_q = 24$ queries per modality with $h = 4$ heads and attention dropout $0.2$ (App.~\ref{app:n_queries_ablation}). We evaluate mean pooling and group means over fractional depths, as fixed baselines on the same layer axis, similar to \citet{dascoli2026tribev1}.

\subsection{Modality Fusion and Brain Encoder}

\begin{figure}[t]
    \centering
    \includegraphics[width=1\linewidth]{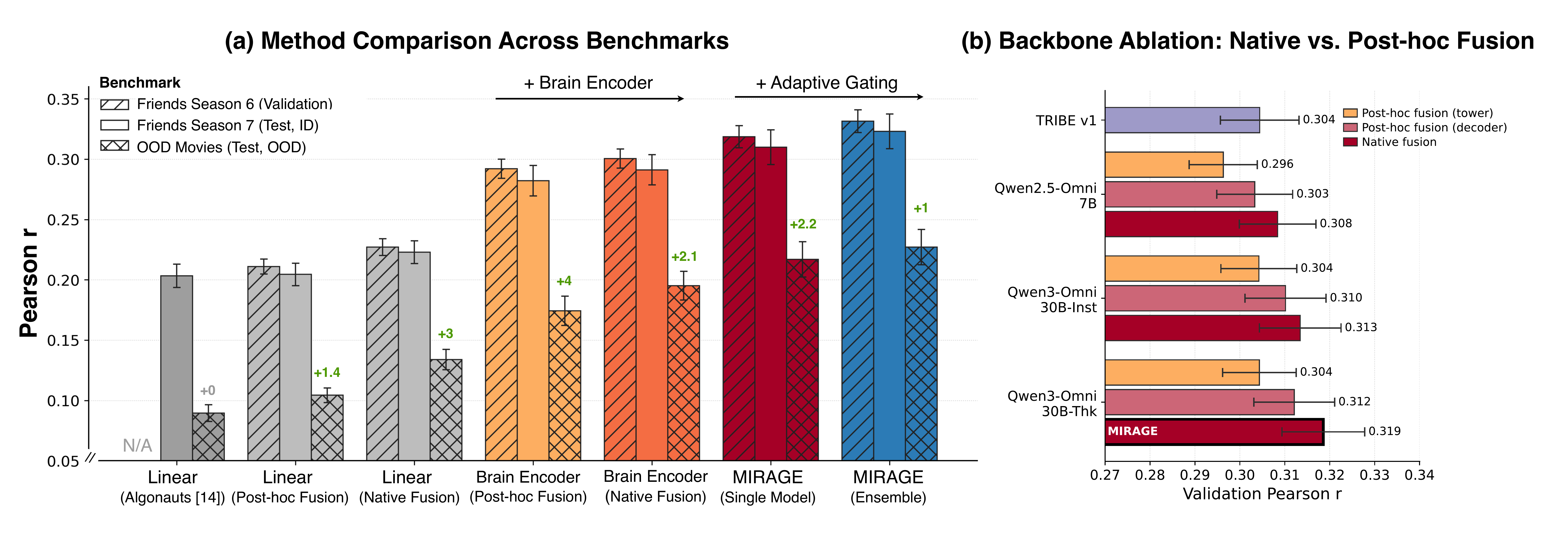}
     \caption{
    \textbf{(a) Method Comparison Across Benchmarks.} Mean Pearson $r$ between predicted and measured BOLD on the validation set (\emph{Friends} S06), the in-distribution test set (\emph{Friends} S07), and the out-of-distribution \emph{Movies} benchmark, grouped by architectural complexity: linear ridge baselines (gray), \textsc{Qwen3-Omni} features with a learned brain encoder but no cross-attention gating (orange), and \ourmodel\ as a single model (red) and as an ensemble (blue); the ensemble's S06 value is in-sample, since its parcel weights are fit on S06. Each group is shown under both post-hoc and native fusion where applicable. %
    \textbf{(b) Backbone Ablation.} 
    Pearson $r$ on the validation set when varying the feature-extraction backbone of \ourmodel: native fusion (red) against post-hoc fusion from the encoder towers (orange) and the depth-matched post-hoc fusion through the decoder (rose) for three Qwen-Omni backbones (App.~\ref{app:fusion-matched-results}), with our TRIBE~v1-style reproduction (App.~\ref{app:parameters-provenance}) as a reference. Error bars denote SEM across the four subjects.
    }
    \label{fig:main-result}
\end{figure}

\ourmodel\ separates the adaptive component from a deliberately simple fusion rule. The main learned mechanism is the per-modality \emph{layer aggregation}, while modality fusion concatenates the resulting streams. Each modality stream is mapped to a common representational space by a modality-specific linear projector, and the projected streams are concatenated to form the fused sequence. During training, modality dropout ($p = 0.3$) randomly zeroes input modalities while keeping at least one modality active, regularizing the projectors and enabling robust inference when one or two modalities are missing.

The fused sequence $u_{1:T}$ is processed by a transformer-based brain encoder \citep{transformer2017} with depth~8, 8~heads, hidden dimension $D = 3072$, and a feed-forward inner dimension of $4D$. Tokens receive both learned absolute temporal position embeddings (added before the trunk) and rotary positional embeddings (applied at each attention layer), and the trunk is shared across all subjects. The trunk outputs contextualized features $r_{1:T}$, which are mapped to parcels by a subject-specific linear head and then reduced to the target fMRI grid by adaptive average pooling along the time axis. This shares the brain encoder across subjects while allowing each subject to have an individualized mapping into parcel space; subject identity enters the model only through this readout.

\subsection{Training and Ensembling}
\label{sec:methods_traning_ensembling}
Models are trained with mean-squared error under AdamW and selected by validation Pearson's $r$. Full optimizer settings, schedule, and compute are reported in Appendix~\ref{app:training-procedure}. Final test predictions are produced by ensembling $15$ trained models, combined by parcel-wise softmax weights over each member's per-parcel validation Pearson's $r$ with temperature $\tau = 0.3$, computed independently per subject and per parcel. Per-subject ridge baselines used in \S\ref{sec:results} are detailed in App.~\ref{app:linear_baselines}.

\paragraph{Statistical testing.}
Every comparison is a paired difference in mean Pearson $r$ between models scored on identical stimuli, subjects, and parcels, with the stimulus or the participant as the independent unit. The 23 episodes of \emph{Friends} S06 (bootstrap intervals and permutation tests), the OOD movies, or the external participants (exact sign-flip and sign tests), with Holm correction within pre-declared families. \emph{Friends} S07 is descriptive, since Codabench returns only subject aggregates. We report here point estimates and consistency across subjects; intervals and corrected $p$-values for every contrast are in App.~\ref{app:statistics}.

%% file: sections/04-results.tex
\subsection{\ourmodel Achieves State-of-the-Art Brain Alignment}
\label{sec:sota}

\begin{table}[t]
\centering
\caption{Per-subject and aggregate Pearson $r$ on the validation set (\emph{Friends} s06), the in-distribution test set (\emph{Friends} s07), and the out-of-distribution benchmark. The four right-most columns report per-subject performance on OOD set. Best result in each column is shown in \textbf{bold}. TRIBE v2 numbers reflect applying its publicly released group head per subject after projecting into the benchmark fMRI space, without subject-specific readout fitting; they are a lower bound under our protocol. 
Post-hoc rows use the tower condition (\S\ref{sec:methods}). Parameter counts and the source of every row are given in App.~\ref{app:parameter-counts} and App.~\ref{app:parameters-provenance}.
}
\label{tab:results}
\small
\setlength{\tabcolsep}{4pt}
\begin{tabular}{l ccc | cccc}
\toprule
\multirow{2}{*}{\textbf{Model}} & \multicolumn{3}{c|}{\textbf{Mean Pearson $r$}} & \multicolumn{4}{c}{\textbf{OOD per-subject ($r$)}} \\
\cmidrule(lr){2-4} \cmidrule(lr){5-8}
 & Val (s06) & Test (s07) & OOD & Sub-01 & Sub-02 & Sub-03 & Sub-05 \\
\midrule
\multicolumn{8}{l}{\emph{Linear}} \\
\quad Linear \citep{gifford2025algonauts} & -- & 0.203 & 0.090 & 0.099 & 0.086 & 0.102 & 0.071 \\
\quad Linear (Post-hoc) & 0.211 & 0.204 & 0.104 & 0.115 & 0.103 & 0.111 & 0.087 \\
\quad Linear (Ours) & 0.227 & 0.223 & 0.134 & 0.152 & 0.131 & 0.141 & 0.112 \\
\midrule
\multicolumn{8}{l}{\emph{Single Model}} \\
\quad TRIBE v1 \citep{dascoli2026tribev1} & -- & 0.303 & 0.196 & 0.221 & 0.191 & 0.214 & 0.157 \\
\quad TRIBE v2 \citep{dAscoli2026tribev2}  & 0.195 & 0.187 & 0.116 & 0.130 & 0.112 & 0.125 & 0.097 \\
\quad Brain Encoder (Post-hoc)  & 0.292 & 0.282 & 0.174 & 0.192 & 0.171 & 0.193 & 0.141 \\
\quad Brain Encoder (Ours) & 0.301 & 0.291 & 0.195 & 0.212 & 0.194 & 0.213 & 0.162 \\
\quad \ourmodel (Ours) & 0.319 & 0.310 & 0.217 & 0.244 & 0.210 & 0.235 & 0.179 \\
\midrule
\multicolumn{8}{l}{\emph{Ensemble}} \\
\quad MedARC \citep{villanueva2025-algonauts2025_medarc} & -- & 0.288 & 0.209 & 0.230 & 0.200 & 0.230 & 0.174 \\
\quad ModalityRNN \citep{eren2025-algonauts2025_rnn}  & -- & 0.313 & 0.209 & 0.223 & 0.207 & 0.227 & 0.180 \\
\quad TRIBE v1 \citep{dascoli2026tribev1} & -- & 0.320 & 0.215 & 0.238 & 0.210 & 0.238 & 0.172 \\
\quad VIBE \citep{schad2025-algonauts2025_vibe} & -- & 0.320 & 0.210 & 0.235 & 0.205 & 0.227 & 0.172 \\
\quad \textbf{\ourmodel (Ours)} & \textbf{0.335} & \textbf{0.323} & \textbf{0.227} & \textbf{0.253} & \textbf{0.221} & \textbf{0.246} & \textbf{0.189} \\
\bottomrule
\end{tabular}
\end{table}

We evaluate \ourmodel\ against a representative set of baselines on three splits of the CNeuroMod data used in the Algonauts 2025 challenge~\citep{gifford2025algonauts}: the validation set (\emph{Friends} S06), the in-distribution test set (\emph{Friends} S07), and the held-out out-of-distribution set. Figure~\ref{fig:main-result}a organizes the comparison along two axes: architectural complexity (linear ridge baselines $\rightarrow$ frozen-backbone with a learned brain encoder $\rightarrow$ \ourmodel) and fusion strategy (native multimodal fusion vs.\ post-hoc fusion of independently extracted unimodal features). Per-subject and aggregate scores are reported in Table~\ref{tab:results}.

Three observations stand out. 
First, native multimodal fusion outperforms post-hoc (tower) fusion at every architectural level: at the level of a linear ridge, swapping post-hoc unimodal features for native \textsc{Qwen3-Omni} features improves Pearson $r$ from $0.204$ to $0.223$ on \emph{Friends} S07 and from $0.104$ to $0.134$ on \emph{OOD Movies}; the same effect persists when the linear readout is replaced by the brain encoder ($r{=}0.282 \rightarrow 0.291$ on S07 and $r{=}0.174 \to 0.195$ on \emph{OOD Movies}).
 At the level of \ourmodel, the S06 gain of $+0.014$
 holds for every subject and every leave-one-episode-out recomputation (App.~\ref{app:stats-validation}). Against the depth-matched post-hoc (decoder) control, native fusion retains a smaller but consistent advantage of $+0.003$ to $+0.007$ on S06 for every backbone and subject, and of $+0.004$ to $+0.014$ on the held-out splits (App.~\ref{app:fusion-matched-results}, \ref{app:fusion-heldout}).

Second, each step up in architectural complexity yields a consistent gain across all three splits, with \ourmodel\ surpassing every baseline on every split, attaining $r=0.310$ on the in-distribution test set and $r=0.217$ on \emph{OOD Movies}, a relative improvement of roughly $10\%$ on OOD over the strongest single-model post-hoc baseline (TRIBE v1, $r=0.196$). Third, the advantage of \ourmodel\ is most pronounced under distribution shift: 
the linear baseline drops $40\%$ of its test-set Pearson on the OOD split $(0.223\to0.134)$, whereas \ourmodel\ drops only $30\%$ $(0.310\to0.217)$, indicating that the learned encoder captures structure that generalizes beyond the narrative and stylistic statistics of \emph{Friends}. 
Ensembling \ourmodel\ with complementary checkpoints (\S\ref{sec:methods_traning_ensembling}, App.~\ref{app:ensembling}) adds $+0.013$ on S07 and $+0.010$ on OOD, attaining $r=0.323$ and $r=0.227$ and outperforming four published ensembles from the Algonauts 2025 leaderboard, with the largest margin on OOD ($0.227$ vs.\ at most $0.215$).
Subject-averaged Codabench accuracy maps are in App.~\ref{app:codabench-maps}.

We additionally ablate the choice of feature-extractor, keeping the brain encoder and layer gating intact. Across three Qwen-Omni backbones (Figure~\ref{fig:main-result}b; a fourth, \textsc{Qwen2.5-Omni-3B}, in App.~\ref{app:fusion-matched-results}), native fusion outperforms both post-hoc conditions at every scale, and its advantage over both controls is preserved on the held-out S07 and OOD splits (App.~\ref{app:fusion-heldout}); \textsc{Qwen3-Omni-30B-Thinking} achieves the highest alignment ($r=0.319$) and is used as the default backbone throughout.

\subsection{Modality-Specific Contributions Across Cortex}
\label{sec:modality_contrib}

\begin{figure}
    \centering
    \includegraphics[width=1\linewidth]{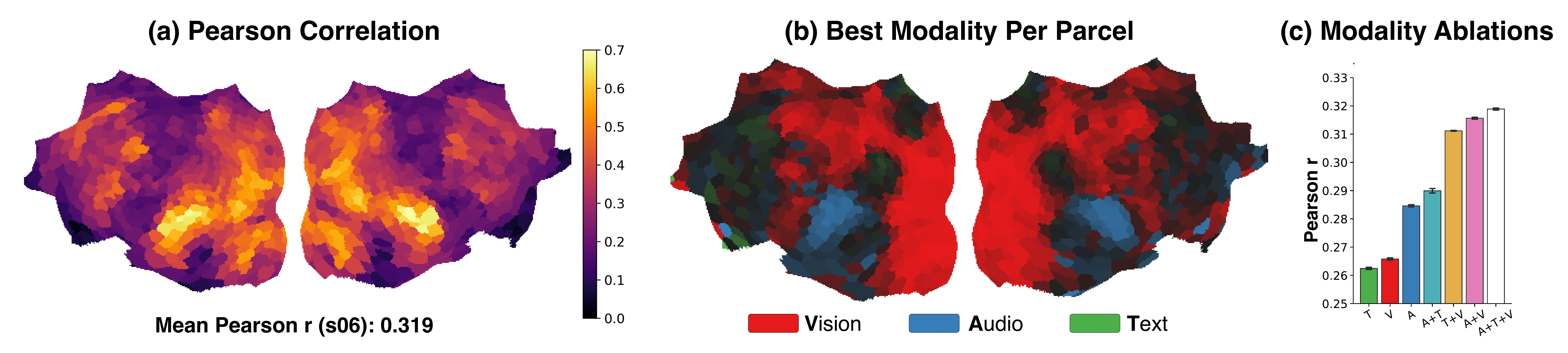}
    \caption{\textbf{Cortical alignment and modality contributions.}
    \textbf{(a)} Per-parcel Pearson $r$ for \ourmodel\ on the validation set, shown on a cortical flatmap.
    \textbf{(b)} Dominant modality per parcel, vision (red), audio (blue), or text (green), defined as the modality whose ablation causes the largest drop in per-parcel Pearson $r$ relative to the full trimodal model. Color saturation encodes dominance strength (the dominant modality's share of the total drop, normalized to $[0,1]$); desaturated parcels reflect distributed multimodal contributions.
    \textbf{(c)} Mean Pearson $r$ across cortex when restricting input to subsets of modalities during training (T = text, V = vision, A = audio); pairwise and trimodal inputs improve over single modalities, indicating complementary contributions.}
    \label{fig:modality_contrib}
\end{figure}

Beyond aggregate accuracy, we ask which cortical regions \ourmodel\ predicts well and which input modalities drive those predictions. Figure~\ref{fig:modality_contrib}a shows the per-parcel Pearson $r$ between predicted and measured BOLD on the validation set, projected onto a cortical flatmap. Alignment is highest in lateral occipitotemporal cortex, superior temporal regions, and lateral-temporal and inferior-frontal areas associated with language processing, with weaker prediction in medial prefrontal and limbic cortex, a pattern broadly consistent with prior encoding studies of naturalistic stimuli~\citep{Huth2016}.

To attribute these predictions to individual modalities, we perform a leave-one-modality-out ablation. For each modality $m \in \{\text{vision}, \text{audio}, \text{text}\}$, we re-evaluate the model with the corresponding input stream zeroed, as during modality dropout, and record the per-parcel drop in Pearson $r$ relative to the full trimodal model. The dominant modality at each parcel is the one whose ablation produces the largest drop; dominance strength is the share of the total drop attributable to it (Figure~\ref{fig:modality_contrib}b). The resulting topography aligns with the canonical functional organization of sensory and language cortex. Vision dominates posterior occipitotemporal cortex (red), audio dominates superior temporal regions around primary and secondary auditory cortex (blue), and text dominates the lateral-temporal and inferior-frontal language network (green; \citep{Fedorenko2024}). Parcels with no single dominant modality (shown desaturated) concentrate in higher-order association areas where multimodal integration is expected, including portions of the temporoparietal junction and lateral prefrontal cortex.

Finally, Figure~\ref{fig:modality_contrib}c quantifies the additive contribution of each modality combination by feeding only specific subsets of modalities into the brain encoder. All three single-modality models achieve relatively high predictions, but each leaves substantial variance unexplained relative to the full model. Pairwise combinations consistently outperform the best single-modality model, and the full trimodal input yields a further gain, indicating that the three modalities contribute complementary rather than redundant information to whole-brain prediction. Shared-scale maps of each modality's unique contribution are given in App.~\ref{app:modality-maps}.

\subsection{Where Does MIRAGE Help, and Which Components Contribute?}
\label{sec:where_what}

\begin{figure}
    \centering
    \includegraphics[width=1\linewidth]{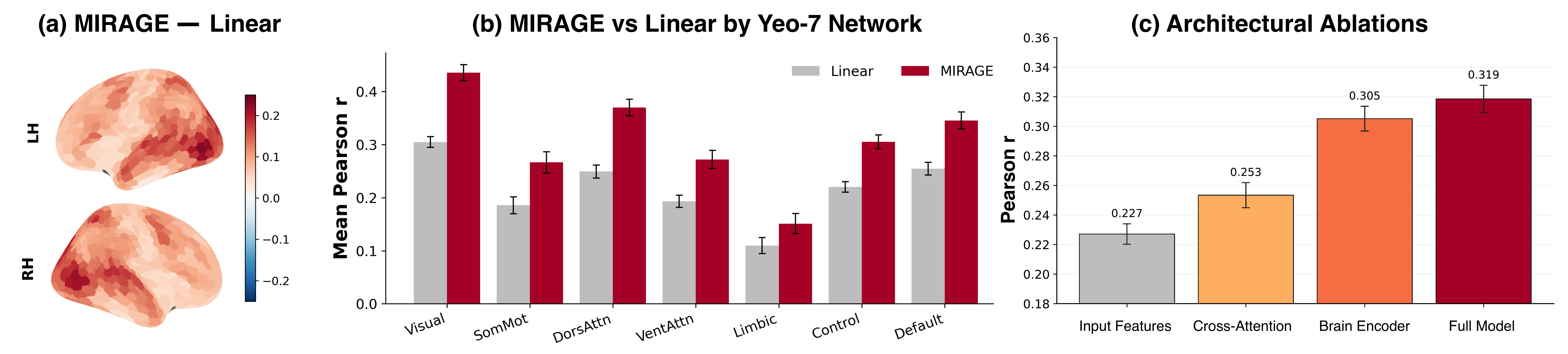}
    \caption{\textbf{Where does \ourmodel\ help, and which components contribute?}
    \textbf{(a)} Parcel-wise difference in Pearson $r$ between \ourmodel\ and the matched \emph{Linear (Native Fusion)} baseline (Fig.~\ref{fig:main-result}), averaged across subjects and projected onto an inflated cortical surface (LH/RH: left/right hemisphere); warmer colors mark parcels where \ourmodel\ improves. Both models share the same input features, so the difference isolates the contribution of the learned encoder.
    \textbf{(b)} Mean Pearson $r$ for \ourmodel\ (red) and the linear baseline (gray) within each of the seven canonical Yeo--Krienen networks~\citep{ThomasYeo2011}: Visual, Somatomotor (SomMot), Dorsal/Ventral Attention (DorsAttn/VentAttn), Limbic, Frontoparietal Control (Control), and Default Mode (Default).
    \textbf{(c)} Pearson $r$ from a per-subject linear probe trained on representations at successive stages of \ourmodel: raw input features, post cross-attention, post \emph{Brain Encoder}, and full model output (no additional fitting).
    Error bars in (b) and (c) denote SEM across the four CNeuroMod subjects.
    }
    \label{fig:ablation}
\end{figure}

Having established that \ourmodel\ achieves state-of-the-art alignment, we now ask \emph{where} on cortex its gains concentrate and \emph{which} architectural components are responsible. To isolate the contribution of the learned encoder from that of the backbone, we compare \ourmodel\ against the matched linear-readout baseline \emph{Linear (Native Fusion)}, which uses the same \textsc{Qwen3-Omni} features but a per-subject linear ridge in place of \ourmodel's encoder. Any difference between the two is attributable to the encoder rather than to the input features.

\paragraph{Cortical distribution of gains.} 
Figure~\ref{fig:ablation}a shows the parcel-wise difference in Pearson $r$ between \ourmodel\ and the linear baseline. Improvements are positive across nearly all of cortex, with the largest gains in lateral occipitotemporal and dorsal frontoparietal regions and the smallest gains in primary sensorimotor and limbic cortex. Aggregating into the seven canonical Yeo--Krienen networks~\citep{ThomasYeo2011} confirms this picture (Figure~\ref{fig:ablation}b): \ourmodel\ outperforms the linear baseline in every network, with the largest absolute gains in the Visual ($\Delta r \approx 0.13$) and Dorsal Attention ($\Delta r \approx 0.12$) networks and the smallest in the Limbic network ($\Delta r \approx 0.04$). 
These peaks overlap with regions specialized for dynamic social-visual processing (LOC, pSTS, EBA, MT+) and the top-down attention network, both heavily recruited by the social-narrative content of \emph{Friends}. Against our TRIBE~v1-style reproduction, a strong non-linear baseline, \ourmodel\ improves 968 of 1{,}000 parcels, with the largest gains in Control and Default cortex (App.~\ref{app:mirage-vs-tribe}).

\paragraph{Architectural attribution.} 
To localize this gain within \ourmodel's architecture, we train a per-subject linear probe at successive stages of \ourmodel: Figure~\ref{fig:ablation}c shows that each component contributes a measurable gain in mean Pearson r ($0.227 \rightarrow 0.253 \rightarrow 0.305 \rightarrow 0.319$). The Brain Encoder produces the largest single jump ($+0.052$); the cross-attention pooler and subject-specific head contribute smaller but consistent gains.
 Replacing the pooler by a fixed read of the most-attended layer of each modality costs $0.016$ on S06, a gap that persists on the held-out splits. The same pooler also improves a model built on TRIBE-style unimodal features, so the aggregation benefit is not specific to Qwen-Omni (App.~\ref{app:single-layer}, \ref{app:tribe-cross-attention}).

\subsection{Adaptive Gating Reveals Modality-Specific Layer Preferences}
\label{sec:layer_preferences}

\begin{figure}[t]
    \centering
    \includegraphics[width=1\linewidth]{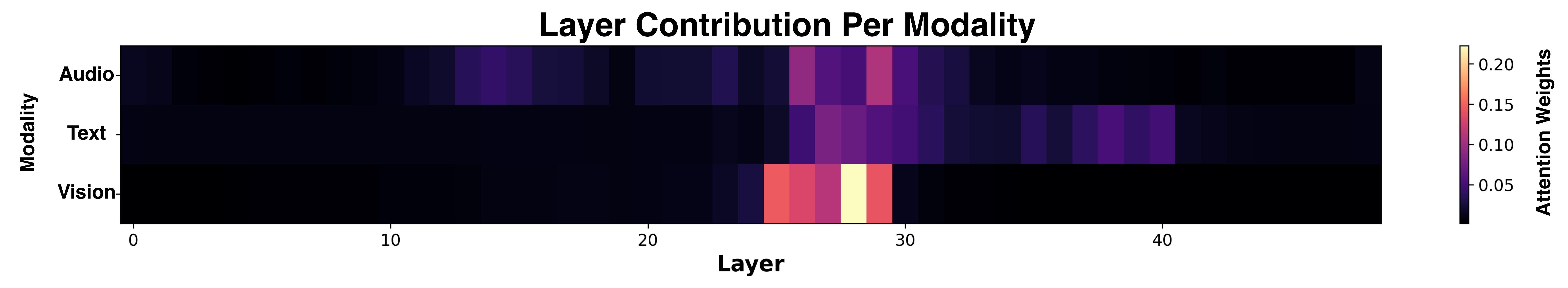}
    \caption{\textbf{Layer-wise contributions of \textsc{Qwen3-Omni} features to each modality.} Cross-attention weights from \ourmodel's per-modality cross-attention modules (vision, text, audio) over the 48 layers of the \textsc{Qwen3-Omni} language module, averaged across attention heads and the 24 latent queries; brighter cells indicate layers that contribute more strongly to the modality-specific readout. Per-head breakdowns are in Appendix~\ref{app:attention_heads}.}
    \label{fig:layer_attention}
\end{figure}

A practical advantage of \ourmodel's per-modality cross-attention design is that the learned attention weights expose, in a directly inspectable form, which layers of \textsc{Qwen3-Omni} contribute to each modality's readout. Figure~\ref{fig:layer_attention} reports the cross-attention weights of each modality's gating module, averaged across attention heads and across the 24 latent queries.

The three modalities exhibit distinct depth profiles. Vision is the most layer-selective: attention concentrates sharply around layers 25--30 and is near zero elsewhere. Text distributes its weight more broadly across mid-to-late layers, with a primary cluster overlapping the vision band and a secondary cluster around layers 35--40. Audio is the most diffuse, spreading weight across a wide range of mid-depth layers rather than committing to a single band. Across all three modalities, the early layers ($0$--$10$) carry little weight.

\subsection{Data Scaling, Subject Adaptation, and Dataset Transfer}
\label{sec:generalization}

The results above are obtained on four CNeuroMod subjects. We therefore ask how much neural data the model needs, whether its representation transfers to a new participant, and whether it transfers to other datasets (Figure~\ref{fig:transfer}; App.~\ref{app:data-scaling}, \ref{app:cross-subject}, \ref{app:external-transfer}). Trained on nested subsets of the 282 training videos, \ourmodel\ reaches $70\%$ of its full-data S06 score with $5\%$ of the videos and $94.7\%$ with half of them, with monotonic curves for every subject (three seeds per fraction). Holding one subject out, a model trained on the other three predicts the held-out subject at $r = 0.250$ without any fMRI from that subject, using the average of the three training subjects' heads. Fitting only a linear readout on $5\%$ of that subject's videos reaches $0.297$, and on half of them $0.316$, about $0.002$ below the model trained on all four subjects; transfer exceeds training from scratch at every data fraction and for every held-out subject (Figure~\ref{fig:transfer}a).

Finally, the frozen representation transfers to two external movie-watching datasets acquired with different scanners, participants, and stimuli. On HCP 7T Movie ($n = 20$) \ourmodel\ reaches $r = 0.183$ zero-shot and $0.252$ after fitting a participant-specific ridge readout, and on CineBrain ($n = 6$) $0.047$ and $0.109$ (Figure~\ref{fig:transfer}b). These datasets also let us repeat the native-versus-post-hoc (decoder) comparison with participants as the unit. After readout fitting, native fusion exceeds the depth-matched control for all three backbones on HCP (by $+0.006$ to $+0.015$) and in the pooled sample of 26 participants, whereas CineBrain alone is directionally consistent but underpowered with six participants (Figure~\ref{fig:transfer}c).

%% file: sections/05-discussion.tex
The empirical results across \S\ref{sec:sota}--\S\ref{sec:layer_preferences} support a coherent picture: \emph{where} multimodal integration matters and learnable layer \emph{fusion} over a natively omnimodal backbone outperforms independently extracted unimodal streams.
Native fusion in \modelname{Qwen3-Omni-30B-A3B-Thinking} beats post-hoc fusion at every architectural level we tested (linear ridge, learned brain encoder, and \ourmodel). Against the depth- and representation-matched control, the benefit of joint conditioning is small ($+0.003$ to $+0.007$ on S06, $+0.006$ to $+0.015$ on HCP after readout fitting) but consistent across backbones and subjects, and for most external participants. Within the Qwen-Omni family, the cross-modal interactions learned during foundation-model training therefore seem to provide generalizable features for whole-brain prediction, and should be favored over a downstream combination of unimodal streams.

\begin{figure}[t]
    \centering
    \includegraphics[width=\linewidth,trim=124 24 41 35,clip]{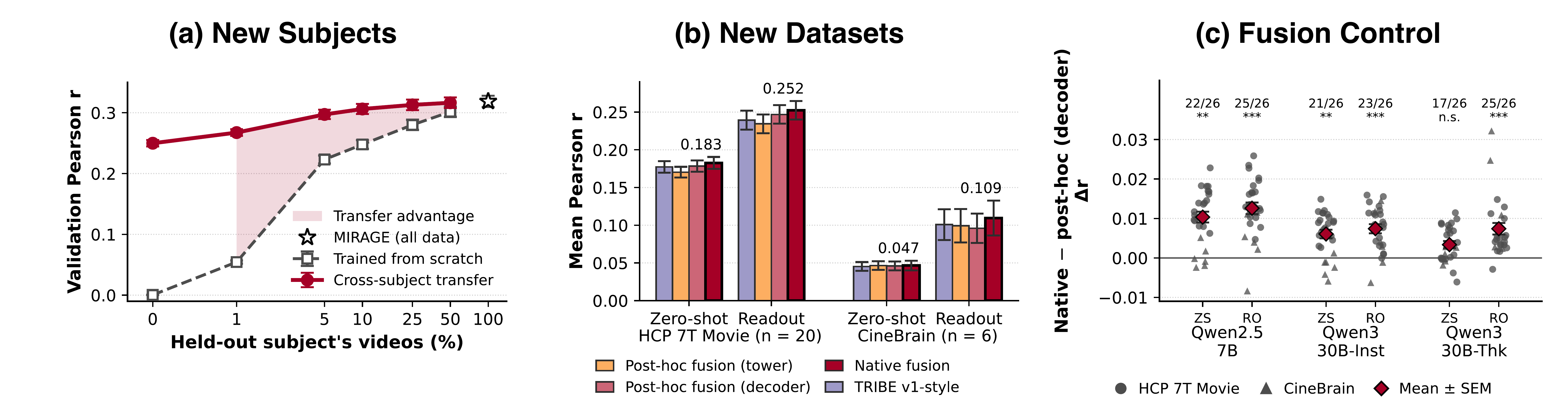}
    \caption{\textbf{Subject adaptation and dataset transfer.}
    \textbf{(a)} Leave-one-subject-out transfer on \emph{Friends} S06: \ourmodel\ trained on three CNeuroMod subjects is frozen and only the held-out subject's linear readout is fitted on the stated fraction of that subject's training videos (red; mean $\pm$ SEM over the four held-out subjects after seed averaging), compared with training from scratch on the same data (gray; shaded: transfer advantage); the star is \ourmodel\ trained on all four subjects.
    \textbf{(b)} Frozen CNeuroMod-trained models on HCP 7T Movie ($n = 20$) and CineBrain ($n = 6$), zero-shot (ZS) and with a participant-specific ridge readout (RO), for the three fusion conditions of \textsc{Qwen3-Omni-30B-Thinking} and our TRIBE~v1-style reproduction (mean $\pm$ SEM over participants; \ourmodel's values printed).
    \textbf{(c)} Native minus post-hoc fusion (decoder) per participant, HCP (circles) and CineBrain (triangles) pooled ($n = 26$); diamonds are means $\pm$ SEM, counts are participants with a positive difference, and stars are Holm-corrected exact sign tests ($^{*}$ $p \le 0.05$, $^{**}$ $p \le 0.01$, $^{***}$ $p \le 0.001$).}
    \label{fig:transfer}
\end{figure}

\paragraph{Modality-specific layer preferences.}
The cross-attention attribution in \S\ref{sec:layer_preferences} shows that vision collapses onto a narrow mid-depth band where visual semantics appear to consolidate, and that text spreads over mid-to-late layers with a second cluster deeper in the language module, consistent with multiple stages of linguistic abstraction \citep{skean2025layer}. Audio integrates diffusely across mid-depth layers, in line with the longer temporal context that speech requires. For every stream the embedding-adjacent layers carry little weight, so brain-relevant representations emerge several blocks into the language module (App.~\ref{app:attention_heads}).

\paragraph{Where the gains come from.}
The component-wise probe in \S\ref{sec:where_what} attributes the largest single gain over a matched linear readout to the temporal brain encoder, indicating that whole-brain prediction benefits from explicitly modeling temporal structure beyond what a ridge regression captures, with smaller but consistent contributions from the cross-attention layer poolers and the subject-specific head. Spatially, the gains concentrate in the Visual and Dorsal Attention networks, the cortical territory most strongly engaged by naturalistic audiovisual stimuli.

\paragraph{Limitations.}
Our models are trained on the four-subject CNeuroMod cohort \citep{boyle2020courtois_cneuromod} as exposed by Algonauts 2025 \citep{gifford2025algonauts}.
The transfer experiments (\S\ref{sec:generalization}) show that the shared trunk generalizes to held-out subjects and to 26 external participants, but all training data come from four individuals, and the external evaluations rest on four stimuli per dataset and one trained checkpoint per configuration (App.~\ref{app:external-protocol}).
The native-vs.-post-hoc result is established within the Qwen-Omni family; whether the same effect transfers to other foundation models, and how it scales beyond 3B active parameters, remains to be tested.  
One practical cost of our design is that the layer pooler operates over the full backbone layer stack, so cached features must retain every layer, which raises the feature footprint and dataloader cost relative to pipelines that pre-aggregate layers at extraction time. Finally, the attention attributions are model-level: they show what the layer poolers prefer to read out, not what individual cortical parcels prefer; parcel-level attribution along the layer axis is left to future work.

\paragraph{Broader impact.}
Brain encoding models trained on naturalistic fMRI advance basic neuroscience and may eventually inform brain--computer interfaces and clinical applications. The same models also raise privacy considerations: predictive mappings from sensory inputs to brain responses can in principle support inferences about mental states from neural recordings. Our experiments use the publicly released, consent-based CNeuroMod corpus and predict responses to externally presented stimuli; we do not perform mental-state decoding. We see no immediate dual-use risk specific to this work, but we recommend that any deployment outside research contexts includes explicit consent, data minimization, and audit trails.

\paragraph{Future Work.}

Two extensions follow naturally. Richer subject models than a linear head (shared low-rank subject embeddings, hypernetwork-conditioned heads, or test-time adaptation) could improve on the transfer results of \S\ref{sec:generalization} for participants with little or no fMRI. On the brain side, the framework is modality-agnostic: an EEG, MEG, or ECoG readout on the same backbone would let the depth profiles favored by each measurement reveal the spatial and temporal scales at which neural signals align with foundation-model representations.

%% file: sections/02-related_work.tex
\paragraph{Brain encoding with task-optimized neural networks.}
Linearly mapping representations of pretrained deep networks to neural activity has become the dominant paradigm for predicting brain responses to naturalistic stimuli, from early demonstrations in ventral stream regions~\citep{Yamins_Hong_Cadieu_Solomon_Seibert_DiCarlo_2014,Khaligh-Razavi2014} to large-scale benchmarks such as Brain-Score~\citep{SchrimpfKubilius2018BrainScore,Schrimpf2021Language,Schrimpf2020integrative} and the Algonauts challenges~\citep{gifford2025algonauts,cichy2019algonauts,cichy2021algonauts,gifford2023algonauts}. Subsequent work has shown that the choice of backbone, training objective, and inductive bias substantially shapes brain predictivity, with self-supervised, language-aligned, and contrastive models often matching or surpassing supervised ones~\citep{Schrimpf2021Language,conwell2024bias,gokce2025scaling,
tang2025manytwoone, alkhamissi2025language2cognition,zhuang2021unsupervised,muttenthaler2023human}. These studies typically fit \emph{unimodal} encoders separately per subject and per measurement, leaving open how to integrate information across modalities and representational depths in a unified model.
A complementary line of work aligns responses across individuals before fitting, through hyperalignment~\citep{haxby2011common,guntupalli2016humancortex}, shared response models~\citep{chen2015reduced_response}, optimal transport~\citep{thual2022aligning}, or shared-subject latent spaces~\citep{scotti2024mindeye2}. We instead share one trunk across subjects with a thin per-subject readout and test its transfer to held-out participants directly (\S\ref{sec:generalization}).

\paragraph{Multimodal encoders of naturalistic stimuli.}
The release of multimodal foundation models~\citep{radford2021clipopenai,assran2025vjepa2,xu2025qwen25omnitechnicalreport,xu2025qwen3omnitechnicalreport,Qwen3-VL} and rich audiovisual neural datasets~\citep{boyle2020courtois_cneuromod,allen2021nsd} has spurred a shift toward multimodal encoding. \citet{tang2023brain} showed that features from multimodal transformers transfer across visual and language regions, and \citet{oota2025multimodal} found that multimodal stimuli are best explained by multimodal models. Building on the Courtois NeuroMod data~\citep{boyle2020courtois_cneuromod}, the Algonauts 2025 challenge~\citep{gifford2025algonauts} produced a wave of whole-brain encoders that combine video, audio, and language backbones in different ways: TRIBE~\citep{dascoli2026tribev1} and its foundation-model extension~\citep{dAscoli2026tribev2} train a trimodal transformer to predict fMRI across cortex, while VIBE~\citep{schad2025-algonauts2025_vibe}, the multimodal-LLM probe of \citet{villanueva2025-algonauts2025_medarc}, the seq2seq transformer of \citet{he2025-algonauts2025_seq2seq}, and the recurrent ensemble of \citet{eren2025-algonauts2025_rnn} explore complementary architectural choices. Our framework shares this multimodal premise but isolates the contribution of \emph{fused} versus modality-specific representations from the same backbone, and replaces fixed layer pooling with learned layer aggregation.

\paragraph{Layer selection, aggregation, and readout design.}
Which layers to read out from is a long-standing question in model–brain comparisons. Early studies pick the single best-matching layer per region~\citep{Yamins_Hong_Cadieu_Solomon_Seibert_DiCarlo_2014,SchrimpfKubilius2018BrainScore}, while more recent work argues that readout design should be treated as a first-class modeling choice that reflects the underlying scientific question~\citep{
gokce2026multimodal_brain_scaling,
ivanova2022readout_mapping, alkhamissi2025language2cognition}. Transformer-based readouts that aggregate features across layers improve fits in high-level visual cortex~\citep{adeli2025nsd_transformer_scale,hwang2025nsd_transformer_gen,saha2025readout}, and response-optimized models recover non-strictly hierarchical structure in visual areas~\citep{styves2023brainoptimized_nature,khosla2022responseoptimized_neurips}. Closest to our setting, brain-tuning and alignment objectives have been used to adapt speech and vision models to neural data~\citep{moussa2025neuralfitting_speech,lu2024realnet,dapello2023aligning}. Our layer poolers are most closely related to Perceiver-style cross-attention~\citep{jaegle2021perceiver, jaegle2022perceiverio}, but are applied along the \emph{layer} axis of frozen extractors, with a small number of learned queries that summarize an entire layer stack into one or a few tokens per timestep.%

%% file: sections/06-conclusion.tex
We presented \ourmodel, a brain encoding framework that pairs a multimodal foundation model with adaptive, per-modality cross-attention over the backbone's layers.
Within the Qwen-Omni family, native multimodal fusion outperforms post-hoc fusion at every architectural level and backbone scale we evaluated, with a small but consistent benefit that survives a depth- and representation-matched control and persists for most external participants.
Adaptive layer-wise aggregation through latent queries adds further gains and keeps the model interpretable, exposing modality-specific depth profiles in the backbone and spatial patterns across cortex. Backbones that integrate modalities during pretraining offer a more generalizable and accurate alternative for whole-brain fMRI encoding that transfers to new participants and datasets.

%% file: appendices/00-appendix.tex
\appendix
\supplementarysection

\input{appendices/A_implementation}
\clearpage
\input{appendices/B_statistics}
\clearpage
\input{appendices/C_fusion_controls}
\clearpage
\input{appendices/D_layer_aggregation}
\clearpage
\input{appendices/E_encoder_ablations}
\clearpage
\input{appendices/F_transfer}
\clearpage
\input{appendices/G_cortical_maps}
\clearpage
\input{appendices/H_licenses}

%% file: appendices/A_implementation.tex
\section{Implementation, Data Alignment, Compute, and Provenance}
\label{app:cr-implementation}
\label{app:training-implementation}

This section gives the details needed to reproduce \ourmodel: data splits, feature extraction, stimulus--BOLD alignment, architecture, training, and ensembling. It also reports parameter counts, compute, and the source of every score in Table~\ref{tab:results}.
\subsection{Objective and Data Splits}
\label{app:training-setup}

\paragraph{Objective and model selection.}
Models are trained with mean-squared error between predicted and measured BOLD responses. Checkpoints are selected by the mean validation Pearson correlation, computed independently for each cortical parcel and then averaged across parcels and subjects.

\paragraph{Data splits.}
We hold out \emph{Friends} season~6 as the in-distribution validation set. The remaining released training stimuli, including the other \emph{Friends} seasons and \emph{Movie10}, are used for fitting. Final in-distribution and out-of-distribution scores are obtained from the Algonauts 2025 evaluation platform on \emph{Friends} season~7 and the held-out movie benchmark, respectively.

\subsection{Feature Extraction}
\label{app:feature-extraction}

All backbones are evaluated on a common $2$~Hz stimulus grid; features are produced in float16 and cached with the full backbone layer axis intact, regardless of the backbone family or fusion variant. Caching is performed once and the backbone is frozen thereafter; only brain-model components (modality-specific layer aggregator, linear projections, shared temporal encoder, subject-specific readout) are optimized. Any depth reduction over backbone layers is a training-time choice (see Section~\ref{app:architecture}), not a property of the cached features.

\paragraph{Unimodal baselines (TRIBE-style).}
The unimodal feature pipeline closely follows TRIBE~\citep{dascoli2026tribev1}. Language features come from \modelname{Llama-3.2-3B}~\citep{grattafiori2024llama3herdmodels} run on chunked transcripts with a $1024$-token context window and projected onto the 2~Hz grid by temporal overlap. Auditory features come from \modelname{Wav2Vec-Bert-2.0}~\citep{chung2021wav2vecbert} extracted on $60$~s audio windows and interpolated to the 2~Hz grid. Visual features come from \modelname{V-JEPA~2}~\citep{assran2025vjepa2} extracted from $64$ frames per $4$~s clip with spatial averaging over patch tokens. Each backbone is run independently and produces a single modality stream.

\paragraph{Qwen-Omni backbones.}
We extract features from four Qwen-Omni backbones: \modelname{Qwen2.5-Omni-3B}~\citep{xu2025qwen25omnitechnicalreport} (used only for the backbone-scale control of App.~\ref{app:fusion-matched-results}), \modelname{Qwen2.5-Omni-7B}~\citep{xu2025qwen25omnitechnicalreport}, \modelname{Qwen3-Omni-30B-A3B-Instruct}~\citep{xu2025qwen3omnitechnicalreport}, and \modelname{Qwen3-Omni-30B-A3B-Thinking}~\citep{xu2025qwen3omnitechnicalreport}. The final \ourmodel\ uses \modelname{Qwen3-Omni-30B-A3B-Thinking}. For each backbone, every input window contains up to $1024$ transcript words, $60$~s of audio, and $4$~s of video at no more than $8$ sampled frames.

\paragraph{Post-hoc fusion (tower) extraction.}
In this variant, each modality is read out from the corresponding Qwen sub-network in isolation: audio features from the audio tower alone, visual features from the vision tower alone, and language features from the language block applied to transcript-only inputs. No cross-modal context is constructed and no modality conditions any other.

\paragraph{Post-hoc fusion (decoder) extraction.}
Each modality is passed alone through the full backbone and read from the same language-module layers as in the native condition; the three streams are concatenated afterwards. This is the depth-matched control defined in App.~\ref{app:fusion-conditions}.

\paragraph{Native fusion extraction.}
In the fused variant, we invoke each Qwen-Omni's internal context construction so that vision tokens and audio tokens are interleaved with text tokens inside the multimodal mixer. We retain three post-fusion streams (language, auditory, and visual), each conditioned on the full audiovisual stimulus and the aligned transcript.

\subsection{Stimulus--BOLD Alignment}
\label{app:temporal-alignment}

Let $\delta = 1.49$\,s be the fMRI repetition time and $\Delta = 3\delta = 4.47$\,s the fixed hemodynamic offset. Each training example contains $N_y = 100$ BOLD samples, so one model window spans
\begin{equation*}
    W = N_y\,\delta = 149\ \text{s},
\end{equation*}
represented by $N_x = \mathrm{round}(W f_x) = 298$ stimulus-feature tokens on the common $f_x = 2$\,Hz grid. Windows tile each run with stride $W$. For window $k$, the stimulus-time start is $s_k = -\Delta + kW$; feature token $v \in \{0,\dots,297\}$ carries nominal time $t^x_{k,v} = s_k + v/f_x$, and BOLD target $u \in \{0,\dots,99\}$ is acquired at $t^y_{k,u} = kW + u\delta$. The stimulus time paired with a BOLD target after the hemodynamic shift is $\tau_{k,u} = t^y_{k,u} - \Delta = s_k + u\delta$, i.e.\ the first window begins at $-4.47$\,s in stimulus time and is paired with BOLD from fMRI time $0$; positions before stimulus onset are zero-padded. \ourmodel\ does not concatenate a bank of feature lags or convolve features with a canonical hemodynamic response function; it applies this fixed three-TR shift and lets the temporal encoder learn the remaining temporal mapping.

After layer aggregation, projection, and fusion, the fused sequence $Z_k \in \mathbb{R}^{298 \times 3072}$ is processed by the temporal encoder and the subject head to give parcel predictions $R_k \in \mathbb{R}^{298 \times 1000}$ at the feature rate. Adaptive average pooling then maps the 298 temporal predictions onto the 100 BOLD targets,
\begin{equation*}
    \widehat y_{k,u,p} = \frac{1}{|I_u|}\sum_{v \in I_u} R_{k,v,p},
    \qquad
    I_u = \Big\{\big\lfloor \tfrac{u N_x}{N_y} \big\rfloor,\;\dots,\;\big\lceil \tfrac{(u+1) N_x}{N_y} \big\rceil - 1\Big\},
\end{equation*}
so each target averages three or four feature-rate outputs, with bins that may overlap at their boundaries.

\paragraph{Causality of the cached features.} Backbone features are computed from raw inputs that are truncated before each forward pass: at every 2\,Hz cutoff $c_i = (i+1)/2$\,s the extractor builds a transcript prefix, audio clip, and video clip ending at $c_i$ and runs a new backbone forward. The trailing context is at most 1{,}024 words whose end time is $\le c_i$ for text, $[\max(0, c_i - 60), c_i]$ for audio, and $[\max(0, c_i - 4), c_i]$ at 2\,fps (at most eight frames) for vision. No stimulus content later than the cutoff is available to any stored feature; this is why caching costs about 700 GPU-hours per backbone (App.~\ref{app:compute_resources}), since one full-episode forward followed by timestamp slicing would be far cheaper but would introduce feature-level look-ahead.

\paragraph{Causal feature extraction and temporal masking.} The causal temporal-encoder control (App.~\ref{app:causal-encoder}) combines these cutoff-truncated features with a causal attention mask. After temporal readout pooling, the latest raw-input cutoff contributing to BOLD target $u$ is $1.49$--$1.98$\,s after its nominal stimulus time $\tau_{k,u}$. With the $4.47$\,s hemodynamic offset, this cutoff precedes the corresponding BOLD acquisition by $2.49$--$2.98$\,s. Causality with respect to acquisition therefore follows from the combination of truncated feature extraction and causal temporal masking. The default bidirectional encoder can additionally use later feature tokens within the same window.

\subsection{Architecture}
\label{app:architecture}

\paragraph{Per-modality layer aggregation.}
The cached layer stack is reduced to a per-modality embedding at training time. \ourmodel\ uses a one-layer cross-attention pooler per modality with $4$ heads, $24$ learned queries, and attention dropout $0.2$; the query outputs are concatenated along the hidden dimension. Baselines that do not use cross-attention apply a fixed reduction instead: layers with relative depth in $[0.5,\,0.75]$ are mean-pooled into one embedding and layers in $[0.75,\,1.0]$ into a second, and the two embeddings are concatenated. Either choice operates on the same cached features.

\paragraph{Fusion, encoder, and readout.}
The per-modality embeddings are linearly projected to a shared hidden dimension and concatenated across modalities. The resulting sequence is processed by an $8$-layer transformer encoder with hidden dimension $3072$, $8$ attention heads, and feed-forward expansion $4\times$; inner dropouts in the transformer are $0$. During training, modality dropout with probability $0.3$ randomly removes input streams while ensuring at least one stream remains active. Subject identity enters only through the readout, a per-subject linear projection over parcels. After the readout, predictions are reduced along the temporal axis by adaptive average pooling to $K = 100$ output time steps.

\subsection{Training Procedure}
\label{app:training-procedure}

\paragraph{Optimization.}
We train with AdamW using a peak learning rate of $10^{-4}$ and decoupled weight decay $10^{-2}$. The learning rate follows a one-cycle cosine schedule with a $10\%$ warm-up phase. Gradients are clipped to an $\ell_2$ norm of $1.0$, and training uses 16-bit mixed precision on a single accelerator. Unless stated otherwise, all final runs use random seed $33$.

\paragraph{Schedule and compute.}
\label{app:compute_resources}
Each model is trained for $15$ epochs with batch size $16$ and $32$ data-loader workers. Training examples target $K=100$ fMRI time points over $P=1000$ cortical parcels for $N_S=4$ subjects. Experiments use Python~3.12 and PyTorch~2.7 on academic clusters, with either a single NVIDIA A100 80GB GPU or a single NVIDIA Grace Hopper GH200 module, $16$ CPU cores, and $256$~GB of RAM per job. A 15-epoch run takes approximately $4$ hours; extracting the complete cached feature set across modalities for one Qwen-Omni model on \emph{Friends} seasons~1--6 and \emph{Movie10} takes approximately $700$ GPU-hours.

\paragraph{Variants we explored.}
Several variants did not improve validation performance and were discarded. Composite objectives that augment MSE with a Pearson-correlation term, centered kernel alignment (CKA), or representational similarity analysis (RSA) did not outperform MSE alone. Training for more epochs did not yield further gains. Larger learning rates frequently produced numerical instabilities (including NaN losses) even with gradient clipping enabled, while AdamW with weight decay $10^{-2}$ and clipping at $1.0$ provided a small but consistent regularization benefit. Increasing the number of cross-attention heads or raising dropout rates inside the encoder did not improve the final model.

\subsection{Linear Baselines and Readouts}
\label{app:linear_baselines}

To isolate the contributions of native fusion over post-hoc fusion and of cross-attention aggregation over fixed pooling, we complement \ourmodel\ with a strong linear encoding baseline. For raw cached features with a layer axis, we first mean-pool across layers so that each modality contributes a single embedding stream; we omit this step when the baseline is fit on the output of a trained cross-attention layer pooler or brain encoder, where the layer axis has already been collapsed. Streams are aligned to the fMRI TR grid by mean pooling on the common 2~Hz feature grid and concatenated across modalities. We then build a per-subject design matrix in which each row corresponds to one TR and concatenates the feature vectors at lags $\{-4,\,-3,\,-2,\,-1,\,0\}$ TR; the resulting design is reduced from its native dimensionality to $1024$ by a sparse random projection. For each subject we fit a multi-output ridge regression onto the $P=1000$ parcels with leave-one-out cross-validation, selecting the regularizer independently for each parcel from a grid of $99$ values log-spaced between $10^{-2}$ and $10^{7}$. Models are fit on the training split and scored on \emph{Friends} season~6 by per-parcel Pearson correlation. %

\subsection{Hyper-parameter Summary}
\label{app:training-hparam-summary}

Table~\ref{tab:training-hparams} consolidates the architecture, optimization, training, data, compute, and ensembling hyper-parameters of the final \ourmodel.

\begin{table}[t]
\centering
\caption{Hyper-parameters of the final \ourmodel\ model.}
\label{tab:training-hparams}
\small
\setlength{\tabcolsep}{3pt}
\renewcommand{\arraystretch}{1.10}
\begin{tabular}{@{}l l >{\raggedright\arraybackslash}p{0.49\linewidth}@{}}
\toprule
\textbf{Block} & \textbf{Hyper-parameter} & \textbf{Value} \\
\midrule
\multirow{4}{*}{Backbone}
  & Model                    & \modelname{Qwen3-Omni-30B-A3B-Thinking} (frozen) \\
  & Streams                  & language / audio / vision (post-fusion) \\
  & Feature dtype / grid     & float16 / $2$~Hz \\
  & Layer access             & full layer stack, no depth pooling before training \\
\cmidrule(l){1-3}
\multirow{3}{*}{Window per example}
  & Transcript words         & $\le 1024$ \\
  & Audio duration           & $60$~s \\
  & Vision duration / frames & $4$~s / $\le 8$ \\
\midrule
\multirow{5}{*}{Layer pooler (per modality)}
  & Kind                     & layer cross-attention \\
  & Depth                    & $1$ \\
  & Heads                    & $4$ \\
  & Queries $n_q$            & $24$ (outputs concatenated) \\
  & Attention dropout        & $0.2$ \\
\cmidrule(l){1-3}
\multirow{6}{*}{Brain encoder}
  & Hidden dimension $D$     & $3072$ \\
  & Depth                    & $8$ \\
  & Heads                    & $8$ \\
  & FF expansion             & $4\times$ \\
  & Inner dropouts           & $0$ \\
  & Modality dropout         & $0.3$ \\
\cmidrule(l){1-3}
\multirow{4}{*}{Readout}
  & Subject embedding        & none (in trunk) \\
  & Head                     & per-subject linear over parcels \\
  & Temporal reducer         & adaptive avg.\ pool, post-head \\
  & Output time steps $K$    & $100$ \\
\midrule
\multirow{6}{*}{Optimization}
  & Loss                     & MSE \\
  & Selection metric         & val Pearson (max) \\
  & Optimizer                & AdamW (decoupled, fused off) \\
  & Peak LR / weight decay   & $10^{-4}$ / $10^{-2}$ \\
  & LR schedule              & OneCycleLR, $10\%$ warm-up, cosine anneal \\
  & Grad-clip / SWA          & $\ell_2 = 1.0$ / disabled \\
\cmidrule(l){1-3}
\multirow{4}{*}{Training}
  & Epochs / batch size      & $15$ / $16$ \\
  & Precision                & 16-bit mixed \\
  & Random seed              & $33$ \\
  & Data-loader workers      & $32$ \\
\midrule
\multirow{3}{*}{Data and target}
  & Validation split         & \emph{Friends} season~6 \\
  & Test splits              & \emph{Friends} season~7 (in-dist.); OOD movies \\
  & Target shape             & $K=100$ time steps, $P=1000$ parcels, $N_S=4$ subjects \\
\midrule
\multirow{5}{*}{Compute}
  & Software                 & Python~3.12, PyTorch~2.7 \\
  & GPU                      & 1$\times$ A100 80GB \emph{or} 1$\times$ GH200 \\
  & CPU / RAM                & $16$ cores / $256$~GB per job \\
  & Training wall-clock      & $\approx 4$~h per $15$-epoch run \\
  & Feature extraction       & $\approx 700$ GPU-h per Qwen-Omni model (all modalities) \\
\midrule
\multirow{2}{*}{Ensembling}
  & Members                  & $15$ \\
  & Weighting                & per-(subject, parcel) softmax over val Pearson, $\tau = 0.3$ \\
\bottomrule
\end{tabular}
\end{table}

\subsection{Parameter Accounting}
\label{app:parameters-compute}
\label{app:parameter-counts}

Comparing encoding models by a single parameter count is misleading because three quantities are routinely conflated: the size of the frozen pretrained feature extractors, the number of parameters trained on the fMRI objective in one checkpoint, and the ensemble multiplicity with which checkpoints are combined. Table~\ref{tab:parameter-counts} separates them for \ourmodel\ and for the Algonauts 2025 systems in Table~\ref{tab:results}. Frozen counts are not multiplied by the number of encoder seeds when all seeds reuse the same cached features; for mixture-of-experts backbones we report both total and per-token active parameters.

\begin{table}[h]
\centering
\caption{\textbf{Frozen, trainable, and ensemble parameters for \ourmodel\ and the compared systems.} Trainable counts are per encoding checkpoint. Counts marked $\dagger$ are author-reported; others are derived from the released architectures and configurations (rounded counts for some frozen backbones).}
\label{tab:parameter-counts}
\small
\renewcommand{\arraystretch}{1.2}
\setlength{\tabcolsep}{4pt}
\begin{tabular}{@{}>{\raggedright\arraybackslash}p{0.21\linewidth} >{\raggedright\arraybackslash}p{0.33\linewidth} >{\raggedright\arraybackslash}p{0.2\linewidth} >{\raggedright\arraybackslash}p{0.18\linewidth}@{}}
\toprule
\textbf{Method} & \textbf{Frozen feature extractors} & \textbf{Trainable per checkpoint} & \textbf{Models combined} \\
\midrule
\ourmodel                       & \modelname{Qwen3-Omni-30B-A3B}: $\approx 30$B total, $\approx 3$B active/token & $1.123$B & 1 (single); 15 (ensemble) \\
Brain encoder, fixed pooling (ours) & same cached features & $0.934$B & 1 \\
Linear ridge (ours)             & same cached features & $4.10$M coefficients & 4 subject fits \\
TRIBE v1                        & $4.3$B$^{\dagger}$ nominal ($\approx 4.85$B from checkpoints) & $0.980$B$^{\dagger}$ & 1{,}000 \\
TRIBE v2                        & $\approx 4.85$B & $1.227$B canonical; $0.178$B released group form & released group form evaluated \\
VIBE                            & $\approx 21.55$B loaded ($\approx 20.64$B on active paths) & $0.329$--$0.722$B per specialist & 60 checkpoints in the S07 merge \\
ModalityRNN                     & $\approx 1.563$B loaded & $71.85$--$99.00$M & 100 \\
MedARC                          & $\approx 25.61$B loaded ($\approx 24.71$B on active paths) & $3.49$M$^{\dagger}$ & top-5 routing from 49 \\
\bottomrule
\end{tabular}
\end{table}

\begin{table}[h]
\centering
\caption{\textbf{Trainable parameters of \ourmodel\ by component.}}
\label{tab:mirage-parameters}
\small
\renewcommand{\arraystretch}{1.1}
\begin{tabular}{l r}
\toprule
\textbf{Component} & \textbf{Trainable parameters} \\
\midrule
Per-modality cross-attention layer poolers (gating) & $\approx 50.5$M \\
Modality projectors                                  & $\approx 151$M \\
Temporal brain encoder                               & $\approx 906$M \\
Learned temporal position embeddings                 & $\approx 3.1$M \\
Subject readouts (4 heads)                           & $\approx 12.3$M \\
\midrule
Total                                                & $\approx 1.123$B \\
\bottomrule
\end{tabular}
\end{table}

Per trained checkpoint, \ourmodel\ ($1.12$B) is comparable to TRIBE~v1 ($0.98$B) and canonical TRIBE~v2 ($1.23$B), larger than VIBE ($0.33$--$0.72$B), and much larger than ModalityRNN and MedARC. Relative to our matched fixed-pooling brain encoder, the cross-attention layer poolers add $188.9$M trainable parameters ($+20.2\%$). Most of this is the query bank: reducing from 24 to 12 queries removes $75.6$M parameters and changes $r$ by less than $0.001$ (Holm $p = 0.61$; App.~\ref{app:stats-validation}), so a substantially lighter configuration is available at negligible cost.

\subsection{Ensembling, Selection, and Compute}
\label{app:ensembling}
\label{app:ensemble-selection}

Final challenge submissions are produced by ensembling $15$ trained models. For model $k$, subject $s$, and parcel $p$, let $\rho^{(k)}_{s,p}$ denote the validation Pearson correlation. We convert these scores into non-negative ensemble weights with a softmax over models,
\begin{equation}
  w^{(k)}_{s,p}
  \;=\;
  \frac{\exp\!\big(\rho^{(k)}_{s,p}/\tau\big)}
       {\sum_{j} \exp\!\big(\rho^{(j)}_{s,p}/\tau\big)},
  \qquad \tau = 0.3,
\end{equation}
and the final prediction for each subject and parcel is the weighted average of the $15$ member predictions under these subject- and parcel-specific weights. We select 15 performance-competitive, configuration-diverse models from 110 completed runs spanning training seeds and architecture hyper-parameters (e.g., the number of cross-attention queries $n_q$).

\begin{table}[h]
\centering
\caption{\textbf{Ensemble candidate pool, selection policy, and downstream compute.}}
\label{tab:ensemble-compute}
\small
\renewcommand{\arraystretch}{1.2}
\setlength{\tabcolsep}{4pt}
\begin{tabular}{@{}>{\raggedright\arraybackslash}p{0.36\linewidth} >{\raggedright\arraybackslash}p{0.58\linewidth}@{}}
\toprule
\textbf{Item} & \textbf{Value} \\
\midrule
Completed candidate runs                    & 110 (varying seeds, architecture, and hyper-parameters) \\
Selected ensemble members                   & 15 \\
Selection policy                            & performance-competitive, configuration-diverse selection from the completed runs \\
Downstream training per run                 & $\approx 4$ GPU-h (1$\times$ A100 80GB) \\
Downstream training, all candidates         & $\approx 440$ GPU-h \\
Downstream training, selected members       & $\approx 60$ GPU-h \\
Cached feature extraction                   & $\approx 700$ GPU-h per Qwen-Omni backbone (all modalities; reported separately) \\
\bottomrule
\end{tabular}
\end{table}

The validation-split ensemble score in Figure~\ref{fig:main-result}a ($r = 0.335$) fits and evaluates the per-parcel softmax weights on the same split and is therefore in-sample; the held-out gains of the ensemble over the single model are $+0.013$ on S07 ($0.310 \to 0.323$) and $+0.010$ on OOD ($0.217 \to 0.227$). No scientific analysis in the paper depends on the ensemble: the cortical, modality, layer-attention, and matched-control analyses use the single \ourmodel\ checkpoint, and the scaling and transfer analyses use individual checkpoints. A single \ourmodel\ checkpoint ($r = 0.217$ on OOD) already exceeds all published ensembles in Table~\ref{tab:results}.

\subsection{Provenance of Compared Scores}
\label{app:parameters-provenance}

Table~\ref{tab:results} mixes numbers from different sources; Table~\ref{tab:provenance} makes the source of each row explicit.

\begin{table}[h]
\centering
\caption{\textbf{Provenance of the rows in Table~\ref{tab:results}.}}
\label{tab:provenance}
\small
\renewcommand{\arraystretch}{1.2}
\setlength{\tabcolsep}{4pt}
\begin{tabular}{@{}>{\raggedright\arraybackslash}p{0.36\linewidth} >{\raggedright\arraybackslash}p{0.58\linewidth}@{}}
\toprule
\textbf{Row} & \textbf{Source} \\
\midrule
Linear (challenge baseline)               & official Algonauts 2025 baseline, leaderboard values \\
Linear (Post-hoc)                         & our ridge fit on cached post-hoc (tower) features (App.~\ref{app:linear_baselines}); S07/OOD via Codabench \\
Linear (Ours)                             & our ridge fit on cached native-fusion features (App.~\ref{app:linear_baselines}); S07/OOD via Codabench \\
TRIBE v1 (single and ensemble)            & published paper / leaderboard values \\
TRIBE v2                                  & our evaluation of the released group head, projected into the benchmark parcel space, without subject-specific readout fitting; a lower bound under our protocol \\
Brain Encoder (Post-hoc)                  & our matched fixed-pooling control on post-hoc (tower) features; S07/OOD via Codabench \\
Brain Encoder (Ours)                      & our matched fixed-pooling control; S07/OOD via Codabench \\
MedARC, ModalityRNN, VIBE                 & published papers / leaderboard values \\
\ourmodel\ (single and ensemble)          & our runs; S07/OOD via Codabench \\
TRIBE v1-style (our reproduction; Apps.~\ref{app:tribe-cross-attention}, \ref{app:mirage-vs-tribe}, \ref{app:external-transfer}) & our re-evaluation of the released TRIBE~v1 pipeline on our split ($r = 0.3044$ on S06) \\
\bottomrule
\end{tabular}
\end{table}

\subsection{Code Availability}
\label{app:code-release}
We release the code, configuration files, and analysis scripts used to produce the results, together with the trained \ourmodel\ checkpoint and a project page:
\begin{center}
Code: \url{https://github.com/epflneuroailab/mirage} \\
Model card and weights: \url{https://huggingface.co/epfl-neuroai/mirage} \\
Project page: \url{https://mirage-brain.epfl.ch}
\end{center}
The repository includes the training and submission pipelines, feature-extraction scripts, experiment configurations, and figure-generation utilities.

%% file: appendices/B_statistics.tex
\section{Statistical Protocol}
\label{app:cr-statistics}
\label{app:statistics}

We report paired differences in mean Pearson correlation. The sections below specify the evaluation units, resampling procedures, and multiple-comparison corrections; individual captions identify the uncertainty shown.

\subsection{Units, Tests, and Corrections}
\label{app:statistical-protocol}

The independent unit of resampling is a stimulus unit or a participant. Parcels, time points, and subject-by-stimulus cells are not treated as independent replicates.

\paragraph{Estimand.} For a comparison between model $A$ and model $B$, the effect is the difference in mean Pearson $r$, where $r$ is computed per parcel over the time points of the evaluation split, averaged over the 1{,}000 parcels within each subject, and then averaged over subjects. Differences are always paired: both models are scored on identical stimuli, subjects, and parcels.

\paragraph{Independent units.}
\begin{itemize}
    \item \emph{Friends S06 (validation).} The unit is the episode (23 available episodes, 49 stimulus parts). Per-episode paired differences are averaged over subjects inside every resample.
    \item \emph{Friends S07 (in-distribution test).} Codabench returns only four subject-level aggregates and no time-aligned targets. S07 effects are reported descriptively as mean $\pm$ SEM across subjects with subject consistency; no $p$-value is attached.
    \item \emph{OOD movies.} Codabench returns six movie-level aggregates. The unit is the movie (six paired units).
    \item \emph{External datasets (App.~\ref{app:external-transfer}).} The unit is the participant (20 for HCP, 6 for CineBrain), after averaging over the four held-out stimuli within each participant.
\end{itemize}

\paragraph{Resampling and tests.} Confidence intervals are bias-corrected and accelerated (BCa) bootstrap intervals over 10{,}000 resamples of the independent units. Two-sided $p$-values on S06 come from 10{,}000 episode-label permutations of the paired differences; on OOD we use exact two-sided sign-flip (randomization) tests over the six movies, and on the external datasets exact two-sided sign tests across participants (ties dropped). A predictive-null diagnostic uses 5{,}000 circular shifts of the predictions with a minimum shift of 20 time points. The resampling seed is $260529850$.

\paragraph{Multiple comparisons.} Comparisons are grouped into pre-declared families (architecture: two contrasts on S06 and three on the held-out splits; native versus post-hoc: three backbones; modality complementarity: three leave-one-modality-out contrasts; layer-query sweep: nine contrasts; external transfer: three backbones within each dataset, alignment, and protocol, and within the pooled sample) and Holm-adjusted within each family. Parcel-wise maps are corrected jointly over the 1{,}000 parcels with both Benjamini--Hochberg false-discovery-rate control and Westfall--Young max-$T$ family-wise control; both parcel counts are reported. Headline verdicts use the adjusted $p$-value.

\paragraph{Robustness diagnostics.} For S06 comparisons we additionally recompute the effect after omitting each of the 23 episodes in turn (leave-one-episode-out) and report the number of omissions for which the sign is preserved.

\subsection{Validation-Split Comparisons}
\label{app:stats-validation}

Table~\ref{tab:stats-validation} lists every \emph{Friends} S06 contrast tested in the paper, grouped by family.

\begin{table}[h]
\centering
\caption{\textbf{Episode-level inference on \emph{Friends} S06.} $\Delta r$ is the paired difference in mean Pearson $r$ with its 95\% BCa episode-bootstrap CI; $p$ is the two-sided episode-label permutation $p$-value followed by the Holm-adjusted value within the family. Consistency reports subjects with a positive effect and leave-one-episode-out recomputations with a positive effect. Seeds: number of matched seed pairs averaged inside each resample.}
\label{tab:stats-validation}
\footnotesize
\renewcommand{\arraystretch}{1.1}
\setlength{\tabcolsep}{2.5pt}
\begin{tabular}{@{}l c c c c@{}}
\toprule
\textbf{Comparison} & $\boldsymbol{\Delta r}$ \textbf{[95\% BCa CI]} & \textbf{Perm.; Holm $p$} & \textbf{Consist.} & \textbf{Seeds} \\
\midrule
\multicolumn{5}{l}{\emph{Family: architecture (\ourmodel\ versus matched controls with the same Qwen3-Omni features)}} \\
Brain Encoder (Ours; native, no gating) & $+0.0181\ [+0.0165,\,+0.0198]$ & $10^{-4}$; $0.0002$ & 4/4; 23/23 & 1 \\
Brain Encoder (post-hoc, no gating)     & $+0.0265\ [+0.0244,\,+0.0285]$ & $10^{-4}$; $0.0002$ & 4/4; 23/23 & 1 \\
\midrule
\multicolumn{5}{l}{\emph{Family: native versus post-hoc fusion (tower) (\ourmodel\ architecture)}} \\
\modelname{Qwen2.5-Omni-7B}             & $+0.0121\ [+0.0102,\,+0.0145]$ & $10^{-4}$; $0.0003$ & 4/4; 23/23 & 1 \\
\modelname{Qwen3-Omni-30B-A3B-Instruct} & $+0.0092\ [+0.0075,\,+0.0107]$ & $10^{-4}$; $0.0003$ & 4/4; 23/23 & 1 \\
\modelname{Qwen3-Omni-30B-A3B-Thinking} & $+0.0142\ [+0.0126,\,+0.0157]$ & $10^{-4}$; $0.0003$ & 4/4; 23/23 & 1 \\
\midrule
\multicolumn{5}{l}{\emph{Family: modality complementarity (full trimodal minus leave-one-modality-out; App.~\ref{app:modality-maps})}} \\
Unique text (full $-$ audio+vision)     & $+0.0032\ [+0.0023,\,+0.0040]$ & $10^{-4}$; $0.0003$ & 4/4; 23/23 & 3 \\
Unique vision (full $-$ audio+text)     & $+0.0289\ [+0.0267,\,+0.0314]$ & $10^{-4}$; $0.0003$ & 4/4; 23/23 & 3 \\
Unique audio (full $-$ text+vision)     & $+0.0077\ [+0.0068,\,+0.0087]$ & $10^{-4}$; $0.0003$ & 4/4; 23/23 & 3 \\
\midrule
\multicolumn{5}{l}{\emph{Family: layer-query sweep (24 queries versus $n_q$; App.~\ref{app:n_queries_ablation})}} \\
24 vs.\ 1   & $+0.0051\ [+0.0045,\,+0.0058]$ & $10^{-4}$; $0.0009$ & 4/4; 23/23 & 3 \\
24 vs.\ 2   & $+0.0037\ [+0.0030,\,+0.0044]$ & $10^{-4}$; $0.0009$ & 4/4; 23/23 & 3 \\
24 vs.\ 3   & $+0.0021\ [+0.0013,\,+0.0029]$ & $10^{-4}$; $0.0009$ & 4/4; 23/23 & 3 \\
24 vs.\ 4   & $+0.0023\ [+0.0013,\,+0.0031]$ & $0.0004$; $0.0024$ & 4/4; 23/23 & 3 \\
24 vs.\ 5   & $+0.0018\ [+0.0010,\,+0.0027]$ & $0.0014$; $0.0070$ & 4/4; 23/23 & 3 \\
24 vs.\ 8   & $+0.0013\ [+0.0006,\,+0.0021]$ & $0.0028$; $0.0112$ & 4/4; 23/23 & 3 \\
24 vs.\ 12  & $+0.0004\ [-0.0004,\,+0.0011]$ & $0.3039$; $0.6077$ & 3/4; 23/23 & 3 \\
24 vs.\ 16  & $+0.0007\ [+0.0001,\,+0.0012]$ & $0.0318$; $0.0954$ & 4/4; 23/23 & 3 \\
24 vs.\ 32  & $-0.0002\ [-0.0006,\,+0.0003]$ & $0.4779$; $0.6077$ & 1/4; 0/23  & 3 \\
\midrule
\multicolumn{5}{l}{\emph{Single comparison (App.~\ref{app:single-layer})}} \\
Most-attended layer (28/29/27) & $+0.0159\ [+0.0143,\,+0.0171]$ & $10^{-4}$; -- & 4/4; 23/23 & 1 \\
\bottomrule
\end{tabular}
\end{table}

\subsection{Held-Out Comparisons}
\label{app:stats-heldout}

Table~\ref{tab:stats-heldout} lists the held-out effects of the architecture family on \emph{Friends} S07 and the OOD movies.

\begin{table}[h]
\centering
\caption{\textbf{Held-out effects for the architecture family.} S07 values are descriptive (mean $\pm$ SEM across the four Codabench subject summaries; no $p$-value). OOD values use the six movies as paired units: movie-bootstrap 95\% CI, exact two-sided sign-flip $p$, and Holm-adjusted $p$ over the three contrasts. Consistency counts subjects (S07) and movies; subjects (OOD) with a positive effect.}
\label{tab:stats-heldout}
\footnotesize
\renewcommand{\arraystretch}{1.1}
\setlength{\tabcolsep}{2.5pt}
\begin{tabular}{@{}l cc ccc@{}}
\toprule
 & \multicolumn{2}{c}{\textbf{S07}} & \multicolumn{3}{c}{\textbf{OOD}} \\
\cmidrule(lr){2-3} \cmidrule(lr){4-6}
\textbf{\ourmodel\ vs.} & $\boldsymbol{\Delta r} \pm$ \textbf{SEM} & \textbf{Consist.} & $\boldsymbol{\Delta r}$ \textbf{[95\% CI]} & \textbf{Exact; Holm $p$} & \textbf{Consist.} \\
\midrule
Brain Encoder (Ours)         & $+0.0188 \pm 0.0021$ & 4/4 & $+0.0218\ [+0.0167,\,+0.0282]$ & $0.0313$; $0.0938$ & 6/6; 4/4 \\
Brain Encoder (post-hoc)     & $+0.0280 \pm 0.0019$ & 4/4 & $+0.0425\ [+0.0346,\,+0.0540]$ & $0.0313$; $0.0938$ & 6/6; 4/4 \\
Linear (Ours)                & $+0.0871 \pm 0.0051$ & 4/4 & $+0.0831\ [+0.0699,\,+0.0946]$ & $0.0313$; $0.0938$ & 6/6; 4/4 \\
\bottomrule
\end{tabular}
\end{table}

\paragraph{Reading the held-out tests.} Every OOD effect is positive for all six movies and all four subjects. With six paired units the smallest attainable exact two-sided $p$ is $0.03125$; after Holm correction over the three architecture contrasts this becomes $0.09375$, which does not pass the family-wise $\alpha = 0.05$. We therefore describe the held-out gating and fusion effects as consistent across every held-out unit and quantified by their intervals, but not as family-wise significant at the movie level. The S06 tests, which have 23 episode units, pass correction comfortably. Together the two splits provide complementary evidence: precise inference in-domain, and fully held-out but low-powered inference out-of-domain.

%% file: appendices/C_fusion_controls.tex
\section{Native Fusion Controls on CNeuroMod}
\label{app:cr-fusion-controls}
\label{app:fusion-controls}

\subsection{Three Feature Conditions from One Backbone}
\label{app:fusion-conditions}

The tower-based comparison of \S\ref{sec:sota} changes both the feature readout location and the availability of cross-modal context. We therefore compare three feature conditions from the same frozen backbone:

\begin{enumerate}
    \item \textbf{Post-hoc fusion (tower).} Vision and audio are read from the corresponding Qwen-Omni encoder towers and text from the language module applied to the transcript alone. No modality conditions any other; fusion happens only downstream in the brain encoder. This is the ``post-hoc'' condition of Figure~\ref{fig:main-result}.
    \item \textbf{Post-hoc fusion (decoder).} Each modality is passed alone through the full backbone, and its stream is read from the same language-module layers used in the native condition: $(V_{\text{vision-only}},\,A_{\text{audio-only}},\,T_{\text{text-only}})$. The streams are concatenated post hoc. This condition matches the native condition in depth, token readout space, temporal alignment, and layer-gating capacity, and differs only in the absence of cross-modal context during extraction.
    \item \textbf{Native fusion.} All three streams are read from the language module after a joint audio--video--text forward pass: $(V_{\text{joint}},\,A_{\text{joint}},\,T_{\text{joint}})$. This is the \ourmodel\ configuration.
\end{enumerate}

Across the three conditions we hold fixed the backbone weights, the 2\,Hz feature grid and hemodynamic alignment (App.~\ref{app:temporal-alignment}), the per-modality 24-query cross-attention layer poolers, the 8-layer brain encoder, the subject readouts, the optimizer and schedule, the data split, and the training seed. The controlled contrast for cross-modal conditioning is therefore \emph{native fusion minus post-hoc fusion (decoder)}; the contrast \emph{post-hoc fusion (decoder) minus post-hoc fusion (tower)} isolates the effect of reading features from the deeper language module rather than from the encoder towers.

\subsection{Results on the Validation Split}
\label{app:fusion-matched-results}

\begin{figure}[h]
    \centering
    \includegraphics[width=0.5\linewidth]{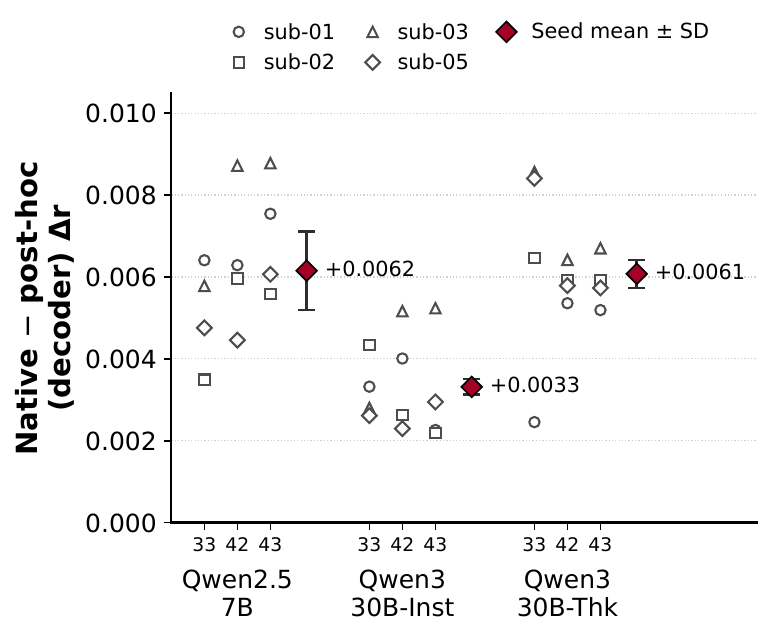}
    \caption{\textbf{Native minus post-hoc fusion (decoder) on \emph{Friends} S06 across training seeds.} Per-subject differences (sub-01 $\circ$, sub-02 $\square$, sub-03 $\triangle$, sub-05 $\diamond$) for seeds 33, 42, and 43 of the three larger backbones; the filled diamond is the mean of the three seed means $\pm$ SD across seeds. Every subject is positive in every seed.}
    \label{fig:fusion-seed-gaps}
\end{figure}

Table~\ref{tab:fusion-three-conditions} reports all three conditions for four Qwen-Omni backbones on \emph{Friends} S06. Joint conditioning improves over the post-hoc (decoder) control for every backbone and for all four subjects, with effects between $+0.003$ and $+0.007$. The decoder controls are themselves above the tower controls by $+0.002$ to $+0.008$, so the original tower-based gap was not a clean estimate of cross-modal conditioning alone: the decoder control retains $35$--$46\%$ of the tower gap for the 7B and 30B backbones and $71\%$ for the 3B backbone. Figure~\ref{fig:fusion-seed-gaps} shows that the decoder contrast is stable across training seeds.

\begin{table}[h]
\centering
\caption{\textbf{Three feature conditions from the same backbone on \emph{Friends} S06.} Mean Pearson $r$ across the four subjects ($\pm$ SEM across subjects). ``Decoder $-$ tower'' is the depth/space difference; ``Native $-$ decoder'' is the fusion difference (paired across subjects, $\pm$ SEM of the paired differences), and the last row counts subjects for which it is positive. One training seed per model.}
\label{tab:fusion-three-conditions}
\footnotesize
\renewcommand{\arraystretch}{1.1}
\setlength{\tabcolsep}{4pt}
\begin{tabular}{@{}l cccc@{}}
\toprule
 & \modelname{Qwen2.5-Omni} & \modelname{Qwen2.5-Omni} & \modelname{Qwen3-Omni-30B} & \modelname{Qwen3-Omni-30B} \\
\textbf{Condition} & \modelname{3B} & \modelname{7B} & \modelname{A3B-Instruct} & \modelname{A3B-Thinking} \\
\midrule
Post-hoc fusion (tower)   & $0.2977 \pm 0.0082$ & $0.2963 \pm 0.0076$ & $0.3042 \pm 0.0084$ & $0.3044 \pm 0.0082$ \\
Post-hoc fusion (decoder) & $0.2995 \pm 0.0082$ & $0.3033 \pm 0.0084$ & $0.3102 \pm 0.0090$ & $0.3121 \pm 0.0090$ \\
Native fusion             & $0.3040 \pm 0.0088$ & $0.3084 \pm 0.0085$ & $0.3134 \pm 0.0091$ & $0.3186 \pm 0.0092$ \\
\midrule
Decoder $-$ tower         & $+0.0018$ & $+0.0070$ & $+0.0059$ & $+0.0078$ \\
Native $-$ decoder        & $+0.0045 \pm 0.0009$ & $+0.0051 \pm 0.0006$ & $+0.0033 \pm 0.0004$ & $+0.0065 \pm 0.0014$ \\
Positive subjects         & 4/4 & 4/4 & 4/4 & 4/4 \\
\bottomrule
\end{tabular}
\end{table}

The matched contrast establishes a collective cross-modal-context effect within the Qwen-Omni family: features extracted with the other modalities present predict cortex better than features of identical depth and space extracted without them. The effect is small in absolute terms ($\Delta r \approx 0.003$--$0.007$) but it is positive at all four scales and for every subject, comparable to the depth and head-count effects of App.~\ref{app:architecture-loss-ablations} at no additional parameter cost. It is also a lower bound on the benefit of cross-modal conditioning: the language module was trained multimodally, so its unimodal-forward representations are already aligned across modalities to a degree even without the other modalities in the forward pass.

\subsection{Native versus Post-hoc Fusion on Held-Out Benchmarks}
\label{app:fusion-heldout}

\begin{figure}[h]
    \centering
    \includegraphics[width=0.48\linewidth]{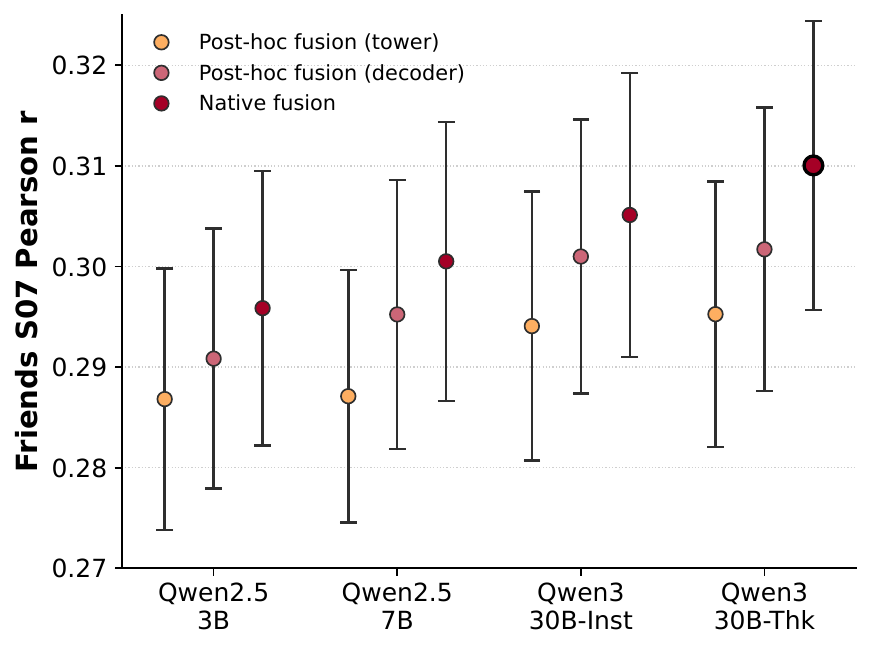}\hfill
    \includegraphics[width=0.48\linewidth]{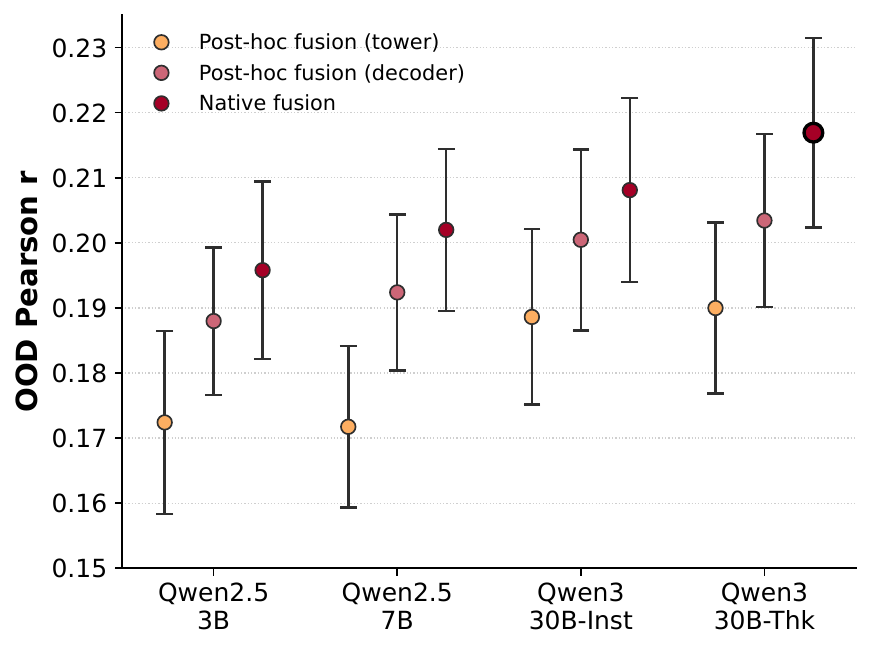}
    \caption{\textbf{Three feature conditions on the held-out benchmarks.} Mean Pearson $r \pm$ SEM across the four subjects (Codabench subject summaries, seed 33) for post-hoc fusion (tower), post-hoc fusion (decoder), and native fusion on \emph{Friends} S07 (left) and the OOD movies (right); the marker with the black edge is \ourmodel. The validation split is shown in Figure~\ref{fig:main-result}b.}
    \label{fig:backbone-scale-heldout}
\end{figure}

The held-out \emph{Friends} S07 and OOD results preserve the native-fusion advantage across all four backbones. Table~\ref{tab:backbone-scale-heldout} reports the tower comparison: gains range from $+0.009$ to $+0.015$ on S07 and from $+0.020$ to $+0.030$ on OOD. Figure~\ref{fig:backbone-scale-heldout} additionally shows the decoder-matched control. Native fusion exceeds this control by $+0.004$--$+0.008$ on S07 and $+0.008$--$+0.014$ on OOD, with a positive difference for every subject and backbone.

These comparisons extend the validation result beyond S06 and the default 30B backbone. Scores are reported descriptively as mean $\pm$ SEM across the four Codabench subject summaries; the formal held-out tests for the architecture comparisons are reported in App.~\ref{app:stats-heldout}.

\begin{table}[h]
\centering
\caption{\textbf{Post-hoc fusion (tower) versus native fusion across backbones and splits.} Mean Pearson $r \pm$ SEM across the four subjects. S06 uses our validation artifacts; S07 and OOD use the Codabench subject summaries. $\Delta r$ is native minus post-hoc with the SEM of the paired subject differences. All models use the \ourmodel\ architecture with cross-attention layer poolers.}
\label{tab:backbone-scale-heldout}
\small
\renewcommand{\arraystretch}{1.1}
\setlength{\tabcolsep}{4pt}
\begin{tabular}{l l ccc}
\toprule
\textbf{Backbone} & \textbf{Split} & \textbf{Post-hoc} & \textbf{Native} & $\boldsymbol{\Delta r}$ \\
\midrule
\multirow{3}{*}{\modelname{Qwen2.5-Omni-3B}}
 & S06 & $0.2977 \pm 0.0082$ & $0.3040 \pm 0.0088$ & $+0.0063 \pm 0.0008$ \\
 & S07 & $0.2868 \pm 0.0130$ & $0.2959 \pm 0.0136$ & $+0.0090 \pm 0.0009$ \\
 & OOD & $0.1724 \pm 0.0140$ & $0.1958 \pm 0.0136$ & $+0.0234 \pm 0.0014$ \\
\midrule
\multirow{3}{*}{\modelname{Qwen2.5-Omni-7B}}
 & S06 & $0.2963 \pm 0.0076$ & $0.3084 \pm 0.0085$ & $+0.0121 \pm 0.0010$ \\
 & S07 & $0.2871 \pm 0.0125$ & $0.3005 \pm 0.0139$ & $+0.0134 \pm 0.0014$ \\
 & OOD & $0.1717 \pm 0.0124$ & $0.2020 \pm 0.0125$ & $+0.0302 \pm 0.0023$ \\
\midrule
\multirow{3}{*}{\modelname{Qwen3-Omni-30B-A3B-Instruct}}
 & S06 & $0.3042 \pm 0.0084$ & $0.3134 \pm 0.0091$ & $+0.0092 \pm 0.0007$ \\
 & S07 & $0.2941 \pm 0.0134$ & $0.3051 \pm 0.0141$ & $+0.0110 \pm 0.0011$ \\
 & OOD & $0.1886 \pm 0.0135$ & $0.2081 \pm 0.0141$ & $+0.0195 \pm 0.0039$ \\
\midrule
\multirow{3}{*}{\modelname{Qwen3-Omni-30B-A3B-Thinking}}
 & S06 & $0.3044 \pm 0.0082$ & $0.3186 \pm 0.0092$ & $+0.0142 \pm 0.0012$ \\
 & S07 & $0.2953 \pm 0.0132$ & $0.3100 \pm 0.0144$ & $+0.0148 \pm 0.0013$ \\
 & OOD & $0.1900 \pm 0.0131$ & $0.2169 \pm 0.0146$ & $+0.0269 \pm 0.0039$ \\
\bottomrule
\end{tabular}
\end{table}

All matched controls in this section use the Qwen-Omni family at the full \ourmodel\ architectural level (cross-attention layer poolers plus brain encoder). App.~\ref{app:external-transfer} extends the native versus post-hoc (decoder) comparison to HCP 7T Movie and CineBrain.

%% file: appendices/D_layer_aggregation.tex
\section{Layer Aggregation and the Cross-Attention Pooler}
\label{app:cr-layer-aggregation}
\label{app:layer-aggregation-controls}

We test whether adaptive aggregation improves on a fixed layer selected from the learned attention profiles and whether it helps on TRIBE-style features. We then examine the number of queries and describe how the pooler attention profiles are extracted.

\subsection{Most-Attended Single-Layer Baseline}
\label{app:single-layer}

We train a model identical to \ourmodel\ except that the cross-attention layer pooler of each modality is replaced by a fixed read of the layer that receives the highest average attention weight in the trained \ourmodel\ (Fig.~\ref{fig:layer_attention}): layer 28 for vision, 29 for audio, and 27 for text. The native-fusion features, brain encoder, readouts, optimizer, schedule, split, and seed are unchanged. As an auxiliary reference we also train the same fixed-layer model on the post-hoc fusion (decoder) features of App.~\ref{app:fusion-conditions}.

\begin{table}[h]
\centering
\caption{\textbf{Adaptive aggregation versus the most-attended single layer.} Mean \emph{Friends} S06 Pearson $r$ across subjects; one matched training seed. The inferential row uses the episode-level protocol of App.~\ref{app:statistical-protocol} (10{,}000 paired episode bootstraps for the BCa interval; 10{,}000 two-sided label permutations).}
\label{tab:single-layer}
\footnotesize
\renewcommand{\arraystretch}{1.1}
\setlength{\tabcolsep}{3pt}
\begin{tabular}{l c c c}
\toprule
\textbf{Layer aggregation} & \textbf{Vision / audio / text layers} & $\boldsymbol{r}$ & $\boldsymbol{\Delta}$ \textbf{vs.\ \ourmodel} \\
\midrule
\ourmodel\ (24-query cross-attention)                    & all 48 / all 48 / all 48 & $0.3186$ & -- \\
Fixed single layer (native fusion)                       & 28 / 29 / 27             & $0.3027$ & $-0.0159$ \\
Fixed single layer (post-hoc fusion, decoder) & 28 / 29 / 27             & $0.2977$ & $-0.0209$ \\
\midrule
\multicolumn{4}{l}{\ourmodel\ $-$ fixed single layer (native): $\Delta r = +0.0159$, 95\% BCa CI $[+0.0143,\,+0.0171]$, permutation $p = 10^{-4}$,} \\
\multicolumn{4}{l}{positive for 4/4 subjects and for 23/23 leave-one-episode-out recomputations.} \\
\bottomrule
\end{tabular}
\end{table}

\begin{figure}[h]
    \centering
    \includegraphics[width=0.30\linewidth]{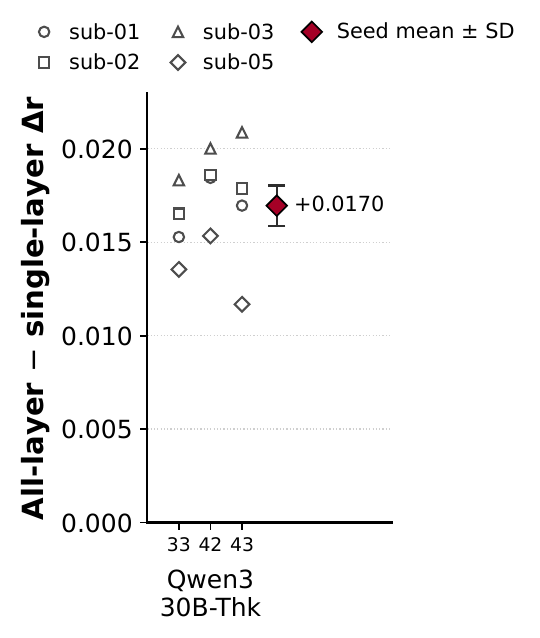}\hspace{0.06\linewidth}
    \includegraphics[width=0.36\linewidth]{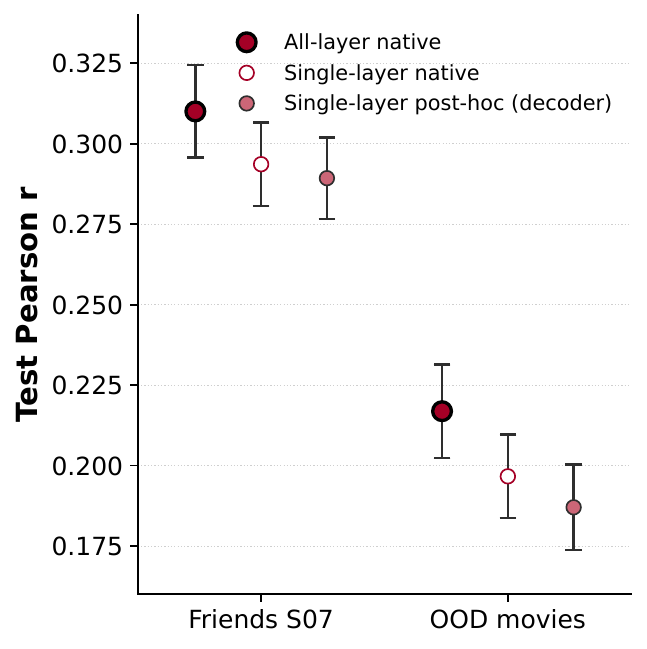}
    \caption{\textbf{Adaptive aggregation versus the most-attended single layer.} Left: all-layer minus single-layer native fusion on \emph{Friends} S06 per subject (sub-01 $\circ$, sub-02 $\square$, sub-03 $\triangle$, sub-05 $\diamond$) for seeds 33, 42, and 43; the filled diamond is the mean of the seed means $\pm$ SD across seeds. Right: held-out \emph{Friends} S07 and OOD scores (mean $\pm$ subject SEM) for all-layer native fusion, single-layer native fusion, and single-layer post-hoc fusion (decoder).}
    \label{fig:single-layer}
\end{figure}

Adaptive aggregation outperforms the most-attended fixed layers by $\Delta r = 0.016$, a gap similar to its gain over the fixed-pooling encoder ($+0.018$; App.~\ref{app:stats-validation}). The seed and held-out comparisons in Figure~\ref{fig:single-layer} show that this advantage extends beyond the original training run and validation split. Together with the query-count sweep below, these results support aggregating information across layers rather than reading only the layers emphasized by the average attention profiles.

\subsection{Cross-Attention Aggregation on TRIBE Features}
\label{app:tribe-cross-attention}

To test whether the aggregation mechanism helps independently of the Qwen-Omni backbone, we apply it to the frozen TRIBE-style unimodal features (\modelname{Llama-3.2-3B}, \modelname{Wav2Vec-BERT~2.0}, \modelname{V-JEPA~2}; App.~\ref{app:feature-extraction}). Table~\ref{tab:tribe-xattn-design} lists the four variants; the first three share the same feature caches, split, seed, and 8-layer brain encoder, and differ only in layer aggregation and training recipe.

\begin{table}[h]
\centering
\caption{\textbf{Design of the TRIBE-feature aggregation control.}}
\label{tab:tribe-xattn-design}
\small
\renewcommand{\arraystretch}{1.2}
\setlength{\tabcolsep}{4pt}
\begin{tabular}{@{}>{\raggedright\arraybackslash}p{0.26\linewidth} >{\raggedright\arraybackslash}p{0.19\linewidth} >{\raggedright\arraybackslash}p{0.28\linewidth} >{\raggedright\arraybackslash}p{0.17\linewidth}@{}}
\toprule
\textbf{Variant} & \textbf{Frozen features} & \textbf{Layer aggregation} & \textbf{Training recipe} \\
\midrule
TRIBE fixed pooling (matched reference) & TRIBE-style unimodal & 3 fractional-depth group means, concatenated & TRIBE (Adam, OneCycle, SWA) \\
TRIBE + \ourmodel\ cross-attention      & same caches          & all layers; 24-query cross-attention per modality & TRIBE \\
TRIBE + cross-attention + \ourmodel\ recipe & same caches      & all layers; 24-query cross-attention per modality & \ourmodel\ (AdamW, OneCycle, no SWA) \\
\ourmodel                                & Qwen3-Omni native fusion & all layers; 24-query cross-attention per modality & \ourmodel \\
\bottomrule
\end{tabular}
\end{table}

\begin{table}[h]
\centering
\caption{\textbf{Adaptive aggregation on fixed TRIBE features (\emph{Friends} S06).} Mean Pearson $r \pm$ SEM across subjects; one seed per model.}
\label{tab:tribe-xattn}
\small
\renewcommand{\arraystretch}{1.1}
\setlength{\tabcolsep}{4pt}
\begin{tabular}{l c c c}
\toprule
\textbf{Variant} & $\boldsymbol{r}$ & $\boldsymbol{\Delta}$ \textbf{vs.\ fixed pooling} & \textbf{Positive subjects} \\
\midrule
TRIBE fixed pooling                          & $0.3012 \pm 0.0087$ & --       & -- \\
TRIBE + cross-attention                      & $0.3054 \pm 0.0090$ & $+0.0042$ & 4/4 \\
TRIBE + cross-attention + \ourmodel\ recipe  & $0.3056 \pm 0.0092$ & $+0.0044$ & 4/4 \\
\ourmodel                                    & $0.3186 \pm 0.0092$ & $+0.0174$ & 4/4 \\
\bottomrule
\end{tabular}
\end{table}

\begin{figure}[h]
    \centering
    \includegraphics[width=0.6\linewidth]{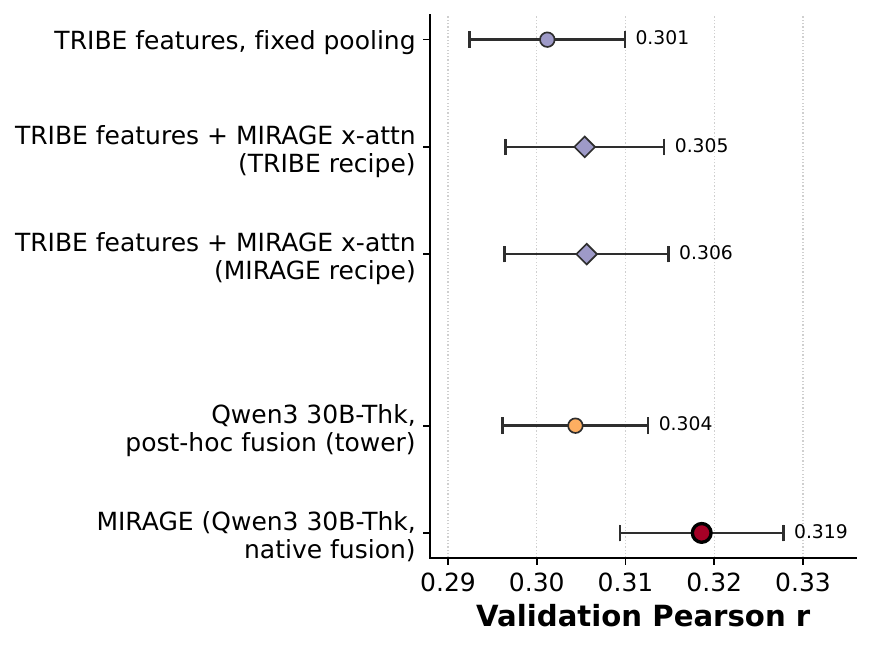}
    \caption{\textbf{Cross-attention layer aggregation on TRIBE features.} \emph{Friends} S06 Pearson $r$ (mean $\pm$ subject SEM) for the TRIBE-feature model with fixed pooling or with the \ourmodel\ cross-attention poolers (TRIBE and \ourmodel\ training recipes), next to the \modelname{Qwen3-Omni-30B-A3B-Thinking} post-hoc fusion (tower) model and \ourmodel.}
    \label{fig:tribe-xattn}
\end{figure}

Cross-attention aggregation improves the TRIBE-feature model from $0.301$ to $0.305$ with the same features, split, seed, encoder, and training recipe, with gains for all four subjects. Switching to the \ourmodel\ training recipe adds less than $0.001$. The aggregation mechanism therefore helps on a second feature family, although it does not close the gap to \ourmodel. The within-Qwen comparisons in App.~\ref{app:fusion-conditions} separately assess the contribution of joint multimodal conditioning.

\subsection{Number of Queries}
\label{app:n_queries_ablation}
The cross-attention layer pooler on top of each modality's frozen feature stream uses a fixed bank of $n_q$ learned queries that pool the layer-axis tokens at each time step into a per-modality, time-aligned representation before fusion (Section~\ref{sec:methods}). The compute and parameter cost of this block scale linearly with $n_q$, while $n_q$ also caps how much representational structure the pooler can preserve per time step. We sweep $n_q \in \{1, 2, 3, 4, 5, 8, 12, 16, 24, 32\}$, holding all other hyper-parameters at their main-paper defaults: \modelname{Qwen3-Omni-30B-A3B-Thinking} post-fusion features for text, audio, and vision; hidden dimension $D = 3072$; AdamW with a one-cycle schedule ($\mathrm{lr} = 10^{-4}$, $10\%$ warm-up, weight decay $10^{-2}$); batch size $16$; $15$ epochs; gradient clipping at $\ell_2 = 1.0$; 16-bit mixed precision. Each setting is trained with $3$ seeds ($42$--$44$); we report \emph{Friends} S06 validation Pearson correlation (mean $\pm$ SEM over seeds).

\begin{figure}[h]
  \centering
  \includegraphics[width=0.6\linewidth]{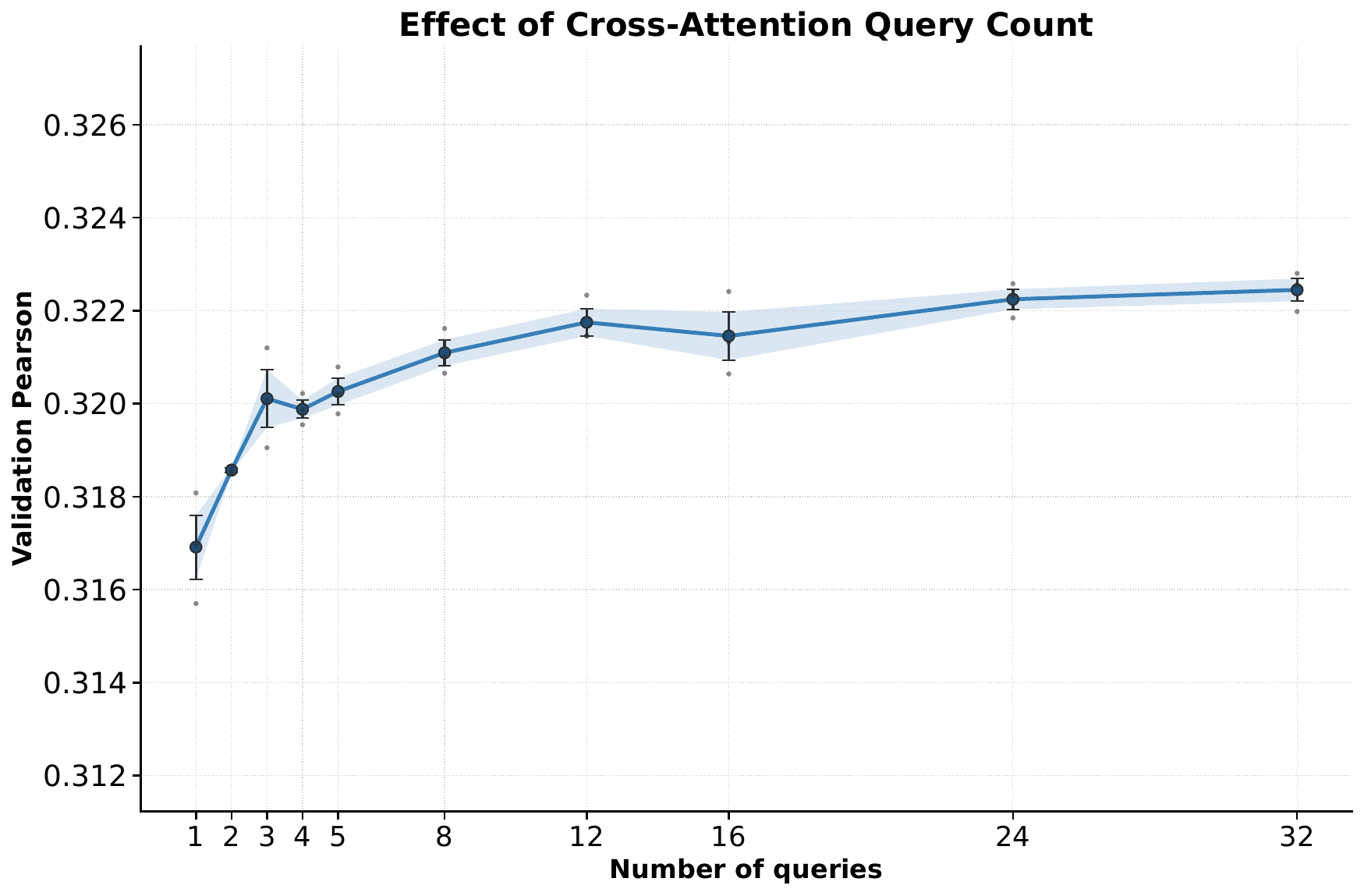}
  \caption{Validation Pearson correlation as a function of the number of cross-attention queries $n_q$ per modality. Solid line: seed mean; shaded band: SEM over $n{=}3$ seeds; gray dots: individual seeds.}
  \label{fig:n_queries_ablation}
\end{figure}

Validation Pearson correlation rises monotonically with $n_q$ on average, with most of the benefit concentrated in the small-$n_q$ regime: moving from $n_q{=}1$ to $n_q{=}4$ recovers roughly $55\%$ of the total $1\!\to\!32$ gain ($\Delta r \approx 0.0030$ out of $0.0055$), $n_q{=}4 \to n_q{=}12$ adds another $\Delta r \approx 0.0019$, and $n_q{=}12 \to n_q{=}32$ contributes only $\Delta r \approx 0.0007$. Seed variance is very small at $n_q \geq 2$ (SEM $< 10^{-3}$), so the saturation pattern is robust rather than driven by noise. Single-query pooling ($n_q{=}1$) is the only setting that is both clearly worse \emph{and} markedly noisier than the rest, indicating that compressing each modality's layer stack to a single weighted layer average under-fits the structure the downstream encoder relies on. Beyond $n_q{=}12$, returns are negligible relative to the added compute and parameter count, so we adopt $n_q = 24$ for the main results to stay on the flat part of the curve while remaining within our compute budget; $n_q \in \{12, 16\}$ is a reasonable lighter-weight alternative with a $\sim\!0.001$ Pearson cost.

\subsection{Cross-Attention Layer-Pooler Contribution}
\label{app:attention-contribution}
We use the attention weights of \ourmodel's cross-attention layer poolers as a model-level diagnostic of which backbone layers each modality stream prefers, with a resolution that lets us condition on attention head, query, and time step. This section specifies how those weights are extracted from a trained model and averaged into the profiles shown in the main paper.

\paragraph{Source of the weights.}
For modality $m \in \{\mathrm{text},\,\mathrm{audio},\,\mathrm{vision}\}$, the layer pooler is a single multi-head cross-attention block (Section~\ref{sec:methods}) whose queries are $n_q$ learned tokens and whose keys and values are the per-time-step layer stack $H^m_{:,t,:} \in \mathbb{R}^{L_m \times d_m}$ of the frozen backbone. We hook the underlying module so that each forward pass yields

\begin{equation*}
    \Pi^m \;\in\; \mathbb{R}^{B \times T \times h \times n_q \times L_m},
\end{equation*}

where $B$ is the batch size, $T$ the number of stimulus frames in the input window, $h$ the number of attention heads, $n_q$ the number of queries, and $L_m$ the number of backbone layers exposed to the pooler. The entry $\Pi^{m}_{b,t,k,q,\ell}$ is the post-softmax weight that head $k$, query $q$ assigns to layer $\ell$ at time step $t$ of stimulus example $b$. For \ourmodel\ we have $h = 4$ and $n_q = 24$ across all three modalities, and $L_m$ matches the layer count exposed by the chosen Qwen-Omni stream.

\paragraph{Aggregation across stimuli and time.}
Profiles are computed in inference mode on the validation split, with backbone features cached and modality dropout disabled. We accumulate $\Pi^m$ across a fixed set of validation batches and average along the stimulus and time axes,
\begin{equation*}
\bar\Pi^{m}_{k,q,\ell}
\;=\;
\frac{1}{B^{\star}\, T^{\star}}\sum_{b=1}^{B^{\star}}\sum_{t=1}^{T^{\star}} \Pi^{m}_{b,t,k,q,\ell}
\;\;\in\;\; \mathbb{R}^{h \times n_q \times L_m},
\end{equation*}
where $B^{\star}$ and $T^{\star}$ are the total numbers of validation stimuli and time steps used in the analysis. $\bar\Pi^m$ is the densest tensor we keep; coarser views in the main paper are obtained by averaging $\bar\Pi^m$ over selected remaining axes.

\paragraph{Reductions used in the figures.}
The per-modality depth profile (one curve per modality) marginalizes both heads and queries,
\begin{equation*}
\overline{\pi}^{m}_{\ell}
\;=\;
\frac{1}{h\,n_q}\sum_{k=1}^{h}\sum_{q=1}^{n_q} \bar\Pi^{m}_{k,q,\ell}.
\end{equation*}
Per-head profiles condition on a head $k$ and average over queries,
\begin{equation*}
\overline{\pi}^{m,k}_{\ell} \;=\; \frac{1}{n_q}\sum_{q=1}^{n_q} \bar\Pi^{m}_{k,q,\ell},
\end{equation*}
while per-query profiles condition on a query $q$ and average over heads,
\begin{equation*}
\overline{\pi}^{m,q}_{\ell} \;=\; \frac{1}{h}\sum_{k=1}^{h} \bar\Pi^{m}_{k,q,\ell}.
\end{equation*}
A TR-resolved profile, which exposes any temporal drift in layer preference within a window, averages over heads, queries, and stimuli but \emph{not} over time:
\begin{equation*}
\overline{\pi}^{m}_{t,\ell}
\;=\;
\frac{1}{B^{\star}\,h\,n_q}\sum_{b=1}^{B^{\star}}\sum_{k=1}^{h}\sum_{q=1}^{n_q} \Pi^{m}_{b,t,k,q,\ell}
\;\;\in\;\; \mathbb{R}^{T^{\star} \times L_m}.
\end{equation*}

\paragraph{Practical notes.}
Each entry of $\Pi^m$ is already a probability over layers (the per-layer softmax of \S\ref{sec:methods}), so every reduction above is an unweighted per-axis mean; no additional normalization is applied and no entry is rescaled. Heads and queries are treated as exchangeable for averaging, since none of our analyses depend on which specific head or query a particular pattern came from. The pooler hook itself is a simple flag that toggles \texttt{need\_weights=True} on the underlying \texttt{MultiheadAttention} module, so the weight tensor is only materialized when explicitly requested.

\subsubsection{Per-Head Cross-Attention Profiles}
\label{app:attention_heads}

The main-text Figure~\ref{fig:layer_attention} reports cross-attention weights averaged across the four heads of each modality-specific gating module. Averaging is convenient for visualization but could in principle obscure structure: a broad head-averaged profile might reflect either a single head with diffuse weight or multiple heads with sharply peaked weight at different depths.

\begin{figure}
    \centering
    \includegraphics[width=1\linewidth]{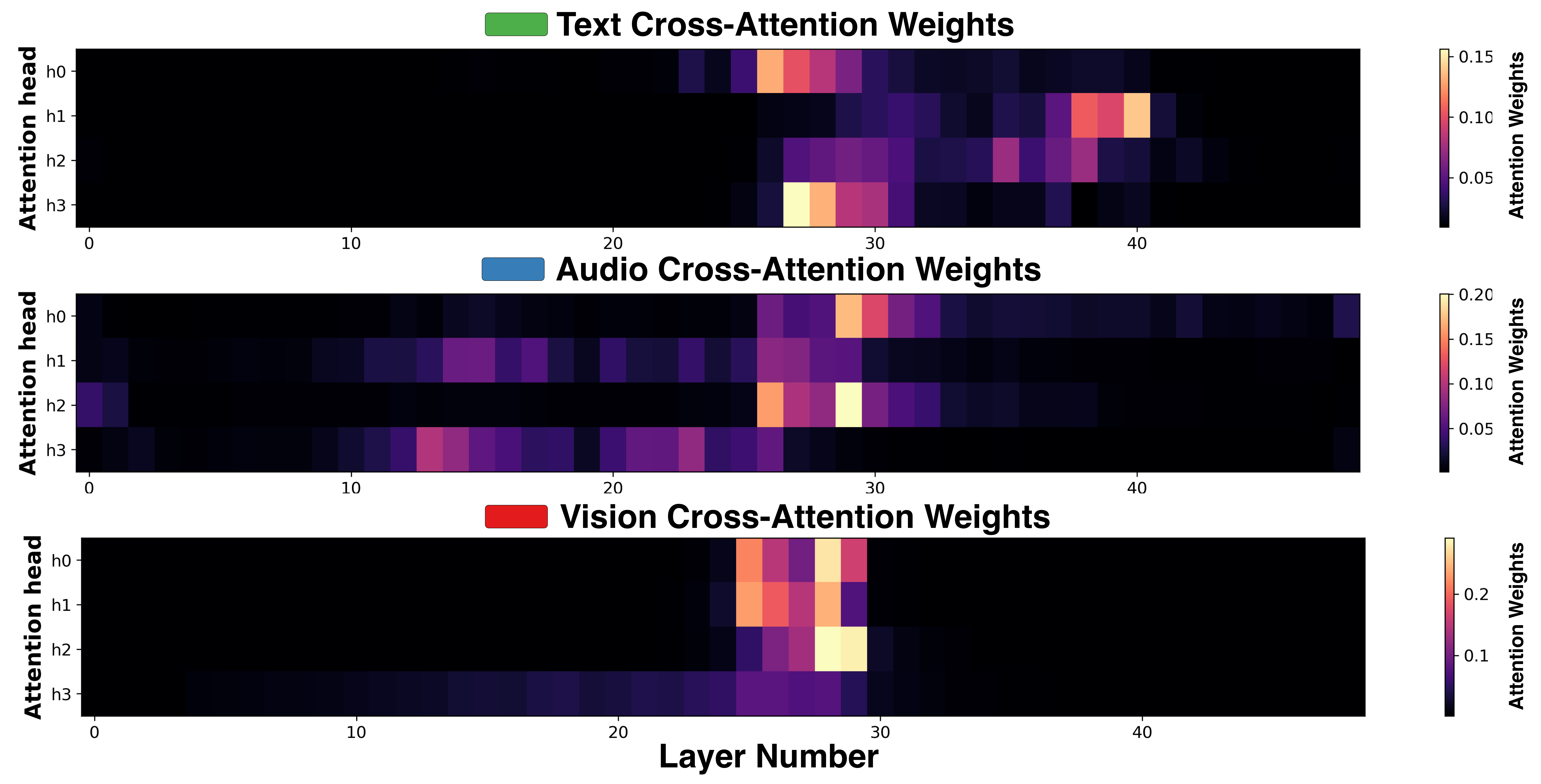}
    \caption{\textbf{Per-head cross-attention weights over \textsc{Qwen3-Omni} layers, by modality.} Attention weights from each of the four heads ($h_0$--$h_3$) of the three modality-specific cross-attention modules (text, audio, vision) over the 48 layers of the \textsc{Qwen3-Omni} language module, averaged across the 24 latent queries; brighter cells indicate stronger contribution to the modality-specific readout. The modality-specific depth profiles reported in Figure~\ref{fig:layer_attention} are preserved at the head level: three of the four vision heads concentrate in a narrow mid-depth band ($\sim$layers 25--30) and the fourth spreads more diffusely over the same range; audio shows greater per-head variability, with some heads sharply peaked in this band and others spreading weight across earlier mid-depth layers; text exhibits the clearest head-level division of labor, with subsets of heads peaking around layers 25--30 ($h_0, h_3$) and around layers 38--40 ($h_1, h_2$). The broader text profile in the head-averaged main-text figure therefore reflects coherent per-head specialization rather than diffuse averaging.}
    \label{fig:app_layer_attention}
\end{figure}

Figure~\ref{fig:app_layer_attention} resolves this ambiguity by displaying the per-head profiles directly, retaining the average over the 24 latent queries. Three observations follow. The vision module is highly homogeneous: three of its four heads concentrate sharply around layers 25--30, while the fourth distributes weight more diffusely over the same mid-depth range; the sharp head-averaged peak in Figure~\ref{fig:layer_attention} therefore reflects a genuine per-head consensus. The audio module shows greater per-head variability, with some heads sharply peaked in the same band and others spreading weight across earlier mid-depth layers ($\sim$10--25). The text module exhibits the clearest head-level division of labor: two heads ($h_0, h_3$) peak around layers 25--30 while the remaining two ($h_1, h_2$) peak around layers 38--40. The broader head-averaged text profile in Figure~\ref{fig:layer_attention} therefore arises from heads that have individually specialized to different stages of the language model, rather than from diffuse, undifferentiated readout.

%% file: appendices/E_encoder_ablations.tex
\section{Brain-Encoder and Temporal Ablations}
\label{app:cr-encoder-ablations}
\label{app:architecture-loss-ablations}

We vary the brain encoder's depth, width, attention heads, and training objective, and evaluate a matched causal temporal encoder. All runs use the native-fusion \modelname{Qwen3-Omni-30B-A3B-Thinking} features, 24-query cross-attention layer poolers, and the training recipe of App.~\ref{app:training-procedure}; only the named factor is varied. Scores are mean \emph{Friends} S06 Pearson $r$ across the four subjects from one training seed, so differences below roughly $0.002$ should not be over-interpreted (the seed SEM in the query-count sweep of App.~\ref{app:n_queries_ablation} is below $10^{-3}$). The tables report the full results, while the text summarizes the main trends.

\subsection{Depth, Width, and Attention Heads}
\label{app:architecture-ablation}

\begin{table}[h]
\centering
\caption{\textbf{Brain-encoder architecture ablations on \emph{Friends} S06.} Each axis varies one factor around the \ourmodel\ configuration (marked $\star$). $\Delta$ is relative to \ourmodel\ ($r = 0.3186$). Head count does not change the parameter count at fixed width.}
\label{tab:architecture-ablation}
\small
\renewcommand{\arraystretch}{1.1}
\begin{tabular}{l r c c}
\toprule
\textbf{Axis} & \textbf{Setting} & $\boldsymbol{r}$ & $\boldsymbol{\Delta}$ \\
\midrule
\multirow{7}{*}{Depth (transformer blocks)}
 & 2  & $0.3126$ & $-0.0060$ \\
 & 4  & $0.3171$ & $-0.0015$ \\
 & 8 $\star$ & $0.3186$ & -- \\
 & 12 & $0.3187$ & $+0.0002$ \\
 & 16 & $0.3198$ & $+0.0012$ \\
 & 24 & $0.3196$ & $+0.0011$ \\
 & 32 & $0.3195$ & $+0.0009$ \\
\midrule
\multirow{5}{*}{Hidden width $D$}
 & 512  & $0.2565$ & $-0.0620$ \\
 & 1024 & $0.2668$ & $-0.0518$ \\
 & 2048 & $0.3126$ & $-0.0060$ \\
 & 3072 $\star$ & $0.3186$ & -- \\
 & 4096 & $0.3215$ & $+0.0030$ \\
\midrule
\multirow{6}{*}{Attention heads}
 & 2  & $0.3176$ & $-0.0010$ \\
 & 4  & $0.3193$ & $+0.0008$ \\
 & 8 $\star$ & $0.3186$ & -- \\
 & 12 & $0.3173$ & $-0.0013$ \\
 & 16 & $0.3173$ & $-0.0013$ \\
 & 32 & $0.3171$ & $-0.0015$ \\
\bottomrule
\end{tabular}
\end{table}

Performance changes little beyond eight transformer blocks, whereas reducing the hidden width below 2{,}048 substantially lowers predictivity (Table~\ref{tab:architecture-ablation}). Varying the number of heads at fixed width has a smaller effect. The default encoder therefore provides a practical balance of predictive accuracy and parameter count (App.~\ref{app:parameter-counts}).

\subsection{Training Objective}
\label{app:loss-ablation}

The paragraph ``Variants we explored'' in App.~\ref{app:training-procedure} summarizes objectives that did not improve over MSE. Table~\ref{tab:loss-ablation} gives the full comparison. ``Pearson'' maximizes the per-parcel correlation between predicted and measured time courses; ``RSA'' and ``CKA'' add or replace the regression loss with a representational-similarity term (representational similarity analysis or centered kernel alignment between predicted and measured response geometries), with and without a logarithmic transform.

\begin{table}[h]
\centering
\caption{\textbf{Training-objective ablation on \emph{Friends} S06.} Mean Pearson $r$ across subjects; $\Delta$ relative to the MSE-trained \ourmodel; one seed per model.}
\label{tab:loss-ablation}
\small
\renewcommand{\arraystretch}{1.1}
\begin{tabular}{l c c}
\toprule
\textbf{Objective} & $\boldsymbol{r}$ & $\boldsymbol{\Delta}$ \\
\midrule
MSE (\ourmodel)     & $0.3186$ & -- \\
Pearson             & $0.3189$ & $+0.0004$ \\
MSE + Pearson       & $0.3178$ & $-0.0008$ \\
MSE + RSA           & $0.3019$ & $-0.0167$ \\
MSE + CKA           & $0.2996$ & $-0.0190$ \\
MSE + log CKA       & $0.2991$ & $-0.0195$ \\
MSE + log RSA       & $0.2950$ & $-0.0236$ \\
CKA                 & $-0.0013$ & $-0.3199$ \\
log CKA             & $-0.0013$ & $-0.3199$ \\
RSA                 & $-0.0016$ & $-0.3202$ \\
log RSA             & $-0.0020$ & $-0.3206$ \\
\bottomrule
\end{tabular}
\end{table}

A pure Pearson objective matches MSE (within $0.001$), and the MSE + Pearson hybrid is within noise of both. Adding a representational-similarity term to MSE reduces $r$ by $0.017$--$0.024$, and representational-similarity objectives alone fail to produce parcel-wise predictions that correlate with the measured signal (they optimize the geometry of the response space rather than the responses themselves). We keep MSE as the default because it is the simplest objective that reaches the plateau and matches the selection metric behavior of the Pearson objective.

\subsection{Matched Causal Temporal Encoder}
\label{app:temporal-causal}
\label{app:causal-encoder}

The brain encoder of \ourmodel\ uses bidirectional self-attention over the 298 stimulus-feature tokens of a window. It never receives measured fMRI as input, so no future BOLD response is exposed; the only non-causality is that the prediction at one position may use stimulus features from later in the same window. To quantify the contribution of this within-window future stimulus context, we train a matched causal control in which every temporal attention layer applies the mask
\begin{equation*}
    M_{ij} = \begin{cases} 0, & j \le i,\\ -\infty, & j > i, \end{cases}
\end{equation*}
so that the representation at token $i$ depends only on fused tokens $Z_{k,0:i}$. The mask adds no parameters; the causal model is trained from scratch as a separate run with the mask active in training, validation, and inference. Table~\ref{tab:causal-design} lists what is matched.

\begin{table}[h]
\centering
\caption{\textbf{Design of the causal temporal-encoder control.} ``Same'' denotes identical inputs and settings, not shared learned weights.}
\label{tab:causal-design}
\small
\renewcommand{\arraystretch}{1.2}
\setlength{\tabcolsep}{4pt}
\begin{tabular}{@{}>{\raggedright\arraybackslash}p{0.22\linewidth} >{\raggedright\arraybackslash}p{0.42\linewidth} >{\raggedright\arraybackslash}p{0.28\linewidth}@{}}
\toprule
\textbf{Component} & \textbf{Bidirectional \ourmodel} & \textbf{Causal control} \\
\midrule
Feature caches           & native \modelname{Qwen3-Omni-30B-A3B-Thinking} & same \\
Layer aggregation        & 24-query, 4-head cross-attention per modality & same \\
Fusion                   & concatenation & same \\
Temporal encoder         & 8 layers, width 3072, 8 heads, FF $4\times$ & same \\
Positions / normalization & learned temporal + rotary positions & same \\
Readout                  & subject-linear head; adaptive $298 \to 100$ pooling & same \\
Training                 & 15 epochs, batch 16, MSE, AdamW, OneCycle, mixed precision & same \\
Split and seed           & \emph{Friends} S06; seed 33 & same \\
Initialization           & trained from scratch & trained from scratch (no checkpoint transfer) \\
Temporal attention       & bidirectional & \textbf{causal mask} \\
\bottomrule
\end{tabular}
\end{table}

Causal temporal attention retains $95.2\%$ of \ourmodel's mean Pearson correlation ($r = 0.3034$ versus $0.3186$), a reduction of $0.015$ that is consistent across all four subjects. Future stimulus context within a window therefore contributes to prediction, while most performance is retained without it. The control combines cutoff-truncated feature extraction with causal temporal masking; together with the hemodynamic offset and readout pooling described in App.~\ref{app:temporal-alignment}, this ensures that its predictions use only stimulus presented before the corresponding BOLD acquisition.

%% file: appendices/F_transfer.tex
\section{Transfer: Data Scaling, Held-Out Subjects, and External Datasets}
\label{app:cr-transfer}
\label{app:data-scaling-subject-transfer}

We test transfer along three axes: less training data, new CNeuroMod subjects, and new participants and stimuli from two external datasets. The first two use CNeuroMod only; the third evaluates frozen checkpoints on HCP 7T Movie and CineBrain.

\subsection{Neural-Data Scaling}
\label{app:data-scaling}

\paragraph{Protocol.} We train \ourmodel\ on nested subsets of the 282 training videos (\emph{Friends} seasons 1--5 and \emph{Movie10}) containing $0\%$, $1\%$, $5\%$, $10\%$, $25\%$, and $50\%$ of the videos (0, 2, 14, 28, 70, and 141 videos). Subsets are nested, so every smaller subset is contained in the larger ones. Each fraction is trained with three independent seeds using the full \ourmodel\ configuration and recipe, and evaluated on the complete \emph{Friends} S06 validation split. The $0\%$ condition evaluates the randomly initialized downstream model without any training on neural data and serves as a chance-level control. \ourmodel\ trained on all data ($100\%$) is shown as a reference.

\begin{table}[h]
\centering
\caption{\textbf{Neural-data scaling of \ourmodel\ on \emph{Friends} S06.} Mean Pearson $r$ across the four subjects, aggregated over three training seeds (mean $\pm$ SD across seeds and 95\% $t$-interval). The last column expresses the seed mean as a fraction of \ourmodel\ trained on all data ($r = 0.3186$).}
\label{tab:data-scaling}
\small
\renewcommand{\arraystretch}{1.1}
\begin{tabular}{r r c c c}
\toprule
\textbf{Training videos} & \textbf{Count} & $\boldsymbol{r}$ \textbf{(mean $\pm$ SD)} & \textbf{95\% CI} & \textbf{\% of full data} \\
\midrule
$0\%$  & 0   & $0.0004 \pm 0.0006$ & $[-0.0011,\,0.0018]$ & $0.1$ \\
$1\%$  & 2   & $0.0544 \pm 0.0002$ & $[0.0538,\,0.0550]$  & $17.1$ \\
$5\%$  & 14  & $0.2231 \pm 0.0089$ & $[0.2009,\,0.2452]$  & $70.0$ \\
$10\%$ & 28  & $0.2478 \pm 0.0026$ & $[0.2413,\,0.2543]$  & $77.8$ \\
$25\%$ & 70  & $0.2800 \pm 0.0014$ & $[0.2765,\,0.2836]$  & $87.9$ \\
$50\%$ & 141 & $0.3016 \pm 0.0004$ & $[0.3006,\,0.3026]$  & $94.7$ \\
\midrule
$100\%$ (\ourmodel, 1 seed) & 282 & $0.3186$ & -- & $100$ \\
\bottomrule
\end{tabular}
\end{table}

\begin{table}[h]
\centering
\caption{\textbf{Per-subject neural-data scaling.} Seed-mean Pearson $r$ (three seeds) for each CNeuroMod subject.}
\label{tab:data-scaling-subjects}
\small
\renewcommand{\arraystretch}{1.1}
\begin{tabular}{r c c c c}
\toprule
\textbf{Training videos} & \textbf{sub-01} & \textbf{sub-02} & \textbf{sub-03} & \textbf{sub-05} \\
\midrule
$0\%$  & $0.0005$ & $0.0004$ & $0.0004$ & $0.0001$ \\
$1\%$  & $0.0529$ & $0.0558$ & $0.0606$ & $0.0484$ \\
$5\%$  & $0.2182$ & $0.2278$ & $0.2383$ & $0.2079$ \\
$10\%$ & $0.2446$ & $0.2524$ & $0.2644$ & $0.2298$ \\
$25\%$ & $0.2752$ & $0.2841$ & $0.3003$ & $0.2605$ \\
$50\%$ & $0.2967$ & $0.3057$ & $0.3221$ & $0.2818$ \\
\bottomrule
\end{tabular}
\end{table}

\begin{figure}[h]
    \centering
    \includegraphics[width=0.55\linewidth]{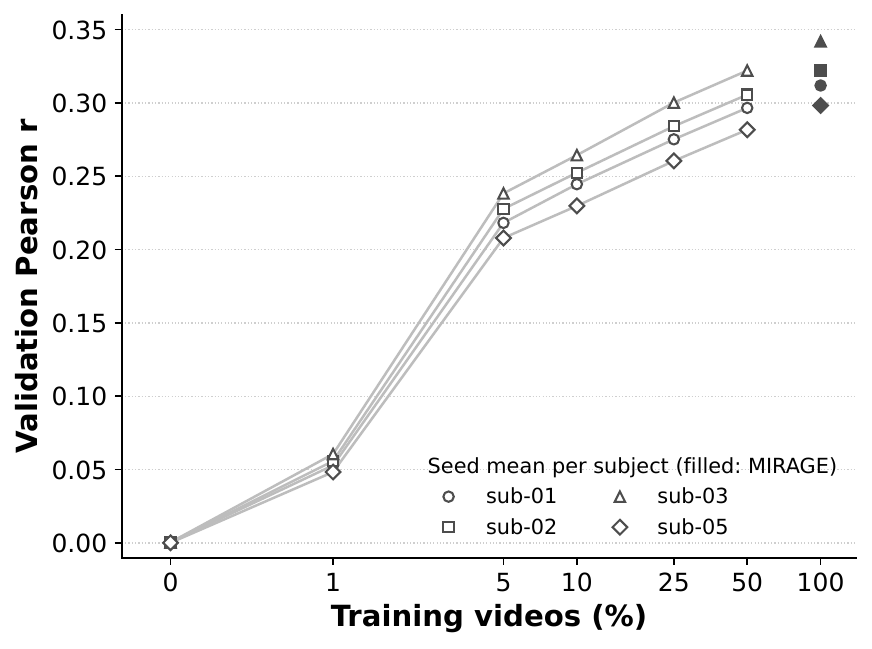}
    \caption{\textbf{Neural-data scaling per subject.} Seed-mean \emph{Friends} S06 Pearson $r$ for each subject (sub-01 $\circ$, sub-02 $\square$, sub-03 $\triangle$, sub-05 $\diamond$) as a function of the fraction of training videos; the filled markers at $100\%$ are \ourmodel\ trained on all data. The horizontal axis is logarithmic above $1\%$.}
    \label{fig:data-scaling}
\end{figure}

Predictivity rises steeply with the first $5\%$ of training videos, reaching $70\%$ of the full-data score with 14 videos. Half of the training videos recover $94.7\%$ of the full-data score (Table~\ref{tab:data-scaling}). The curves rise monotonically for every subject (Figure~\ref{fig:data-scaling}), with smaller gains as more training stimuli are added. The untrained downstream model is at chance.

\subsection{Held-Out-Subject Adaptation}
\label{app:cross-subject}

\paragraph{Protocol.} For each of the four CNeuroMod subjects we train a \emph{source} \ourmodel\ on the other three subjects only, using the full training stimuli and three seeds. The source models reach $r = 0.308$--$0.324$ on their own subjects. The trained model is then frozen, and only a linear readout for the held-out subject is fitted:
\begin{itemize}
    \item \textbf{$0\%$ (zero-shot).} The held-out readout is the mean of the three source-subject heads, and no held-out-subject fMRI is used.
    \item \textbf{$1$--$50\%$.} Starting from the mean source head, the held-out readout is trained on the stated fraction of the held-out subject's training videos (the same nested subsets as in App.~\ref{app:data-scaling}) while all shared parameters remain frozen.
\end{itemize}
The source checkpoint is selected using validation performance on the training subjects. During adaptation, the shared model remains frozen, and the target readout epoch is selected using the held-out subject's S06 validation data from epochs 0--15, including the zero-shot initialization at epoch 0.

As a matched reference we use the from-scratch data-scaling runs of App.~\ref{app:data-scaling} restricted to the same subject, seed, and video fraction. Seeds are averaged within each held-out subject before computing the descriptive subject SEM.

\begin{table}[h]
\centering
\caption{\textbf{Leave-one-subject-out transfer on \emph{Friends} S06.} Transfer $r$ is the mean over the four held-out subjects ($\pm$ subject SEM) after seed averaging. ``From scratch'' is the matched data-scaling run at the same subject, seed, and fraction. The last column counts held-out subjects for which transfer exceeds from-scratch training.}
\label{tab:cross-subject}
\small
\renewcommand{\arraystretch}{1.1}
\setlength{\tabcolsep}{5pt}
\begin{tabular}{r r c c c c c}
\toprule
\textbf{Held-out data} & \textbf{Videos} & \textbf{Transfer $r$} & \textbf{From scratch $r$} & $\boldsymbol{\Delta}$ \textbf{vs.\ $0\%$} & $\boldsymbol{\Delta}$ \textbf{vs.\ scratch} & \textbf{Subjects} \\
\midrule
$0\%$  & 0   & $0.2498 \pm 0.0053$ & $0.0004$ & --       & $+0.2495$ & 4/4 \\
$1\%$  & 2   & $0.2674 \pm 0.0055$ & $0.0544$ & $+0.0176$ & $+0.2129$ & 4/4 \\
$5\%$  & 14  & $0.2972 \pm 0.0077$ & $0.2231$ & $+0.0474$ & $+0.0741$ & 4/4 \\
$10\%$ & 28  & $0.3062 \pm 0.0083$ & $0.2478$ & $+0.0564$ & $+0.0584$ & 4/4 \\
$25\%$ & 70  & $0.3129 \pm 0.0086$ & $0.2800$ & $+0.0631$ & $+0.0329$ & 4/4 \\
$50\%$ & 141 & $0.3164 \pm 0.0087$ & $0.3016$ & $+0.0666$ & $+0.0148$ & 4/4 \\
\bottomrule
\end{tabular}
\end{table}

\begin{table}[h]
\centering
\caption{\textbf{Per-subject held-out transfer.} Seed-mean Pearson $r$ for each held-out subject as a function of the fraction of that subject's training videos used to fit its readout.}
\label{tab:cross-subject-subjects}
\small
\renewcommand{\arraystretch}{1.1}
\begin{tabular}{l c c c c c c}
\toprule
\textbf{Held-out subject} & $\boldsymbol{0\%}$ & $\boldsymbol{1\%}$ & $\boldsymbol{5\%}$ & $\boldsymbol{10\%}$ & $\boldsymbol{25\%}$ & $\boldsymbol{50\%}$ \\
\midrule
sub-01 & $0.2401$ & $0.2615$ & $0.2922$ & $0.3010$ & $0.3071$ & $0.3099$ \\
sub-02 & $0.2557$ & $0.2717$ & $0.3009$ & $0.3098$ & $0.3161$ & $0.3194$ \\
sub-03 & $0.2618$ & $0.2806$ & $0.3163$ & $0.3266$ & $0.3348$ & $0.3387$ \\
sub-05 & $0.2416$ & $0.2556$ & $0.2794$ & $0.2873$ & $0.2939$ & $0.2976$ \\
\bottomrule
\end{tabular}
\end{table}

\begin{figure}[h]
    \centering
    \includegraphics[width=0.55\linewidth]{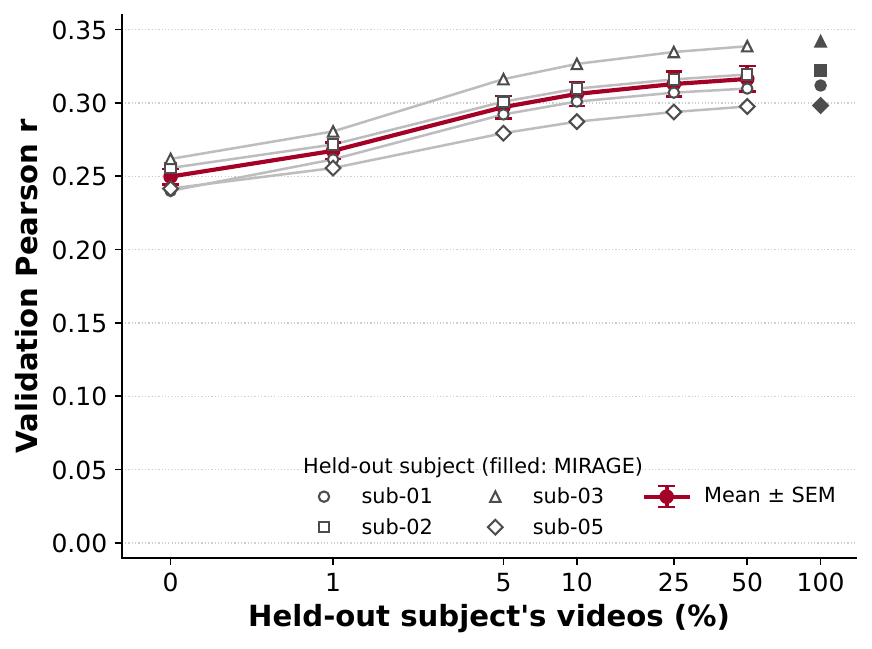}
    \caption{\textbf{Held-out-subject adaptation per subject.} Seed-mean \emph{Friends} S06 Pearson $r$ for each held-out subject (open markers) as a function of the fraction of that subject's training videos used to fit its readout, with the mean $\pm$ subject SEM (red); the filled markers at $100\%$ are \ourmodel\ trained on all four subjects.}
    \label{fig:cross-subject}
\end{figure}

Without target-subject training data, the mean source head reaches $r = 0.250$. Fitting only the readout on $5\%$ of the target subject's videos raises this to $0.297$, and using $50\%$ reaches $0.316$, within $0.002$ of the full-data model. Transfer exceeds the from-scratch reference at every fraction and for every subject, with the largest advantage when target-subject data are scarce (Table~\ref{tab:cross-subject}, Figure~\ref{fig:cross-subject}).

These results support reusing the shared representation and adapting a new participant with a linear readout. They concern CNeuroMod participants evaluated on \emph{Friends} S06; transfer to other datasets is assessed below.

\subsection{External Datasets}
\label{app:external-transfer}
\label{app:external-datasets}

We evaluate frozen CNeuroMod-pretrained representations on two independent naturalistic audiovisual fMRI datasets that differ from CNeuroMod in scanner, participants, stimuli, and acquisition protocol.

\paragraph{HCP 7T Movie.} The Human Connectome Project 7T movie-watching dataset~\citep{vanrssen2013hcp} contains four movie runs per participant composed of short film excerpts, acquired at 7T with a 1\,s repetition time. We use twenty participants with complete data on all four runs and the version-1 stimulus set, selected by participant identifier before any model was evaluated. The preprocessed fsLR dense time series (\texttt{MSMAll}, cleaned) from the HCP minimal preprocessing pipelines~\citep{glasser2013} are used as released.

\paragraph{CineBrain.} CineBrain~\citep{gao2026cinebrain} contains fMRI from six participants watching episodes of \emph{The Big Bang Theory}, with synchronized four-second video and audio clips and preprocessed CIFTI time series. We evaluate all six released participants on episodes s07e01--s07e04 and reconstruct continuous episode-level video and audio from the four-second clips; the released visual captions are not used as stimulus input. As a sensitivity analysis, CineBrain is also evaluated with the stimulus-to-BOLD alignment delayed by a further $4$\,s.

\paragraph{Common target space.} Both datasets are projected from fsLR grayordinates onto the same 1{,}000-parcel Schaefer atlas~\citep{schaefer2017} used for CNeuroMod, with the atlas label order matched to the Algonauts target order. Medial-wall and unassigned vertices are excluded from parcel means, parcel coverage is verified for every participant and run, and subcortical grayordinates are not folded into cortical parcels. Stimulus features use the same cached-feature pipeline as for CNeuroMod (App.~\ref{app:feature-extraction}), with dataset-specific temporal alignment. HCP uses a $4$\,s hemodynamic shift. The original CineBrain fMRI store already incorporates a $4$\,s shift, so no additional shift is applied when loading it; the sensitivity analysis uses an $8$\,s shift in total.

\subsection{External Transfer Protocol}
\label{app:external-protocol}

Ten CNeuroMod-pretrained checkpoints are evaluated: the native fusion, post-hoc fusion (decoder), and post-hoc fusion (tower) models of App.~\ref{app:fusion-conditions} for \modelname{Qwen2.5-Omni-7B}, \modelname{Qwen3-Omni-30B-A3B-Instruct}, and \modelname{Qwen3-Omni-30B-A3B-Thinking} (whose native model is \ourmodel), and our TRIBE~v1-style reproduction. No checkpoint is retrained on the external data. Within each dataset we use four pre-declared leave-one-stimulus-out folds over the four movie runs (HCP) or four stimulus segments (CineBrain), and two conditions:

\begin{itemize}
    \item \textbf{Zero-shot.} The frozen model predicts the external participant's parcels using the mean of its four pretrained CNeuroMod subject heads. No external fMRI is used for fitting.
    \item \textbf{Subject readout.} The representation (the 3{,}072-dimensional brain-encoder output) is frozen, and a participant-specific ridge readout over the 1{,}000 parcels is fitted on the three training stimuli of the fold, with the regularization strength selected per parcel by grouped cross-validation over the training stimuli from a log-spaced grid of 1{,}000 values. The readout is then evaluated on the held-out stimulus.
\end{itemize}

Scores are computed as parcel-wise Pearson $r$ averaged over parcels, then averaged over the four held-out stimuli within each participant, and finally averaged over participants (mean $\pm$ participant SEM). Inference uses participants as the independent units: exact two-sided sign tests on the paired participant differences (ties dropped) and Holm correction over the three backbones within each dataset, alignment, and protocol (App.~\ref{app:statistical-protocol}); the pooled tests treat the HCP and CineBrain participants as one sample of 26. With six CineBrain participants the smallest attainable two-sided exact $p$ is $0.03125$, so CineBrain results are reported as inconclusive rather than negative when they do not pass correction.

\subsection{External Transfer Results}
\label{app:external-results}

\begin{table}[h]
\centering
\caption{\textbf{Transfer of frozen CNeuroMod-pretrained checkpoints to HCP 7T Movie ($n = 20$) and CineBrain ($n = 6$).} Mean parcel-wise Pearson $r \pm$ participant SEM, averaged over four leave-one-stimulus-out folds per participant; CineBrain with the original alignment. All checkpoints are seed 33.}
\label{tab:external-summary}
\footnotesize
\renewcommand{\arraystretch}{1.1}
\setlength{\tabcolsep}{3pt}
\begin{tabular}{@{}l cc cc@{}}
\toprule
\multirow{2}{*}{\textbf{Checkpoint}} & \multicolumn{2}{c}{\textbf{HCP 7T Movie}} & \multicolumn{2}{c}{\textbf{CineBrain}} \\
\cmidrule(lr){2-3} \cmidrule(lr){4-5}
 & Zero-shot & Subject readout & Zero-shot & Subject readout \\
\midrule
\multicolumn{5}{@{}l}{\emph{\modelname{Qwen2.5-Omni-7B}}} \\
\quad Native fusion & $0.1832 \pm 0.0077$ & $0.2498 \pm 0.0126$ & $0.0426 \pm 0.0060$ & $0.0980 \pm 0.0207$ \\
\quad Post-hoc fusion (decoder) & $0.1699 \pm 0.0071$ & $0.2348 \pm 0.0123$ & $0.0424 \pm 0.0053$ & $0.0937 \pm 0.0204$ \\
\quad Post-hoc fusion (tower) & $0.1701 \pm 0.0072$ & $0.2281 \pm 0.0120$ & $0.0428 \pm 0.0056$ & $0.0935 \pm 0.0190$ \\
\multicolumn{5}{@{}l}{\emph{\modelname{Qwen3-Omni-30B-A3B-Instruct}}} \\
\quad Native fusion & $0.1809 \pm 0.0075$ & $0.2473 \pm 0.0124$ & $0.0449 \pm 0.0063$ & $0.1079 \pm 0.0230$ \\
\quad Post-hoc fusion (decoder) & $0.1724 \pm 0.0076$ & $0.2390 \pm 0.0125$ & $0.0472 \pm 0.0059$ & $0.1035 \pm 0.0215$ \\
\quad Post-hoc fusion (tower) & $0.1783 \pm 0.0076$ & $0.2386 \pm 0.0121$ & $0.0467 \pm 0.0064$ & $0.0949 \pm 0.0204$ \\
\multicolumn{5}{@{}l}{\emph{\modelname{Qwen3-Omni-30B-A3B-Thinking}}} \\
\quad Native fusion (\ourmodel) & $0.1827 \pm 0.0079$ & $0.2524 \pm 0.0122$ & $0.0466 \pm 0.0064$ & $0.1095 \pm 0.0232$ \\
\quad Post-hoc fusion (decoder) & $0.1785 \pm 0.0075$ & $0.2468 \pm 0.0123$ & $0.0460 \pm 0.0058$ & $0.0959 \pm 0.0193$ \\
\quad Post-hoc fusion (tower) & $0.1703 \pm 0.0072$ & $0.2345 \pm 0.0124$ & $0.0466 \pm 0.0058$ & $0.0993 \pm 0.0221$ \\
\midrule
TRIBE v1-style (our reproduction) & $0.1773 \pm 0.0075$ & $0.2392 \pm 0.0125$ & $0.0452 \pm 0.0059$ & $0.1008 \pm 0.0204$ \\
\bottomrule
\end{tabular}
\end{table}

\begin{table}[h]
\centering
\caption{\textbf{Native minus post-hoc fusion on the external datasets, by participant.} $\Delta r$ is native minus post-hoc fusion (decoder), mean $\pm$ SEM over participants; ``positive'' counts participants with a positive difference (ties dropped); $p$ is the exact two-sided sign test and Holm $p$ its adjustment over the three backbones within each dataset, alignment, and protocol. The last column gives the native minus post-hoc fusion (tower) difference (descriptive).}
\label{tab:external-inference}
\footnotesize
\renewcommand{\arraystretch}{1.1}
\setlength{\tabcolsep}{3pt}
\begin{tabular}{@{}l l c c c c c@{}}
\toprule
\textbf{Protocol} & \textbf{Backbone} & \textbf{Native $-$ decoder} & \textbf{Positive} & \textbf{Exact $p$} & \textbf{Holm $p$} & \textbf{Native $-$ tower} \\
\midrule
\multicolumn{7}{@{}l}{\emph{HCP 7T Movie ($n = 20$)}} \\
Zero-shot & \modelname{Qwen2.5-Omni-7B} & $+0.0134 \pm 0.0010$ & 20/20 & $<10^{-4}$ & $<10^{-4}$ & $+0.0131$ \\
Zero-shot & \modelname{Qwen3-Omni-30B-A3B-Instruct} & $+0.0086 \pm 0.0007$ & 20/20 & $<10^{-4}$ & $<10^{-4}$ & $+0.0026$ \\
Zero-shot & \modelname{Qwen3-Omni-30B-A3B-Thinking} & $+0.0042 \pm 0.0011$ & 14/20 & $0.12$ & $0.12$ & $+0.0124$ \\
Readout & \modelname{Qwen2.5-Omni-7B} & $+0.0151 \pm 0.0012$ & 20/20 & $<10^{-4}$ & $<10^{-4}$ & $+0.0218$ \\
Readout & \modelname{Qwen3-Omni-30B-A3B-Instruct} & $+0.0083 \pm 0.0011$ & 19/20 & $<10^{-4}$ & $<10^{-4}$ & $+0.0087$ \\
Readout & \modelname{Qwen3-Omni-30B-A3B-Thinking} & $+0.0056 \pm 0.0009$ & 19/20 & $<10^{-4}$ & $<10^{-4}$ & $+0.0179$ \\
\midrule
\multicolumn{7}{@{}l}{\emph{CineBrain ($n = 6$)}} \\
Zero-shot & \modelname{Qwen2.5-Omni-7B} & $+0.0002 \pm 0.0011$ & 2/6 & $0.69$ & $1$ & $-0.0002$ \\
Zero-shot & \modelname{Qwen3-Omni-30B-A3B-Instruct} & $-0.0022 \pm 0.0010$ & 1/6 & $0.22$ & $0.66$ & $-0.0018$ \\
Zero-shot & \modelname{Qwen3-Omni-30B-A3B-Thinking} & $+0.0006 \pm 0.0008$ & 3/6 & $1$ & $1$ & $+0.0001$ \\
Readout & \modelname{Qwen2.5-Omni-7B} & $+0.0043 \pm 0.0030$ & 5/6 & $0.22$ & $0.44$ & $+0.0045$ \\
Readout & \modelname{Qwen3-Omni-30B-A3B-Instruct} & $+0.0043 \pm 0.0032$ & 4/6 & $0.69$ & $0.69$ & $+0.0129$ \\
Readout & \modelname{Qwen3-Omni-30B-A3B-Thinking} & $+0.0135 \pm 0.0050$ & 6/6 & $0.031$ & $0.094$ & $+0.0102$ \\
\midrule
\multicolumn{7}{@{}l}{\emph{CineBrain, alignment delayed by $4$\,s ($n = 6$)}} \\
Zero-shot & \modelname{Qwen2.5-Omni-7B} & $+0.0027 \pm 0.0007$ & 6/6 & $0.031$ & $0.094$ & $+0.0024$ \\
Zero-shot & \modelname{Qwen3-Omni-30B-A3B-Instruct} & $+0.0006 \pm 0.0015$ & 4/6 & $0.69$ & $0.69$ & $+0.0013$ \\
Zero-shot & \modelname{Qwen3-Omni-30B-A3B-Thinking} & $+0.0031 \pm 0.0013$ & 5/6 & $0.22$ & $0.44$ & $+0.0064$ \\
Readout & \modelname{Qwen2.5-Omni-7B} & $+0.0028 \pm 0.0026$ & 5/6 & $0.22$ & $0.66$ & $+0.0048$ \\
Readout & \modelname{Qwen3-Omni-30B-A3B-Instruct} & $+0.0035 \pm 0.0037$ & 5/6 & $0.22$ & $0.66$ & $+0.0106$ \\
Readout & \modelname{Qwen3-Omni-30B-A3B-Thinking} & $+0.0104 \pm 0.0043$ & 5/6 & $0.22$ & $0.66$ & $+0.0126$ \\
\bottomrule
\end{tabular}
\end{table}

\begin{table}[h]
\centering
\caption{\textbf{Pooled native minus post-hoc fusion (decoder) effects ($n = 26$).} HCP and CineBrain participants treated as one sample: mean $\Delta r$ (participants with a positive difference in parentheses); asterisks mark contrasts with Holm $p \le 0.05$ over the three backbones within each alignment and protocol.}
\label{tab:external-native-vs-posthoc}
\small
\renewcommand{\arraystretch}{1.1}
\setlength{\tabcolsep}{4pt}
\begin{tabular}{@{}l c c c@{}}
\toprule
\textbf{Protocol} & \textbf{Qwen2.5-7B} & \textbf{Qwen3-Instruct} & \textbf{Qwen3-Thinking} \\
\midrule
\multicolumn{4}{@{}l}{\emph{Original alignment}} \\
Zero-shot & $+0.0103^{*}$ (22/26) & $+0.0061^{*}$ (21/26) & $+0.0034$ (17/26) \\
Readout & $+0.0126^{*}$ (25/26) & $+0.0074^{*}$ (23/26) & $+0.0074^{*}$ (25/26) \\
\midrule
\multicolumn{4}{@{}l}{\emph{CineBrain delayed by $4$\,s}} \\
Zero-shot & $+0.0109^{*}$ (26/26) & $+0.0067^{*}$ (24/26) & $+0.0039^{*}$ (19/26) \\
Readout & $+0.0122^{*}$ (25/26) & $+0.0072^{*}$ (24/26) & $+0.0067^{*}$ (24/26) \\
\bottomrule
\end{tabular}
\end{table}

\begin{figure}[h]
    \centering
    \includegraphics[width=0.48\linewidth]{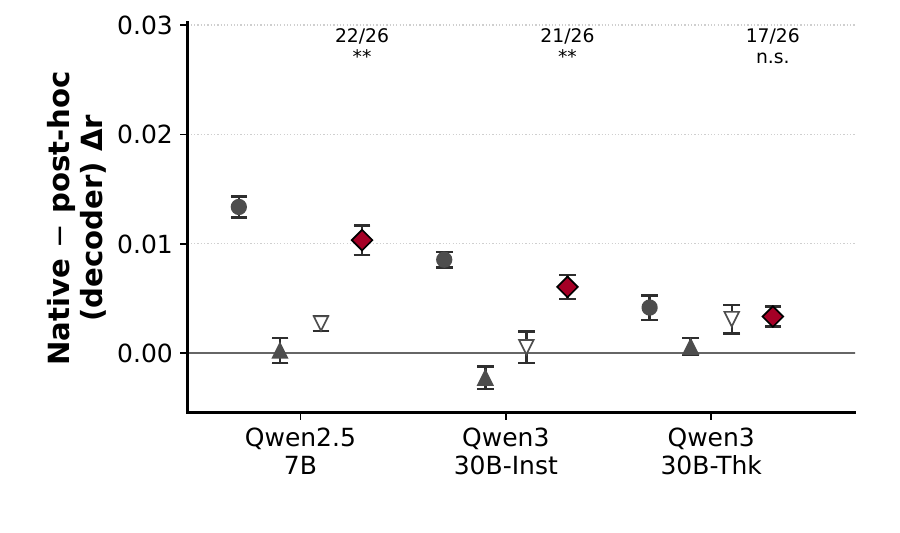}\hfill
    \includegraphics[width=0.48\linewidth]{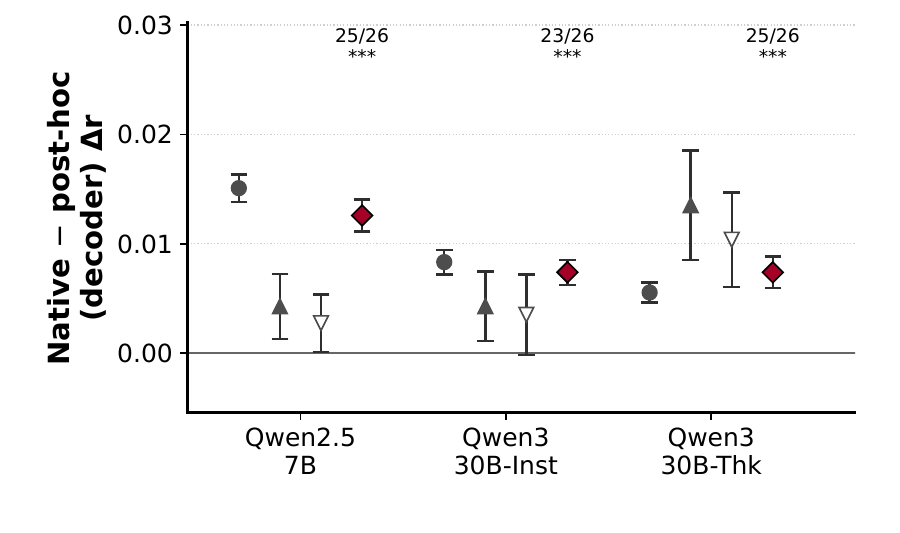}\\[2pt]
    \includegraphics[width=0.48\linewidth]{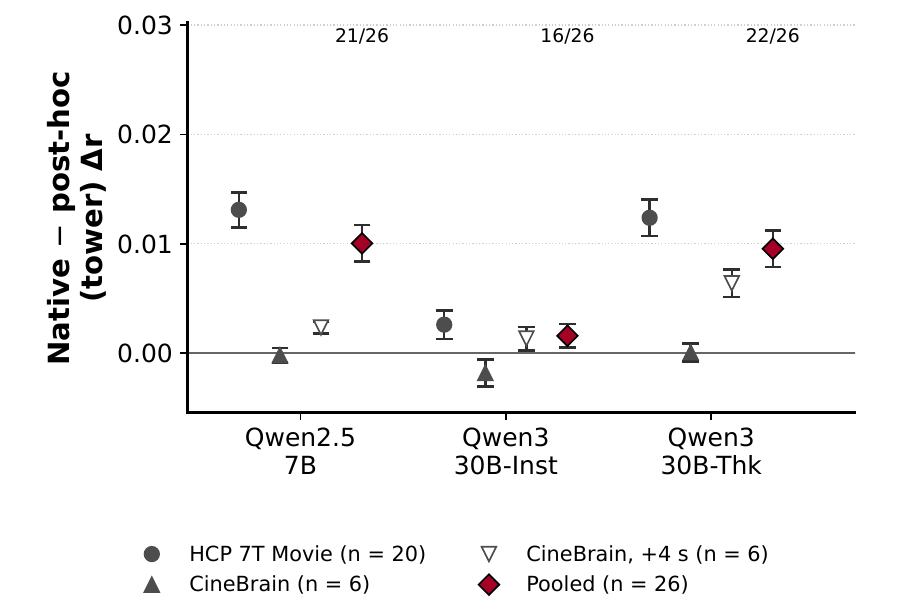}\hfill
    \includegraphics[width=0.48\linewidth]{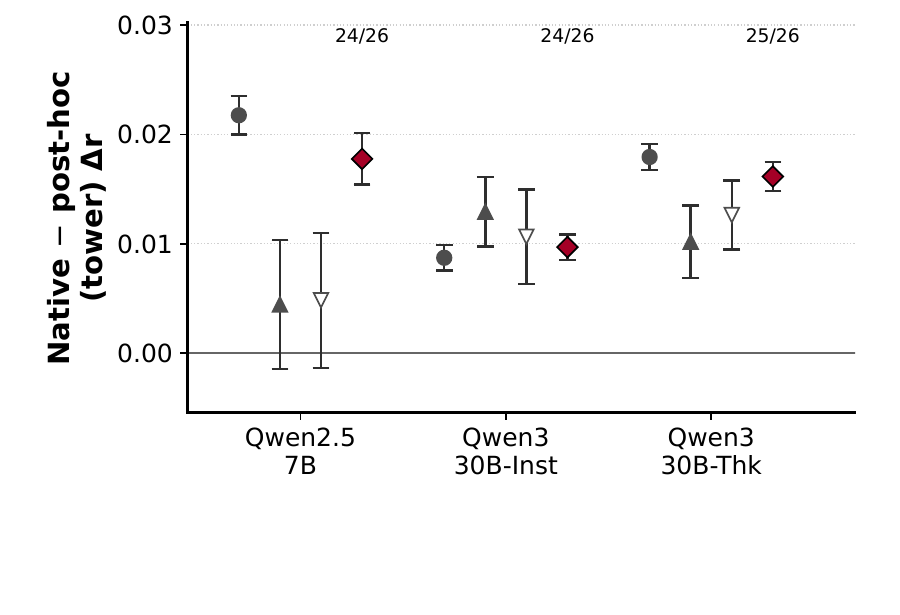}
    \caption{\textbf{Native minus post-hoc fusion on the external datasets.} Top: native minus post-hoc fusion (decoder); bottom: native minus post-hoc fusion (tower). Left: zero-shot; right: subject readout. Mean $\pm$ SEM over participants for HCP 7T Movie ($n = 20$), CineBrain ($n = 6$), CineBrain with the alignment delayed by $4$\,s (sensitivity analysis), and the pooled HCP and CineBrain sample ($n = 26$). Counts are pooled participants with a positive difference; stars (decoder contrasts) are exact two-sided sign tests on the pooled sample, Holm-adjusted over the three backbones within each protocol ($^{*}$ $p \le 0.05$, $^{**}$ $p \le 0.01$, $^{***}$ $p \le 0.001$).}
    \label{fig:external-fusion-contrasts}
\end{figure}

On HCP, native fusion outperforms post-hoc fusion (decoder) after participant-specific readout fitting for all three backbones, with gains of $0.006$--$0.015$ in mean Pearson correlation (all Holm $p < 10^{-4}$). Zero-shot gains also survive correction for 7B and 30B-Instruct; the 30B-Thinking difference is smaller and does not pass correction. Fitting a participant-specific readout improves \ourmodel\ by $0.070$ on average, with gains for all 20 participants.

CineBrain shows a weaker separation. After readout fitting, native fusion has higher mean performance for all three backbones, but none of the contrasts survives correction across backbones. Delaying the alignment by a further $4$\,s yields positive mean native-minus-decoder differences for every backbone in both evaluation protocols (Table~\ref{tab:external-inference}). Fitting a participant-specific readout improves \ourmodel\ by $0.063$ under the original alignment, with gains for all six participants.

In the pooled sample, readout-fitted native fusion exceeds the decoder control for all three backbones after correction; the zero-shot effect also survives correction for 7B and 30B-Instruct. With the delayed CineBrain alignment, all six pooled contrasts (three backbones in each of the two protocols) survive correction (Table~\ref{tab:external-native-vs-posthoc}). The evidence is strongest on HCP; CineBrain provides directionally supportive but inconclusive results. Full scores, participant consistency, and tests are reported in Tables~\ref{tab:external-summary} and~\ref{tab:external-inference}, with the contrasts summarized in Figure~\ref{fig:external-fusion-contrasts}. \ourmodel\ also has higher mean scores than our TRIBE~v1-style reproduction on HCP in both protocols and on CineBrain after readout fitting.

%% file: appendices/G_cortical_maps.tex
\section{Cortical Maps}
\label{app:cr-cortical-maps}

This section collects parcel-wise views of \ourmodel's predictions: the difference from our TRIBE~v1-style reproduction, the unique contribution of each modality on a shared scale, and the held-out Codabench accuracy maps. All maps use the 1{,}000-parcel Schaefer atlas and the seven Yeo--Krienen networks.

\subsection{\ourmodel\ versus TRIBE v1 Across Cortex}
\label{app:mirage-vs-tribe}

The cortical maps of \S\ref{sec:where_what} compare \ourmodel\ against a matched linear baseline, which isolates the contribution of the learned encoder but does not show whether \ourmodel\ improves on a strong non-linear baseline in the same regions. Here we compare \ourmodel\ directly against our TRIBE~v1-style reproduction on the same validation artifacts.

\paragraph{Alignment protocol.} Both models were evaluated on the identical set of 49 stimulus parts from the 23 available \emph{Friends} S06 validation episodes, with predictions and targets aligned time step by time step before scoring. Pearson correlation was computed separately for each subject and each of the 1{,}000 Schaefer parcels~\citep{schaefer2017}, then averaged across subjects for the parcel map and within the seven Yeo--Krienen networks~\citep{ThomasYeo2011} for the network summary. The TRIBE~v1 row is our re-evaluation of the released TRIBE~v1 pipeline on our split (\S\ref{app:parameters-provenance}).

\begin{table}[h]
\centering
\caption{\textbf{\ourmodel\ minus TRIBE~v1 on \emph{Friends} S06, whole cortex and per subject.} Mean Pearson $r$ over 1{,}000 parcels.}
\label{tab:mirage-vs-tribe-overall}
\small
\renewcommand{\arraystretch}{1.1}
\begin{tabular}{l c c c}
\toprule
 & \textbf{\ourmodel} & \textbf{TRIBE v1 (repro.)} & $\boldsymbol{\Delta r}$ \\
\midrule
sub-01 & $0.3119$ & $0.3004$ & $+0.0115$ \\
sub-02 & $0.3221$ & $0.3055$ & $+0.0166$ \\
sub-03 & $0.3420$ & $0.3271$ & $+0.0149$ \\
sub-05 & $0.2983$ & $0.2847$ & $+0.0136$ \\
\midrule
Mean ($\pm$ subject SEM) & $0.3186$ & $0.3044$ & $+0.0142 \pm 0.0011$ \\
Improved parcels (subject-averaged map) & \multicolumn{3}{c}{968 / 1000 (96.8\%)} \\
\bottomrule
\end{tabular}
\end{table}

\begin{table}[h]
\centering
\caption{\textbf{\ourmodel\ minus TRIBE~v1 by Yeo-7 network.} Parcel-averaged Pearson $r$ within each network and the number of parcels whose subject-averaged $\Delta r$ is positive.}
\label{tab:mirage-vs-tribe-networks}
\small
\renewcommand{\arraystretch}{1.1}
\begin{tabular}{l c c c c c}
\toprule
\textbf{Network} & \textbf{Parcels} & \textbf{Improved parcels} & \textbf{\ourmodel} & \textbf{TRIBE v1} & $\boldsymbol{\Delta r}$ \\
\midrule
Visual                      & 162 & 137 (84.6\%)  & $0.4333$ & $0.4257$ & $+0.0075$ \\
Somatomotor                 & 194 & 193 (99.5\%)  & $0.2621$ & $0.2534$ & $+0.0087$ \\
Dorsal Attention            & 122 & 120 (98.4\%)  & $0.3662$ & $0.3480$ & $+0.0182$ \\
Salience / Ventral Attention & 121 & 121 (100\%)  & $0.2686$ & $0.2551$ & $+0.0135$ \\
Limbic                      & 60  & 56 (93.3\%)   & $0.1452$ & $0.1380$ & $+0.0072$ \\
Control                     & 129 & 129 (100\%)   & $0.3030$ & $0.2820$ & $+0.0211$ \\
Default                     & 212 & 212 (100\%)   & $0.3423$ & $0.3222$ & $+0.0201$ \\
\bottomrule
\end{tabular}
\end{table}

\begin{figure}[h]
    \centering
    \begin{tabular}{@{}m{3.05in}@{\hspace{6pt}}m{2.3in}@{}}
    \includegraphics[height=1.7in]{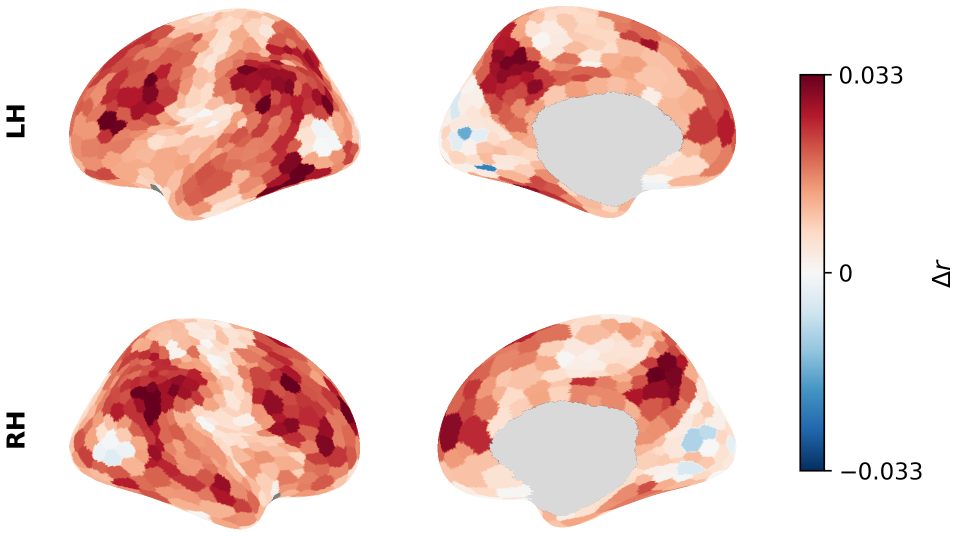} &
    \includegraphics[height=1.7in]{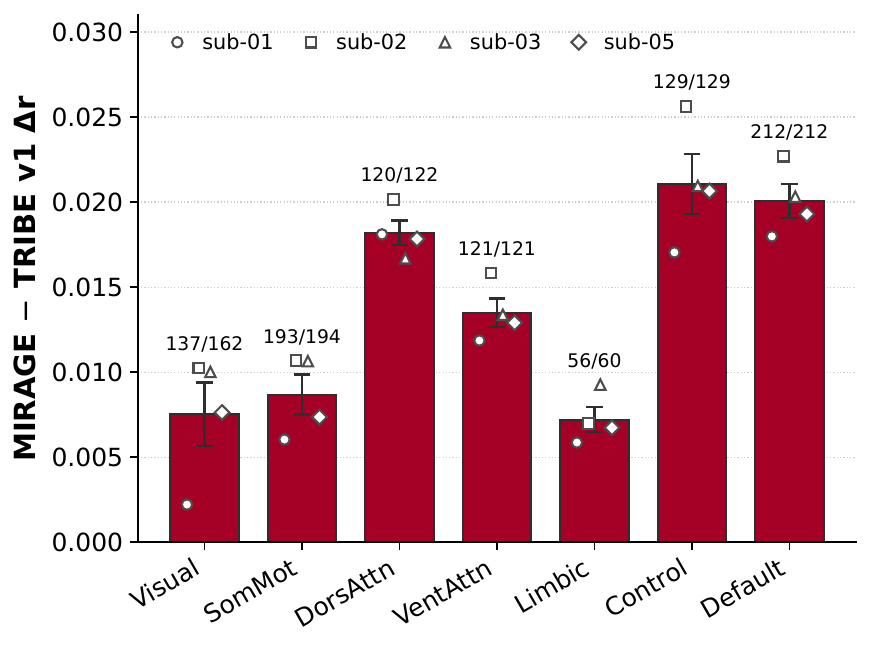} \\
    \end{tabular}
    \caption{\textbf{Where \ourmodel\ improves over TRIBE~v1.} Left: subject-averaged parcel-wise $\Delta r$ (\ourmodel\ minus TRIBE~v1-style reproduction) on the inflated fsaverage surface, lateral and medial views, left hemisphere above right; the color range is $\pm 0.033$ (99th percentile of $|\Delta r|$) and the medial wall is gray. Right: mean $\Delta r$ within each Yeo-7 network (bars: mean $\pm$ SEM across subjects; markers: individual subjects; labels: improved parcels out of parcels in the network).}
    \label{fig:mirage-vs-tribe}
\end{figure}

\ourmodel\ improves on the TRIBE~v1-style reproduction for all four subjects and for 968 of 1{,}000 parcels. The largest network-level gains occur in Control, Default, and Dorsal Attention cortex (Figure~\ref{fig:mirage-vs-tribe}, right). This differs from the comparison against the linear baseline, where the largest gains occur in Visual and Dorsal Attention cortex. The strong non-linear baseline therefore reveals additional improvements in higher-order association regions. The 32 parcels with slightly higher TRIBE~v1 scores lie mostly in early visual cortex on the medial surface (Figure~\ref{fig:mirage-vs-tribe}, left).

\subsection{Shared-Scale Modality Contribution Maps}
\label{app:modality-maps}

The winner-take-all map of Figure~\ref{fig:modality_contrib}b assigns each parcel to the modality whose removal hurts it most. That view is compact but hides parcels to which several modalities contribute jointly. Here we report the unique contribution of each modality on a shared scale with corrected parcel-wise inference.

\paragraph{Definition.} For modality $m \in \{\text{vision},\,\text{audio},\,\text{text}\}$, the unique contribution at parcel $p$ is
\begin{equation*}
    \Delta r_m(p) \;=\; r_{\text{full}}(p) \;-\; r_{\neg m}(p),
\end{equation*}
where $r_{\text{full}}$ is the per-parcel Pearson $r$ of the trimodal model and $r_{\neg m}$ that of a model trained and evaluated with modality $m$ removed (the leave-one-modality-out models of Figure~\ref{fig:modality_contrib}c). Unlike Figure~\ref{fig:modality_contrib}b, which ablates a stream at inference, $r_{\neg m}$ comes from models retrained without modality $m$. Each condition is trained with three seeds; seed-matched differences are averaged inside every resample. Positive values mean that adding modality $m$ to the other two improves prediction of that parcel. The three maps are displayed on one color scale so that their magnitudes are directly comparable.

\paragraph{Inference.} Overall effects (mean over parcels and subjects) are tested with the S06 episode-level protocol of App.~\ref{app:statistical-protocol} and Holm-corrected over the three modalities. Parcel-wise effects are corrected jointly over the 1{,}000 parcels and the three maps with Benjamini--Hochberg FDR and with Westfall--Young max-$T$; we report the number of parcels surviving each correction.

\begin{table}[h]
\centering
\caption{\textbf{Unique modality contributions on \emph{Friends} S06.} Overall $\Delta r$ (mean over parcels and subjects) with 95\% BCa episode-bootstrap CI and Holm-adjusted permutation $p$, and the number of parcels (out of 1{,}000) whose contribution survives joint FDR and joint max-$T$ correction.}
\label{tab:modality-inference}
\small
\renewcommand{\arraystretch}{1.1}
\begin{tabular}{l c c c c}
\toprule
\textbf{Modality} & $\boldsymbol{\Delta r}$ \textbf{[95\% BCa CI]} & \textbf{Holm $p$} & \textbf{FDR parcels} & \textbf{max-$T$ parcels} \\
\midrule
Vision & $+0.0289\ [+0.0267,\,+0.0314]$ & $0.0003$ & 968 & 774 \\
Audio  & $+0.0077\ [+0.0068,\,+0.0087]$ & $0.0003$ & 741 & 234 \\
Text   & $+0.0032\ [+0.0023,\,+0.0040]$ & $0.0003$ & 443 & 61  \\
\bottomrule
\end{tabular}
\end{table}

\begin{table}[h]
\centering
\caption{\textbf{Unique modality contributions by Yeo-7 network.} Mean $\Delta r_m$ over the parcels of each network (descriptive; the corrected tests are in Table~\ref{tab:modality-inference}).}
\label{tab:modality-networks}
\small
\renewcommand{\arraystretch}{1.1}
\begin{tabular}{l c c c}
\toprule
\textbf{Network} & \textbf{Vision} & \textbf{Audio} & \textbf{Text} \\
\midrule
Visual                       & $0.0665$ & $0.0025$ & $0.0019$ \\
Somatomotor                  & $0.0201$ & $0.0118$ & $0.0007$ \\
Dorsal Attention             & $0.0549$ & $0.0068$ & $0.0020$ \\
Salience / Ventral Attention & $0.0207$ & $0.0085$ & $0.0017$ \\
Limbic                       & $0.0053$ & $0.0044$ & $0.0014$ \\
Control                      & $0.0145$ & $0.0077$ & $0.0076$ \\
Default                      & $0.0135$ & $0.0089$ & $0.0060$ \\
\bottomrule
\end{tabular}
\end{table}

\begin{figure}[h]
    \centering
    \setlength{\tabcolsep}{3pt}
    \begin{tabular}{@{}ccc@{}}
    \textbf{Vision} & \textbf{Audio} & \textbf{Text} \\
    \includegraphics[height=1.22in]{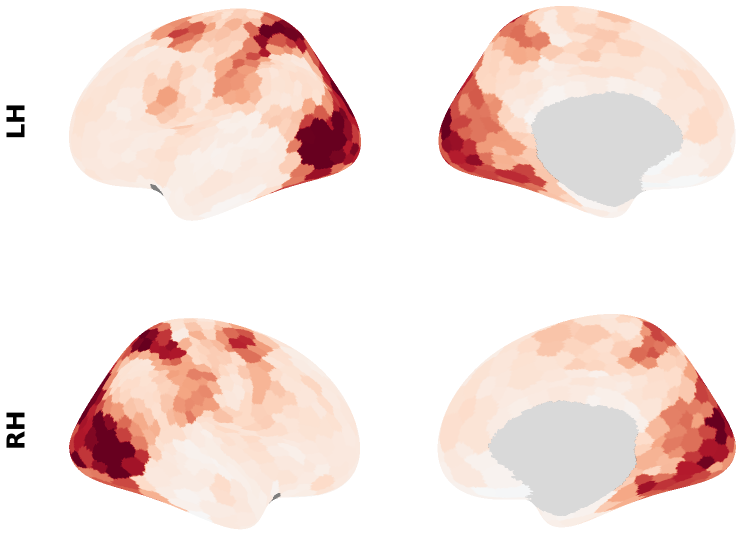} &
    \includegraphics[height=1.22in]{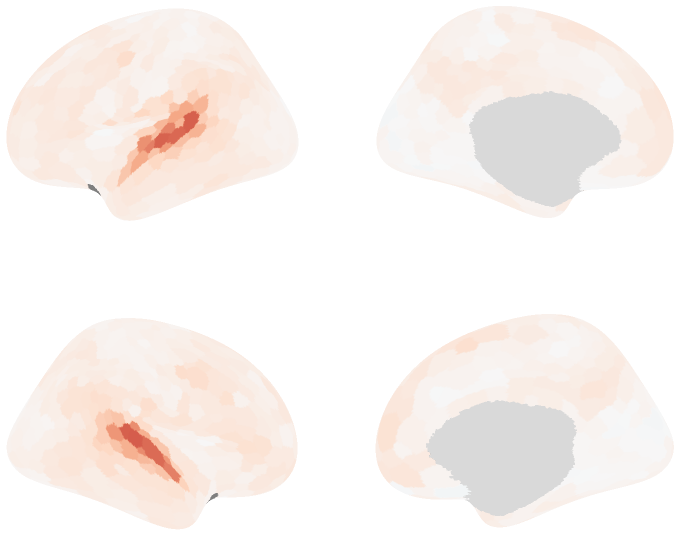} &
    \includegraphics[height=1.22in]{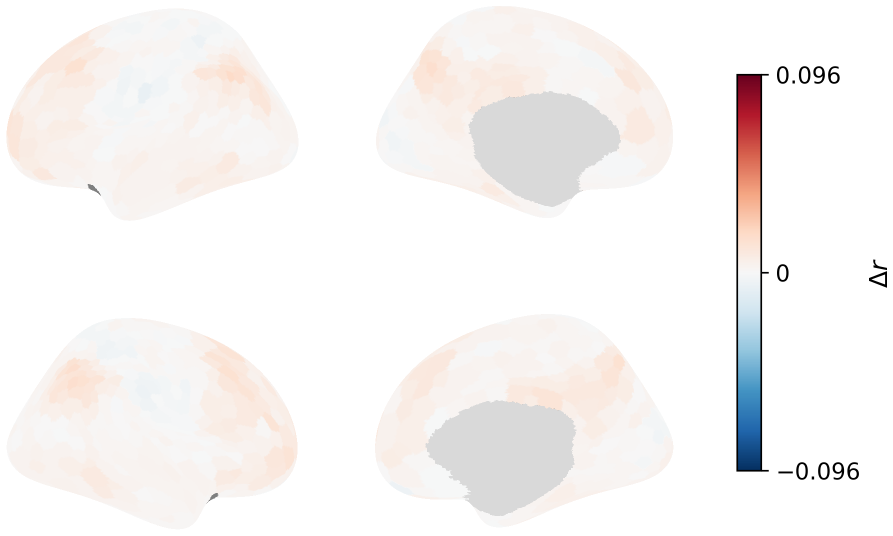} \\
    \end{tabular}
    \caption{\textbf{Unique contribution of each modality on a shared color scale.} Subject-averaged $\Delta r_m$ (full trimodal minus leave-one-modality-out) on the inflated fsaverage surface, lateral and medial views, left hemisphere above right, for vision (left), audio (middle), and text (right). The same color range ($\pm 0.096$, the 99th percentile of $|\Delta r_m|$ over the three maps) is used throughout; the medial wall is gray.}
    \label{fig:modality-surfaces}
\end{figure}

\begin{figure}[h]
    \centering
    \includegraphics[width=0.85\linewidth]{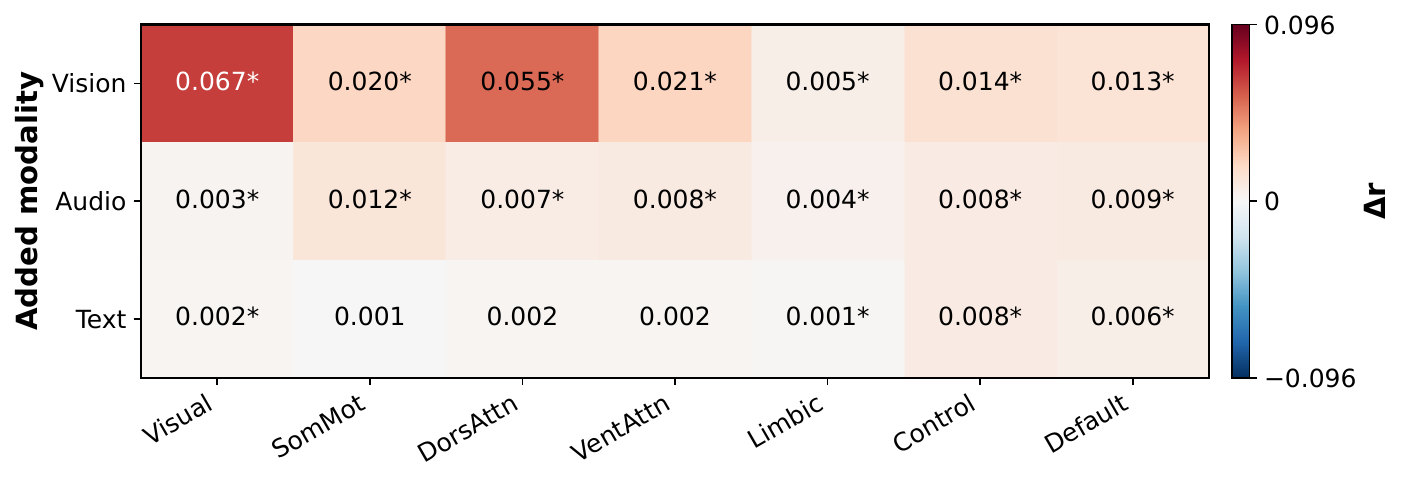}
    \caption{\textbf{Unique modality contribution by Yeo-7 network.} Cell values are the network means of Table~\ref{tab:modality-networks}; asterisks mark networks whose mean contribution is positive under a two-sided episode-permutation test, Holm-adjusted over the seven networks within each modality ($p \le 0.05$).}
    \label{fig:modality-heatmap}
\end{figure}

All three modalities contribute to prediction, with distinct magnitudes and cortical distributions. Vision has the broadest contribution, particularly in Visual and Dorsal Attention cortex. Audio contributes most in Somatomotor cortex around auditory regions, while text is relatively concentrated in Control and Default networks (Figure~\ref{fig:modality-surfaces}).

The network summaries reveal contributions that a winner-take-all map hides. In the Control network, vision, audio, and text contribute $0.015$, $0.008$, and $0.008$, respectively (Table~\ref{tab:modality-networks}, Figure~\ref{fig:modality-heatmap}). These maps describe the encoding model's predictive dependence on its input streams and provide hypotheses for evaluation on independent datasets.

\subsection{Codabench Encoding-Accuracy Maps}
\label{app:codabench-maps}
\begin{figure}[h]
\centering
\textbf{Linear (Native Fusion)}\\
\includegraphics[width=0.48\linewidth]{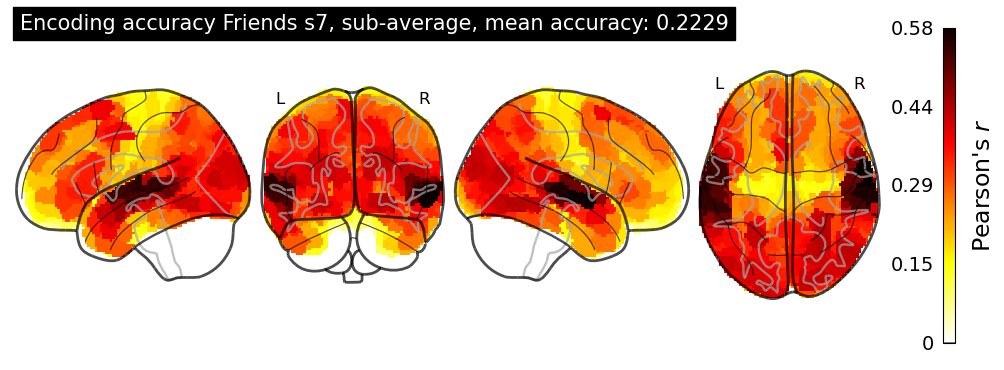}\hfill
\includegraphics[width=0.48\linewidth]{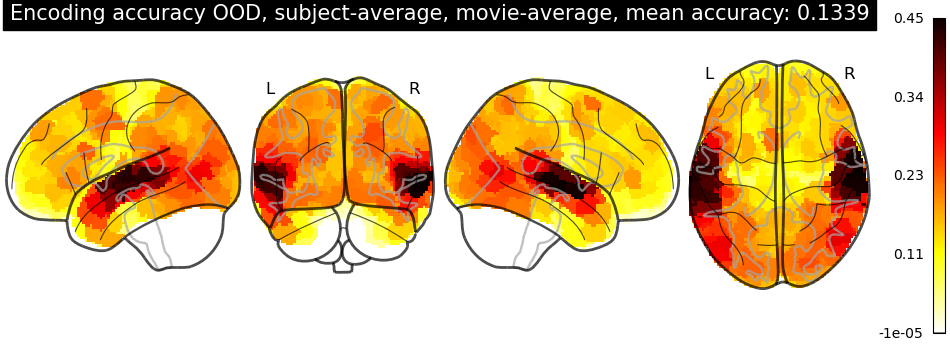}\\[0.6em]
\textbf{\ourmodel\ (single model)}\\
\includegraphics[width=0.48\linewidth]{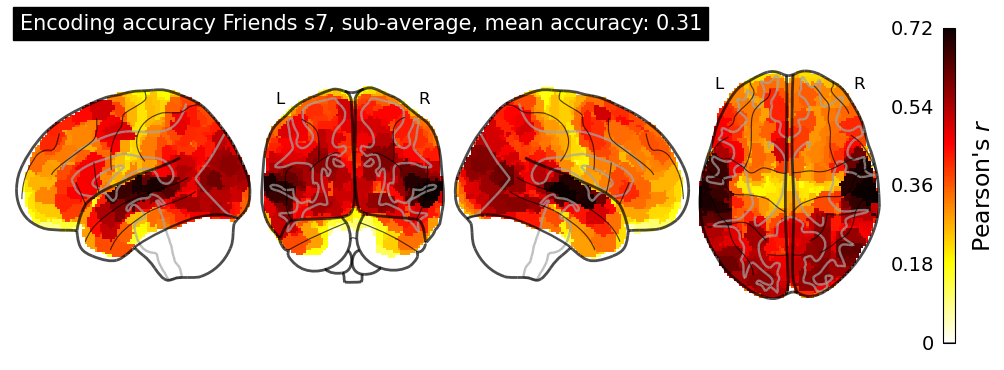}\hfill
\includegraphics[width=0.48\linewidth]{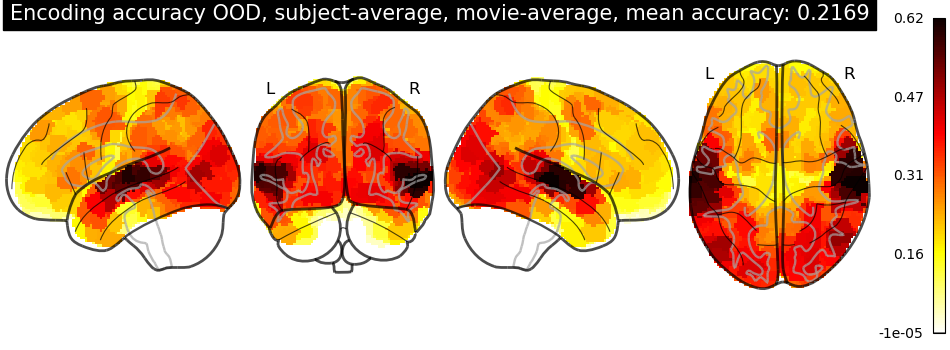}\\[0.6em]
\textbf{\ourmodel\ ($15$-member ensemble)}\\
\includegraphics[width=0.48\linewidth]{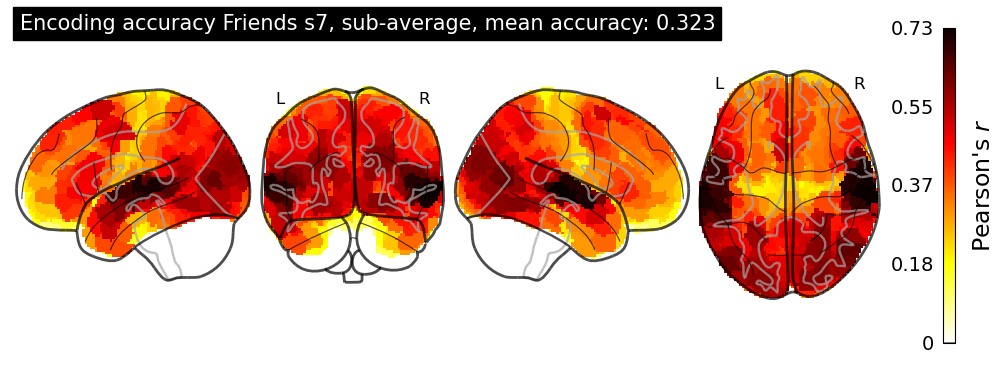}\hfill
\includegraphics[width=0.48\linewidth]{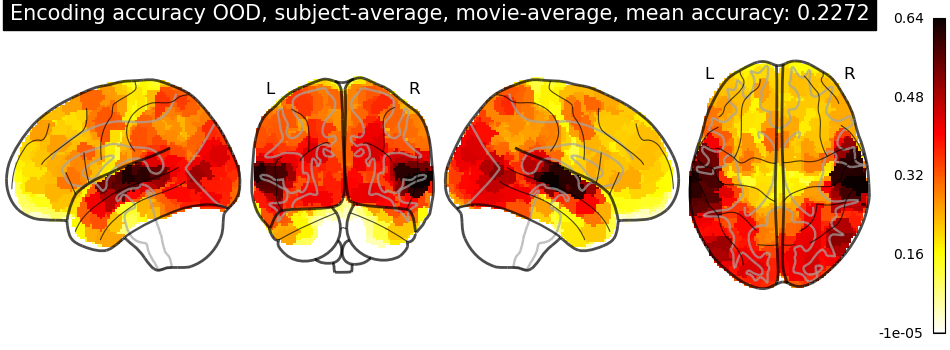}
\caption{\textbf{Subject-averaged encoding-accuracy maps from the Algonauts 2025 Codabench evaluation platform.} Left column: in-distribution test set (\emph{Friends} S07). Right column: out-of-distribution movie set (subject- and movie-averaged). Rows: linear ridge baseline with native-fusion features (top); \ourmodel\ as a single model (middle); \ourmodel\ as the $15$-member ensemble used for our final submission (bottom). Each map shows per-parcel Pearson correlation between predicted and measured BOLD, averaged across the four CNeuroMod subjects.}
\label{fig:codabench-maps}
\end{figure}

For an additional qualitative view of the test-set predictions in Table~\ref{tab:results}, Figure~\ref{fig:codabench-maps} reproduces the subject-averaged encoding-accuracy maps rendered by the Algonauts 2025 Codabench evaluation platform~\citep{gifford2025algonauts} for three of our submitted models: the linear ridge baseline with native-fusion features (\emph{Linear (Native Fusion)}), \ourmodel\ as a single model, and the $15$-member \ourmodel\ ensemble used for our final submission. For each model, the platform produces per-subject and subject-averaged maps on both the in-distribution test set (\emph{Friends} S07) and the out-of-distribution movies; we show the subject-averaged maps here and release the per-subject maps with the code (Appendix~\ref{app:code-release}).

%% file: appendices/H_licenses.tex
\section{Licenses for Existing Assets}
\label{app:licenses}
All existing assets used in this work (pretrained model weights, datasets, atlases, and major software dependencies) are listed in Table~\ref{tab:asset-licenses} together with their providers and current licenses. Pretrained model weights are downloaded from each provider's official model hub (e.g., HuggingFace) and used unmodified for feature extraction; we do not redistribute model weights. The Courtois NeuroMod dataset~\citep{boyle2020courtois_cneuromod} is accessed via its open-data release, and the Algonauts 2025 challenge test sets~\citep{gifford2025algonauts} are accessed only through the official Codabench evaluation platform. Where individual licenses include non-commercial clauses, our use of those assets is restricted to academic research and is consistent with those terms.

\begin{table}[h]
\centering
\caption{Licenses for existing assets used in this work. Researchers should consult each provider's model card or repository for current license terms before redistribution.}
\label{tab:asset-licenses}
\small
\setlength{\tabcolsep}{2pt}
\renewcommand{\arraystretch}{1}
\begin{tabular}{@{}lll@{}}
\toprule
\textbf{Asset} & \textbf{Provider} & \textbf{License} \\
\midrule
\multicolumn{3}{@{}l}{\emph{Multimodal foundation models}} \\
\modelname{Qwen3-Omni-30B-A3B-Thinking}~\citep{xu2025qwen3omnitechnicalreport} & Alibaba Cloud & Apache-2.0 \\
\modelname{Qwen3-Omni-30B-A3B-Instruct}~\citep{xu2025qwen3omnitechnicalreport} & Alibaba Cloud & Apache-2.0 \\
\modelname{Qwen2.5-Omni-7B}~\citep{xu2025qwen25omnitechnicalreport} & Alibaba Cloud & Apache-2.0 \\
\modelname{Qwen2.5-Omni-3B}~\citep{xu2025qwen25omnitechnicalreport} & Alibaba Cloud & Qwen Research License \\
\midrule
\multicolumn{3}{@{}l}{\emph{Unimodal baselines (TRIBE-style)}} \\
\modelname{Llama-3.2-3B}~\citep{grattafiori2024llama3herdmodels} & Meta AI & Llama 3.2 Community License \\
\modelname{Wav2Vec-BERT-2.0}~\citep{chung2021wav2vecbert} & Meta AI & MIT \\
\modelname{V-JEPA~2}~\citep{assran2025vjepa2} & Meta AI & MIT \\
\midrule
\multicolumn{3}{@{}l}{\emph{Datasets and atlases}} \\
Courtois NeuroMod~\citep{boyle2020courtois_cneuromod} & Courtois Project & CC-BY-SA~4.0 \\
Algonauts 2025 challenge data~\citep{gifford2025algonauts} & Algonauts organizers & CNeuroMod-derived \\
HCP 7T movie-watching data~\citep{vanrssen2013hcp} & WU-Minn HCP Consortium & HCP Open Access Data Use Terms \\
CineBrain~\citep{gao2026cinebrain} & Gao et al. & Apache-2.0 \\
Schaefer 1000-parcel atlas~\citep{schaefer2017} & Yeo Lab & MIT \\
Yeo--Krienen 7 networks~\citep{ThomasYeo2011} & Yeo Lab / FreeSurfer & Open non-commercial research use \\
\midrule
\multicolumn{3}{@{}l}{\emph{Software}} \\
PyTorch~2.7 & Linux Foundation / Meta & BSD-3-Clause (modified) \\
Python~3.12 & Python Software Foundation & PSF License \\
HuggingFace Transformers & HuggingFace, Inc. & Apache-2.0 \\
\bottomrule
\end{tabular}
\end{table}